\documentclass[sn-nature]{sn-jnl}

\usepackage{graphicx}%
\usepackage{multirow}%
\usepackage{amsmath,amssymb,amsfonts}%
\usepackage{amsthm}%
\usepackage{mathrsfs}%
\usepackage[title]{appendix}%
\usepackage{xcolor}%
\usepackage{textcomp}%
\usepackage{manyfoot}%
\usepackage{booktabs}%
\usepackage{algorithm}%
\usepackage{algorithmicx}%
\usepackage{algpseudocode}%
\usepackage{listings}%
\usepackage{array}
\usepackage{colortbl}
\usepackage{xcolor}
\usepackage{subcaption}

\theoremstyle{thmstyleone}%
\theoremstyle{thmstyletwo}%

\theoremstyle{thmstylethree}%

\begin{document}

\title[Decoding Continuous Character-based Language from Non-invasive Brain Recordings]{Decoding Continuous Character-based Language from Non-invasive Brain Recordings}



\author[1,2]{\fnm{Cenyuan} \sur{Zhang}}
\equalcont{These authors contributed equally to this work.}

\author*[1,2]{\fnm{Xiaoqing} \sur{Zheng}}\email{zhengxq@fudan.edu.cn}
\equalcont{These authors contributed equally to this work.}

\author[1,2]{\fnm{Ruicheng} \sur{Yin}}

\author[3,4]{\fnm{Shujie} \sur{Geng}}

\author[1,2]{\fnm{Jianhan} \sur{Xu}}

\author[1,2]{\fnm{Xuan} \sur{Gao}}

\author[1,2]{\fnm{Changze} \sur{Lv}}

\author[1,2]{\fnm{Zixuan} \sur{Ling}}

\author[1,2]{\fnm{Xuanjing} \sur{Huang}}

\author*[3,4]{\fnm{Miao} \sur{Cao}}\email{mcao@fudan.edu.cn}

\author*[3,4]{\fnm{Jianfeng} \sur{Feng}}\email{jianfeng64@gmail.com}

\affil[1]{\orgdiv{School of Computer Science}, \orgname{Fudan University}, \orgaddress{\city{Shanghai}, \country{China}}}

\affil[2]{\orgdiv{Shanghai Key Laboratory of Intelligent Information Processing}, \orgname{Fudan University}, \orgaddress{\city{Shanghai}, \country{China}}}

\affil[3]{\orgdiv{Institute of Science and Technology for Brain-Inspired Intelligence}, \orgname{Fudan University}, \orgaddress{\city{Shanghai}, \country{China}}}

\affil[4]{\orgdiv{Key Laboratory of Computational Neuroscience and Brain-Inspired Intelligence (Fudan University)}, \orgname{Ministry of Education}, \orgaddress{\country{China}}}


\abstract{
Deciphering natural language from brain activity through non-invasive devices remains a formidable challenge. Previous non-invasive decoders either require multiple experiments with identical stimuli to pinpoint cortical regions and enhance signal-to-noise ratios in brain activity, or they are limited to discerning basic linguistic elements such as letters and words. We propose a novel approach to decoding continuous language from single-trial non-invasive fMRI recordings, in which a three-dimensional convolutional network augmented with information bottleneck is developed to automatically identify responsive voxels to stimuli, and a character-based decoder is designed for the semantic reconstruction of continuous language characterized by inherent character structures. The resulting decoder can produce intelligible textual sequences that faithfully capture the meaning of perceived speech both within and across subjects, while existing decoders exhibit significantly inferior performance in cross-subject contexts. The ability to decode continuous language from single trials across subjects demonstrates the promising applications of non-invasive language brain-computer interfaces in both healthcare and neuroscience.

}




\maketitle

\section*{Introduction}\label{sec:introduction}

Over the past decade, advancements in brain-computer interfaces have demonstrated the feasibility of decoding various forms of communication, such as speech sounds \cite{pei2011decoding,akbari2019towards}, hand gestures \cite{stavisky2018decoding,willett2021high}, articulatory movements \cite{moses2021neuroprosthesis,anumanchipalli2019speech}, and other signals \cite{pasley2012reconstructing} from intracranial recordings. 
Despite their efficacy, the requirement for invasive brain surgery limits the applicability of these decoding methods to patients with severe impediments in speech or communication due to neurodegenerative diseases, strokes, or traumatic brain injuries.
In contrast, non-invasive recordings, particularly those employing functional magnetic resonance imaging (fMRI) \cite{tang2023semantic,xi2023unicorn}, magnetoencephalography (MEG) and electroencephalography (EEG) \cite{defossez2023decoding}, have demonstrated the ability to record rich linguistic information, and decoding natural language from such non-invasive recordings holds the potential for broader applications in both restorative interventions and augmentative technologies.

Previous efforts to decode natural language from non-invasive recordings have primarily focused on recognizing letters, words, or fragments within a predetermined set of possibilities \cite{farwell1988talking, mitchell2008predicting, pereira2018toward, dash2020decoding, xi2023unicorn, defossez2023decoding}. 
A recent breakthrough has demonstrated the feasibility of decoding continuous language from non-invasive recordings of native English speakers \cite{fedorenko2010new}.
However, their approach requires a crucial pre-processing step in which multiple experiments with identical stimuli should be repeatedly conducted to locate associated cortical regions \cite{tang2023semantic}.
This requirement is due to the inherent noise in non-invasive recordings, which varies across trials and subjects. Additionally, prior studies have shown the variability in the anatomical location of speech networks across individuals \cite{fedorenko2010new}.
While effective, a decoder using linear regression to model the influence of speech sounds on brain responses is limited to a specific individual for whom the decoder was learned and may not generalize well across different individuals.

\begin{figure}
    \centering
    \includegraphics[width = 0.96\textwidth]{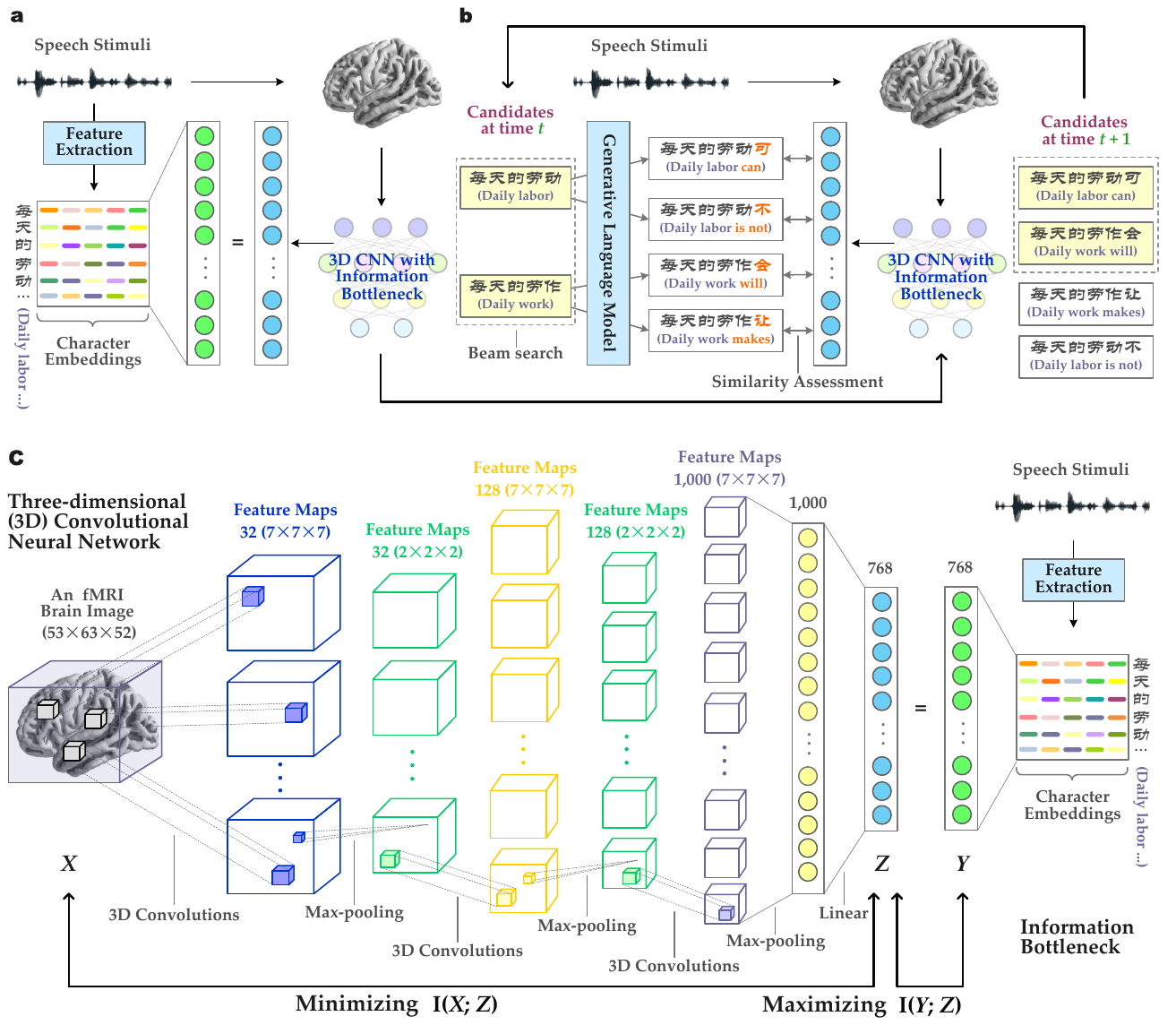}
    \vspace{3mm}
    \caption{\small Character-based language decoder. \textbf{(a)} BOLD fMRI responses were recorded from 20 subjects during a 2.7-hour of listening to naturally spoken narratives. A neural network encoder was learned for each subject (in a within-subject setting) or multiple subjects (in a cross-subject setting) to extract the feature representations from fMRI brain images. Those feature representations are expected to match the semantic features of the stimulus character sequence. 
    \textbf{(b)} To reconstruct continuous language from novel brain recordings, the decoder maintains a set of candidate character sequences. As new characters are detected, a large language model (LLM) proposes continuations for each sequence, and the encoder estimates the likelihood of each proposed sequences based on the recorded brain responses. The most likely continuations are retained until no further characters are detected.
    \textbf{(c)} A three-dimensional (3D) convolutional neural network enhanced with an information bottleneck (IB) was developed to extract semantic features from fMRI brain images. The 3D convolutions were used to enhance invariance to distortions in fMRI images, and the IB was introduced to maximize the predictive power of the extracted feature representations by the network while mitigating the inclusion of irrelevant and noisy information in BOLD fMRI recordings.}
    \label{fig:illustration}
\end{figure}

Here we introduce a decoder designed to reconstruct continuous natural language from deep representations of non-invasive brain recordings obtained through functional magnetic resonance imaging (fMRI). 
Our apporach integrates three-dimensional (3D) convolutional neural networks (CNNs) with information bottleneck (IB), facilitating the extraction of semantic features from brain responses elicited by naturally spoken narratives. 
By incorporating an information bottleneck, we nonlinearly compress brain recordings, enabling the identification of more responsive voxels and thereby enhancing signal-to-noise ratios in brain activity (Fig. \ref{fig:illustration}). 
The application of deep neural networks enables the learning of invariant feature representations across different subjects exposed to the same stimuli of spoken narratives. 
Consequently, the trained decoders demonstrate the potential for generalization across different individuals. 
Furthermore, the application of end-to-end deep neural network architecture for language decoding allow us to identify cortical areas that are primarily engaged or activated during semantic processing.

\begin{figure}
    \centering
    \includegraphics[width = 1.0\textwidth]{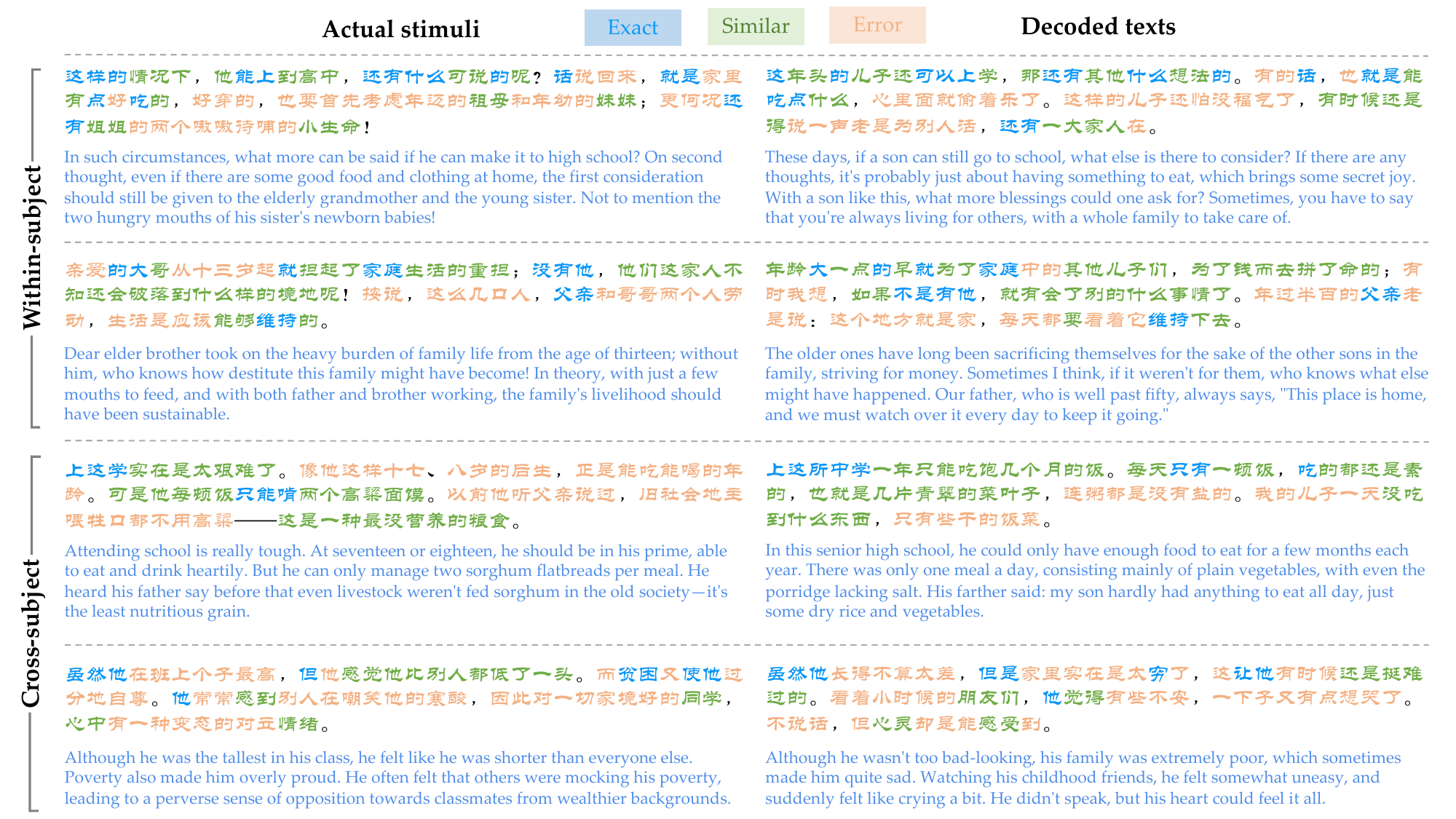}
    \vspace{3mm}
    \caption{\small Decoders were evaluated on single-trial brain responses recorded while the subjects listened to the test articles, which were not used in model training.
    The comparisons between decoder predictions and the actual stimuli are shown for both within-subject and cross-subject settings. 
    The examples were manually selected and annotated to demonstrate typical decoder behaviors. The decoder can exactly reproduce some characters, words and phrases, as well as effectively grasp the similar meanings of many more. English translations corresponding to the actual stimuli and decoded texts are provided below for reference.}
    \label{fig:decoding-results}
\end{figure}

Although fMRI offers exceptional spatial resolution, its measurement of the blood-oxygen-level-dependent (BOLD) signal is notably slow. 
The rise and fall of BOLD signal caused by an impulse of neural activity unfold gradually over a duration of approximately ten seconds \cite{logothetis2003underpinnings}.
This implies that each brain image can be affected by multiple words, posing a challenge as there are considerably more words to decode than brain images.
In line with the approach proposed in \cite{tang2023semantic}, we address this challenge by employing a large language model (LLM) to generate candidate textual sequences, estimating the similarity between the distributed embedding of each candidate sequence and the deep representation of brain recordings, and then choosing the best candidate.   

To efficiently search for the most probable textual sequence, we generates candidate sequences character by character.
Unlike English, innate character-based languages such as Chinese and Japanese, lack clearly defined boundaries for words. 
The definition of Chinese words, or segmentation units, exhibits significant variability across individuals. 
While previous research has established that English words are processed holistically, evidenced by more efficient recognition of related words than unrelated letters \cite{cattell1886time}, the holistic processing of Chinese words remains uncertain.
This uncertainty arises from the challenge of reaching a consensus on word segmentation among individuals, even though language comprehension does not appear to be influenced by discrepancies in word segmentation results \cite{wang1999reading}.

\begin{table}[t]
  \setlength{\tabcolsep}{1.1mm}
  \caption{\label{tab:main_result_within} Language similarity scores in a within-subject setting}
    \begin{tabular}{l|ccc|ccc|ccc|ccc}
      
    \hline
   \rowcolor{gray!20} \multirow{2}{*}{} & \multicolumn{3}{c|}{\textbf{BLEU}} & \multicolumn{3}{c|}{\textbf{METEOR}} &
    \multicolumn{3}{c|}{\textbf{BERT}} & \multicolumn{3}{c}{\textbf{SBERT}} \\
   \rowcolor{gray!20} & NULL & BL &  3dC-IB  & NULL & BL & 3dC-IB & NULL & BL &  3dC-IB & NULL & BL & 3dC-IB \\
    \hline
    
    S1 & 0.167 & 0.184 & \textbf{0.216} & 0.096 & 0.107 & \textbf{0.119} & 0.519 & \textbf{0.536} & \textbf{0.536} & 0.268 & 0.374 & \textbf{0.413} \\
    
    \rowcolor{gray!20} S2 & 0.167 & 0.168 & \textbf{0.203} & 0.099 & 0.100 & \textbf{0.118} & 0.520 & 0.531 & \textbf{0.537} & 0.274 & 0.344 & \textbf{0.409} \\
    
    S3 & 0.170 & 0.179 & \textbf{0.223} & 0.099 & 0.107 & \textbf{0.129} & 0.519 & 0.533 & \textbf{0.536} & 0.263 & 0.302 & \textbf{0.395} \\
    
    \rowcolor{gray!20} S4 & 0.172 & 0.154 & \textbf{0.215} & 0.100 & 0.091 & \textbf{0.121} & 0.522 & 0.525 & \textbf{0.531} & 0.279 & 0.278 & \textbf{0.406} \\
    
    S5 & 0.172 & 0.160 & \textbf{0.229} & 0.099 & 0.096 & \textbf{0.132} & 0.519 & 0.526 & \textbf{0.540} & 0.287 & 0.305 & \textbf{0.397} \\
    
    \rowcolor{gray!20} S6 & 0.160 & 0.139 & \textbf{0.217} & 0.100 & 0.091 & \textbf{0.131} & 0.521 & 0.518 & \textbf{0.537} & 0.287 & 0.245 & \textbf{0.371} \\
    
    \hline
    \end{tabular}
    {Decoder predictions for a test article were compared to the actual stimulus sequence of $2,040$ characters using four language similarity metrics: BLEU, METEOR, BERT, and SBERT, in a within-subject setting. To establish a null distribution, the mean similarity between the actual stimulus and $200$ null sequences generated by a language model was computed with brain data only used to estimate the number of perceived characters. The abbreviation ``BL'' denotes the baseline model \cite{tang2023semantic} employing a regularized linear model for decoding. The 3D convolutional network with an information bottleneck (3dC-IB) consistently outperformed the baseline across all subjects and metrics.}

\end{table}

To address this uncertainty, we introduce a character-based decoder designed to identify the most likely perceived speech stimuli. 
This is accomplished by comparing recorded brain responses to possible character sequences using a beam search algorithm \cite{tillmann2003word}. 
At each time step, a beam contains the $k$ most likely candidate sequences, and the language model generates the next character for each sequence in the beam using the previously-decoded sequence as context. 
The encoder then estimates the similarity between each extended sequence and the recorded brain responses in the form of their distributed representations, and retains the $k$ most likely sequences in the beam for the subsequent time step. This iterative process continuously approximates the most likely stimulus characters over an arbitrary duration.
Experimental results indicate that such character-based decoders can produce intelligible character sequences (Fig. \ref{fig:decoding-results}), faithfully capturing and recovering the meaning of naturalistic speech in both within-subject and cross-subject contexts.


\begin{figure} 
    \centering
    \begin{subfigure}[valign=t]{0.49\textwidth}
        \includegraphics[width=\textwidth]{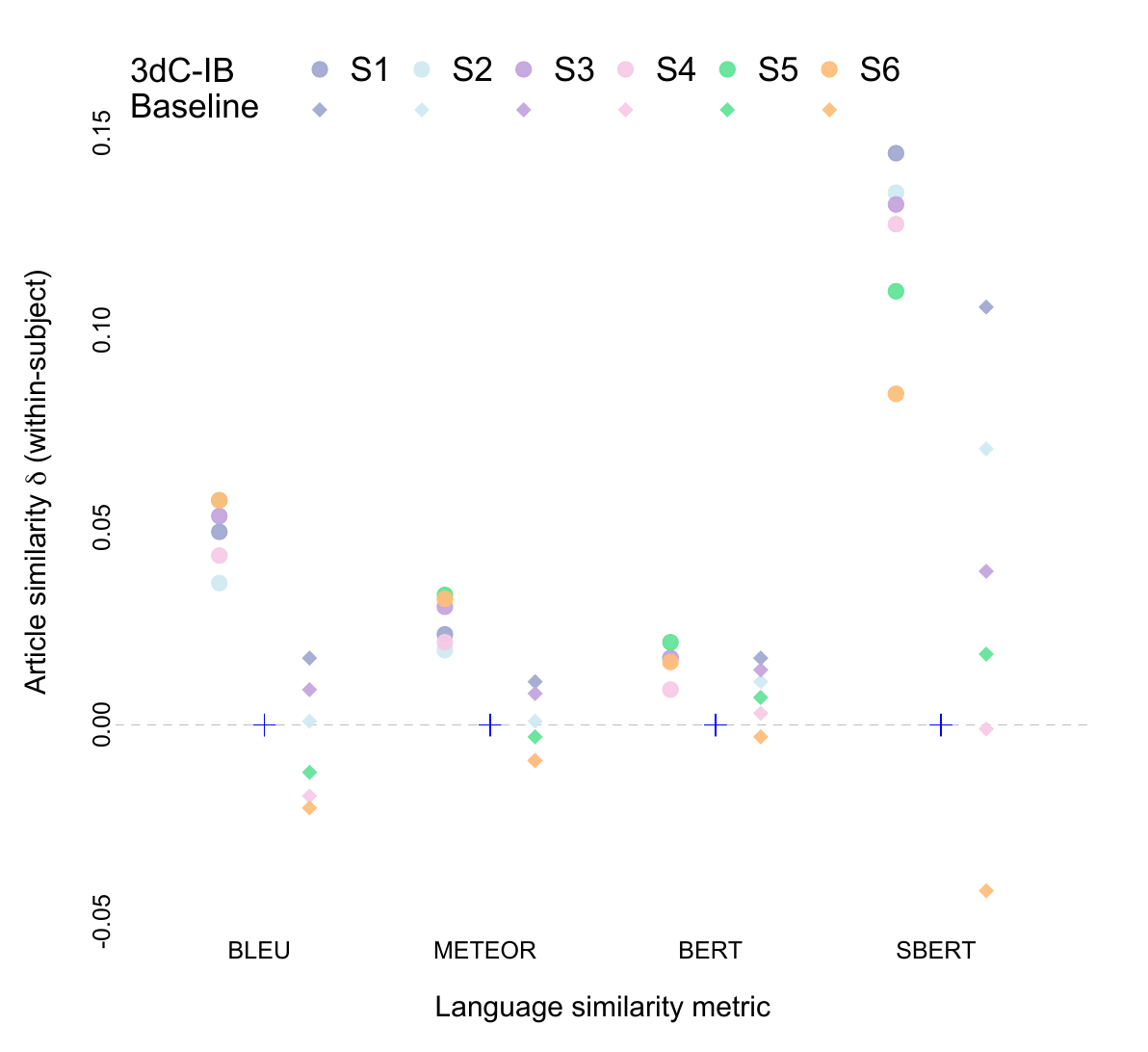}
        \label{fig:within-subject}
    \end{subfigure}
    \begin{subfigure}[valign=t]{0.49\textwidth}
        \includegraphics[width=\textwidth]{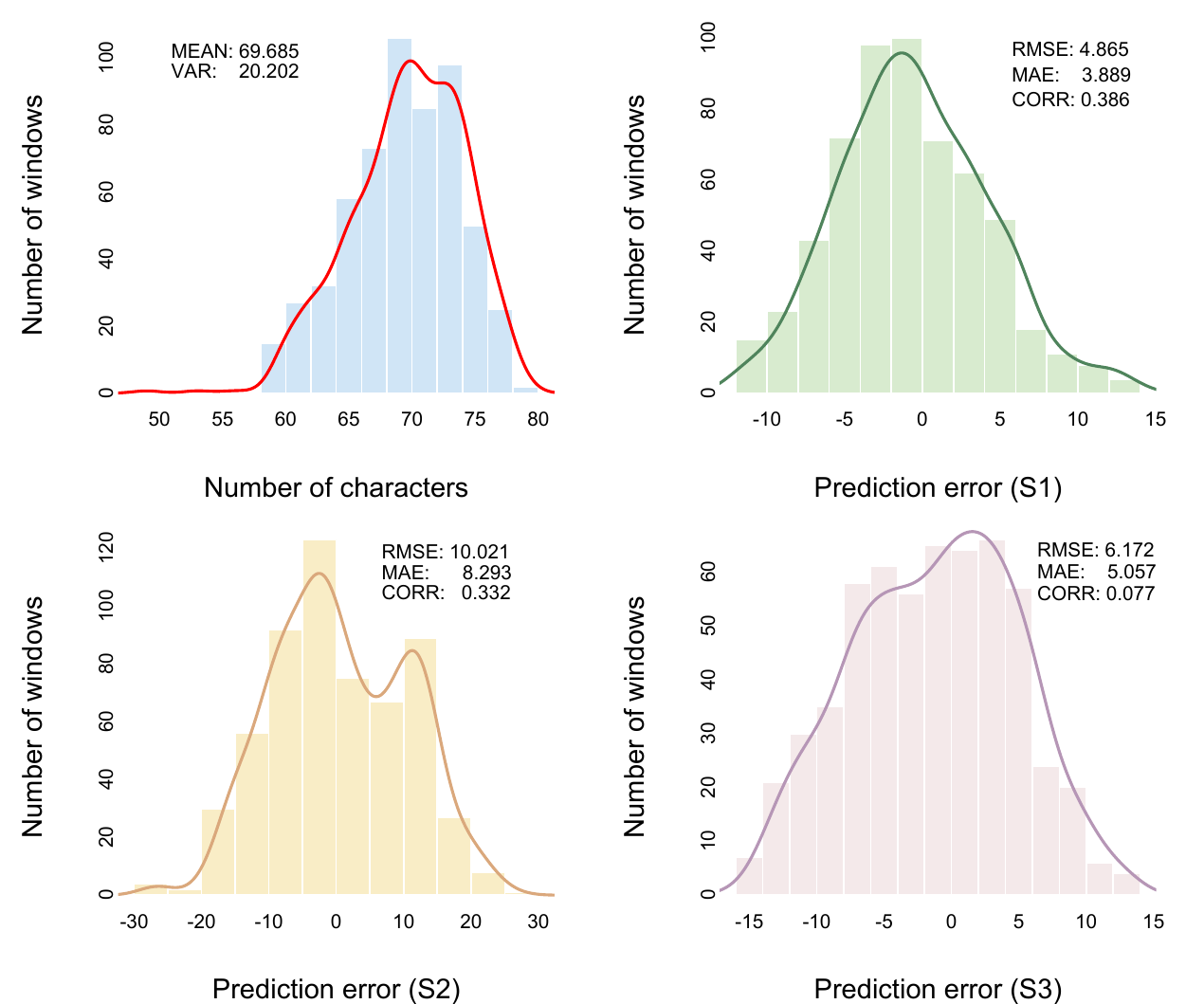}
        \label{fig:within-count}
    \end{subfigure}
    \begin{subfigure}[valign=t]{0.49\textwidth}
        \includegraphics[width=\textwidth]{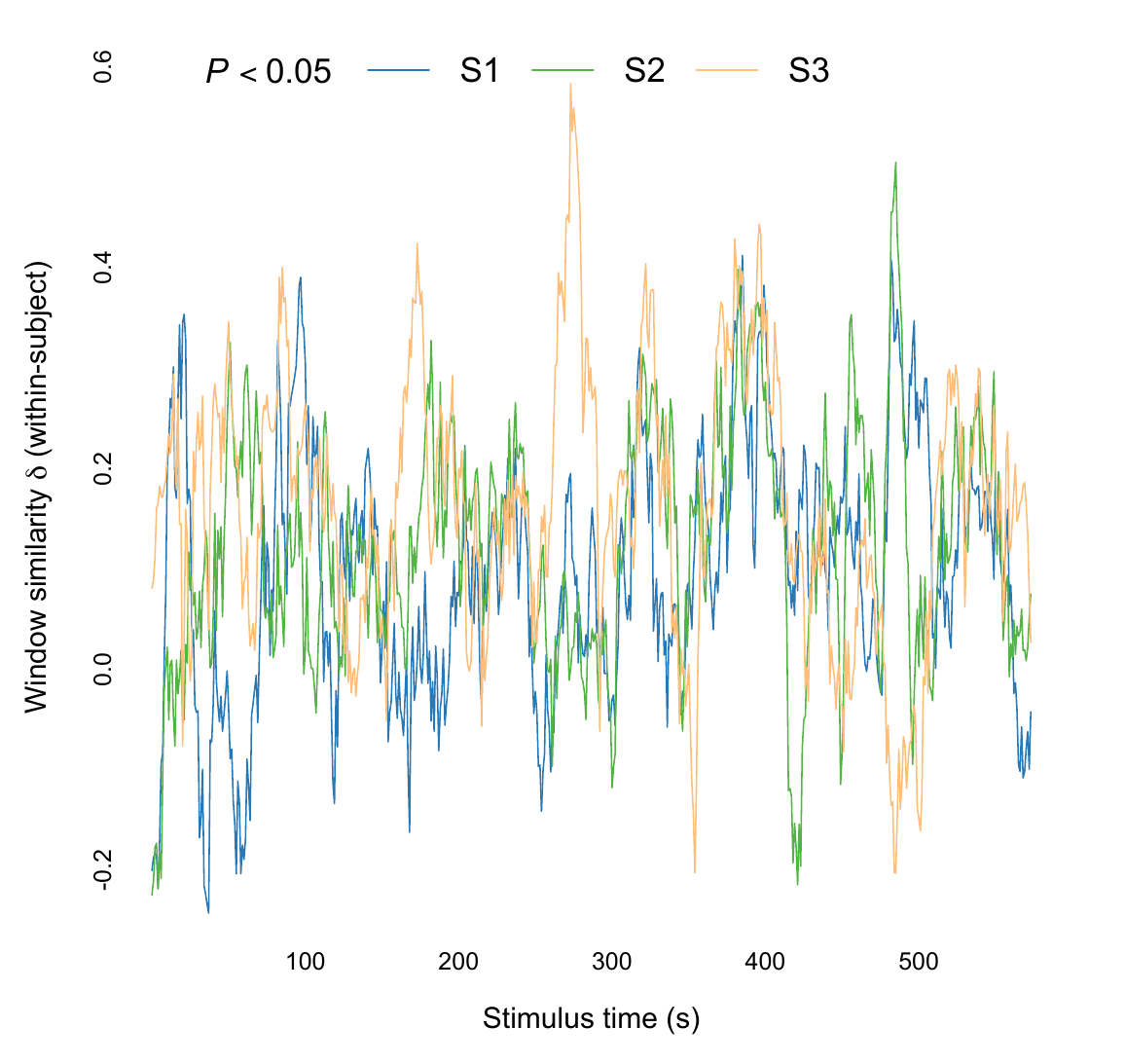}
        \label{fig:within-time-null}
    \end{subfigure}
    \begin{subfigure}[valign=t]{0.45\textwidth}
        \includegraphics[width=\textwidth]{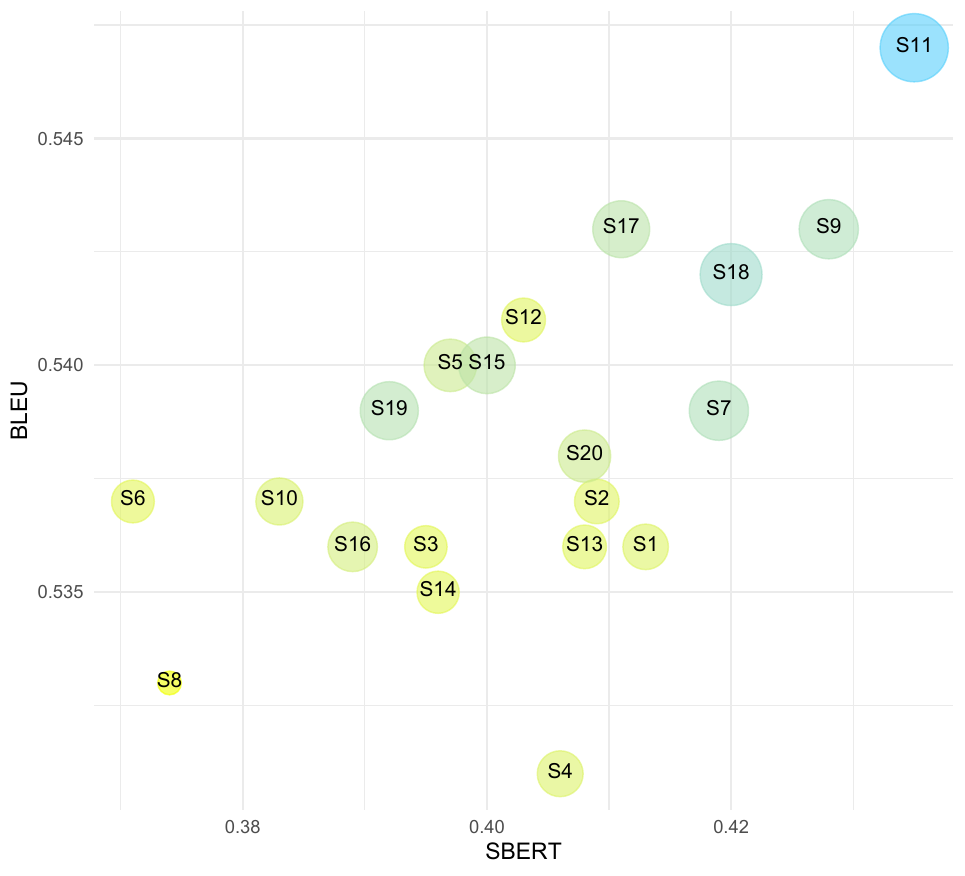}
        \label{fig:time-windows}
    \end{subfigure}

    \caption{\small Language similarity scores in a within-subject setting. \textbf{(a)} Decoder predictions for a test article (2,040 characters) exhibited significantly greater similarity to the actual stimulus character sequence than both the baseline and expected by chance ($P < 0.05$ for all subjects, one-sided non-parametric test) across all language similarity metrics. To compare across metrics, results are shown as deviations away from the mean of the null distribution (Methods).
    \textbf{(b)} The actual distribution of character rate for a test article (top-left corner) alongside three distributions of character rates predicted for different subjects. 
    Each subject's character rate model was independently trained and evaluated by predicting the character rate of a test article, and then assessing the linear correlation between the predicted and actual character rate distributions.
    Predicted distributions were significantly higher than expected by chance ($P < 0.05$, one-sided non-parametric test).
    \textbf{(c)} Decoding scores were significantly higher than expected by chance ($P < 0.05$, one-sided non-parametric test) for most timepoints under the SBERT metric. 
    \textbf{(d)} Circle size is proportional to the average window similarity between the feature vectors extracted by 3dC-IB from brain responses and the semantic representations of a test article. The degree of similarity closely aligns with both SBERT and BERT scores.
    } 
    \label{fig:within-analysis}
\end{figure}

\section*{Results}\label{sec:resutls}

\subsection*{Can character-based decoders reconstruct continuous language?}


We trained a decoder for each of $20$ subjects on a training set comprising fMRI brain recordings elicited by naturally spoken narratives over a duration of $151$ minutes, and evaluated each subject's decoder on novel, single-trial brain responses recorded while the subject listened to a test article ($11$ minutes in duration, containing $2,040$ characters) that was not present in the training set.
During the training phase, the neural network decoder learns to identify and interpret the semantic features from fMRI brain recordings which align with the stimulus character sequences. 
The decoder predictions should reflect the semantic content conveyed by the stimuli rather than low-level articulatory or auditory features.
Results show that the character sequences decoded from the brain activity not only captured the meaning of the presented stimuli but, in many instances, matched exact characters, words, or phrases (Fig. \ref{fig:decoding-results}), demonstrating that continuous language and semantic information can be reconstructed from  BOLD signals on a character-by-character basis (Fig. \ref{fig:within-analysis}a and Extended Data Fig. \ref{fig:within-all}).
Note that while the decoder reconstructs the stimulus texts in a character-by-character manner, the evaluation of candidate sequences was performed based on sequence-level similarity (Fig. \ref{fig:illustration}b) with the assistance of a generative language model.

\begin{table}[t]
  \setlength{\tabcolsep}{1.1mm}
  \caption{\label{tab:main_result_across} Language similarity scores in a cross-subject setting}
    \begin{tabular}{l|ccc|ccc|ccc|ccc}
      
    \hline
   \rowcolor{gray!20} \multirow{2}{*}{} & \multicolumn{3}{c|}{\textbf{BLEU}} & \multicolumn{3}{c|}{\textbf{METEOR}} &
    \multicolumn{3}{c|}{\textbf{BERT}} & \multicolumn{3}{c}{\textbf{SBERT}} \\
   \rowcolor{gray!20} & NULL & BL &  3dC-IB  & NULL & BL & 3dC-IB & NULL & BL &  3dC-IB & NULL & BL & 3dC-IB \\
    \hline
    
    S1 & 0.167 & 0.184 & \textbf{0.218} & 0.096 & 0.108 & \textbf{0.119} & 0.519 & 0.532 & \textbf{0.533} & 0.268 & 0.365 & \textbf{0.418} \\
    
    \rowcolor{gray!20} S2 & 0.167 & 0.162 & \textbf{0.246} & 0.099 & 0.094 & \textbf{0.144} & 0.520 & 0.526 & \textbf{0.545} & 0.274 & 0.325 & \textbf{0.422} \\
    
    S3 & 0.170 & 0.167 & \textbf{0.224} & 0.099 & 0.098 & \textbf{0.127} & 0.519 & 0.523 & \textbf{0.541} & 0.263 & 0.308 & \textbf{0.387} \\
    
    \rowcolor{gray!20} S4 & 0.172 & 0.163 & \textbf{0.228} & 0.100 & 0.103 & \textbf{0.136} & 0.522 & 0.529 & \textbf{0.550} & 0.279 & 0.300 & \textbf{0.427} \\
    
    S5 & 0.172 & 0.169 & \textbf{0.215} & 0.099 & 0.098 & \textbf{0.120} & 0.519 & 0.523 & \textbf{0.535} & 0.287 & 0.276 & \textbf{0.376} \\
    
    \rowcolor{gray!20} S6 & 0.160 & 0.170 & \textbf{0.227} & 0.100 & 0.101 & \textbf{0.130} & 0.521 & 0.523 & \textbf{0.544} & 0.287 & 0.328 & \textbf{0.434} \\
    
    \hline
    \end{tabular}
    {Decoder predictions for a test article were compared to the actual stimulus textual sequence of $2,040$ characters using a range of language similarity metrics: BLEU, METEOR, BERT, and SBERT, in a cross-subject setting. To establish a null distribution, the mean similarity between the actual stimulus characters and $200$ null sequences generated by a large language model was computed with brain data only used to estimate the number of perceived characters. The abbreviation ``BL'' denotes the baseline model \cite{tang2023semantic} applying a regularized linear model for decoding.
    Remarkably, the baseline model exhibits inferior performance compared to the null distribution on certain metrics and subjects, while the 3D convolutional network with an information bottleneck (3dC-IB) consistently outperformed both the baseline and null distributions across all subjects and metrics.}

\end{table}

To qualify the decoding performance, we compared decoded and actual character sequences for a test article ($2,040$ characters) using a range of language similarity metrics (Methods). 
Traditional metrics, such as BLEU and METEOR, were used to measure the overlap of characters or words between two textual sequences, a standard method commonly applied in machine translation system evaluations.
Recognizing that different characters can convey the same or similar meaning, the BERT score was used to qualify the semantic similarity between two sequences.
The above three metrics still focus on character-based evaluations. 
Given the importance of sequential order and compositional nuances in natural language, it is crucial to consider how characters and words are organized.
To address this, we also used SBERT, a newer method that accounts for the sequential arrangement of characters and the variability in expressing same meanings. 
This metric enhances our ability to measure the similarity of textual sequences by considering both the order of textual elements and their compositionally, thereby offering a more comprehensive evaluation of decoding performance.
The decoding performance achieved by the 3dC-IB model was significantly higher than both the baseline and what would be expected by chance ($P < 0.05$, one-sided non-parametric test; Fig. \ref{fig:within-analysis}a and Extended Data Fig. \ref{fig:within-all}; see Table \ref{tab:main_result_within} for raw scores).
A majority of timepoints in the article---over $80\%$---had a significantly higher SBERT than expected by chance based on the similarities evaluated between the decoded and actual characters (Fig. \ref{fig:within-analysis}c).
We also evaluated the character rate model's performance by examining the linear correlation between the predicted and actual character rate distributions.
A separate character rate model was constructed for each individual subject.
In general, there is a strong correlation between the predicted rate distributions and the actual character rate distribution (Fig. \ref{fig:within-analysis}b).
It is noteworthy that the generative language model can express the same or similar meanings using varying numbers of characters or words. 
Consequently, the differences between the predicted and actual character counts have a limited impact on decoding performance  if such differences are not substantially large. 
These results indicate a notable accuracy in the model's ability to recover the stimulus character sequences from non-invasive BOLD signals.

\begin{figure}
    \centering
    \begin{subfigure}[valign=t]{0.49\textwidth}
        \includegraphics[width=\textwidth]{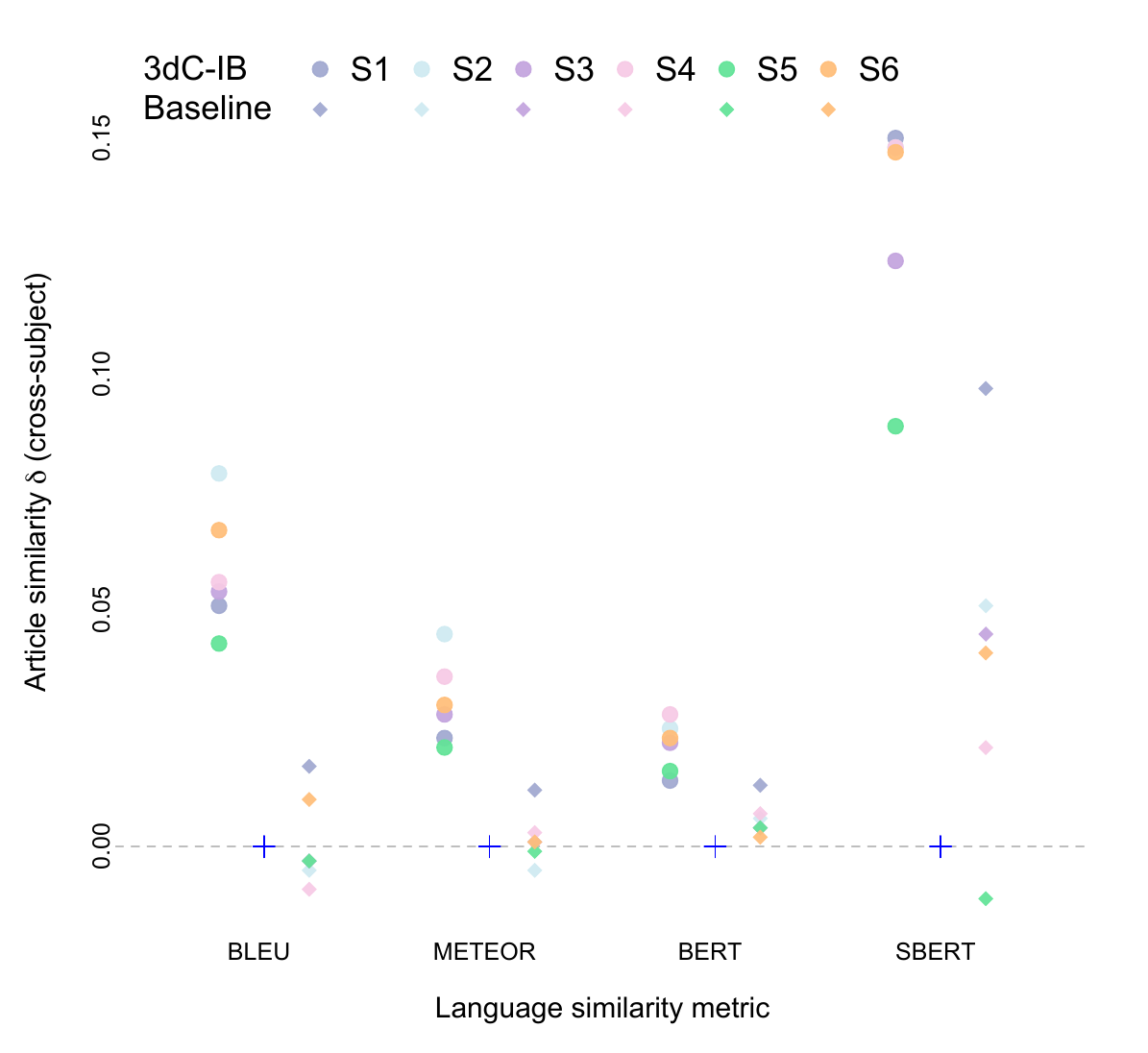}
        \label{fig:across-subject}
    \end{subfigure}
    \begin{subfigure}[valign=t]{0.49\textwidth}
        \includegraphics[width=\textwidth]{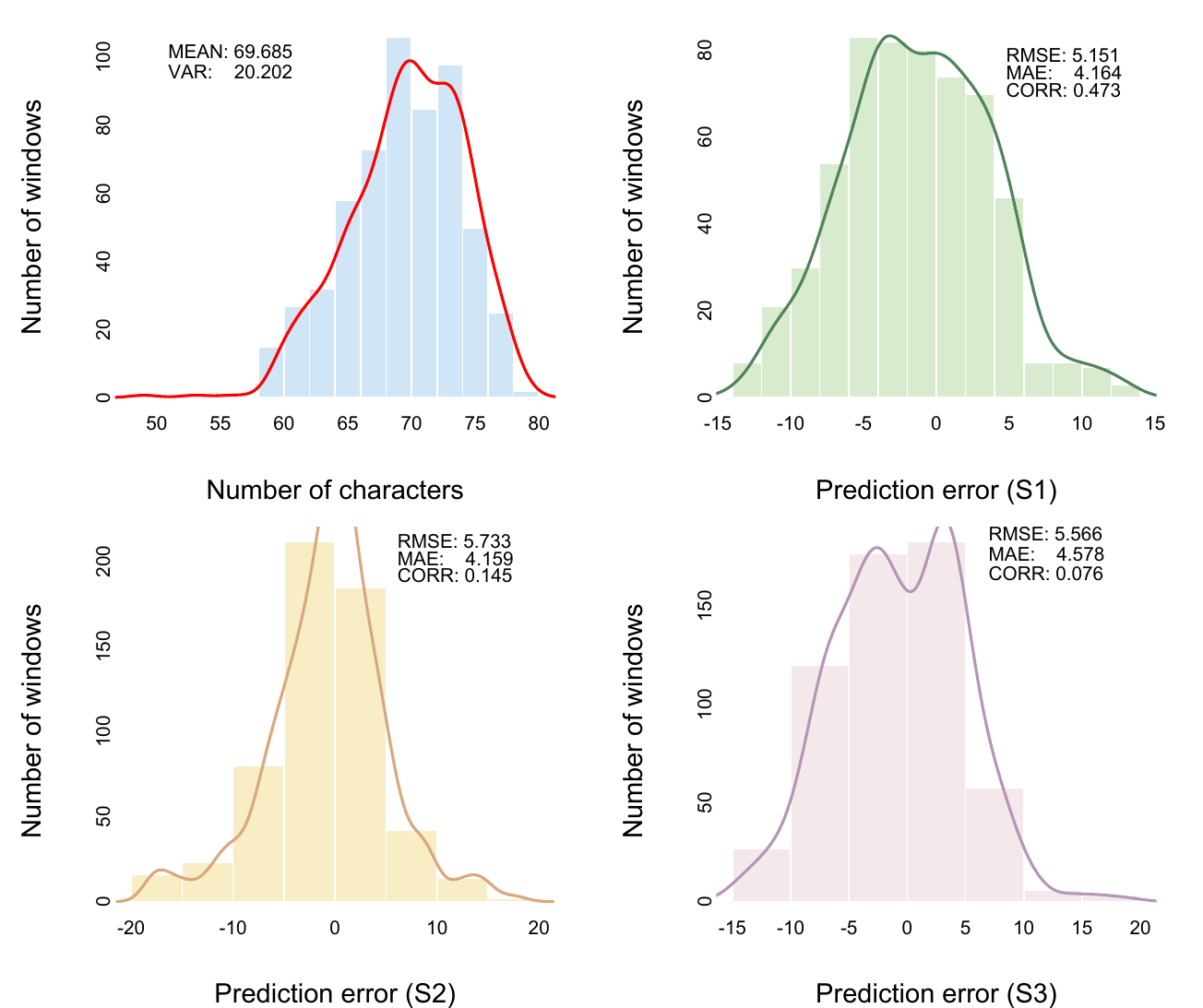}
        \label{fig:across-count}
    \end{subfigure}
    \begin{subfigure}[valign=t]{0.49\textwidth}
        \includegraphics[width=\textwidth]{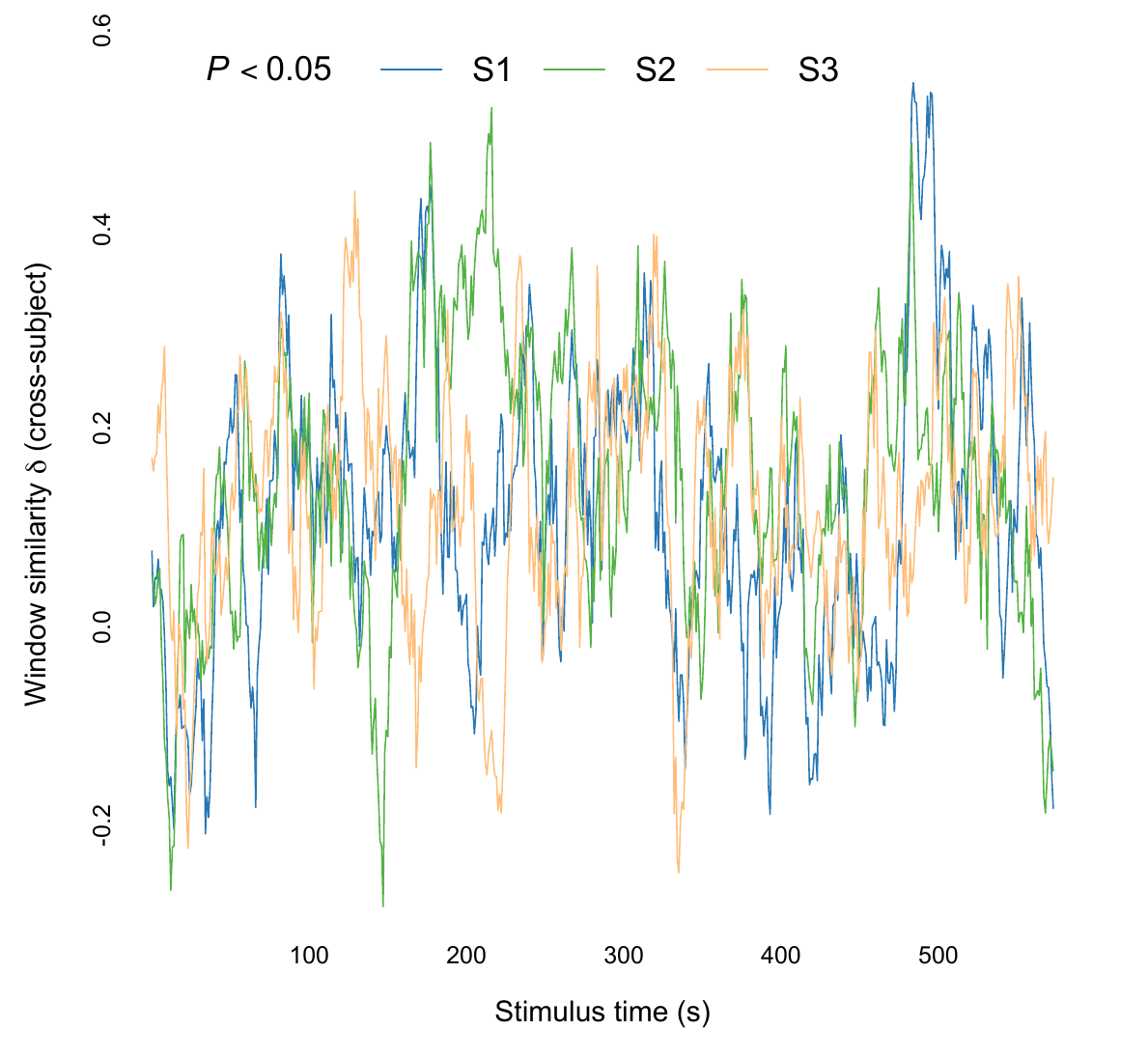}
        \label{fig:across-time-null}
    \end{subfigure}
    \begin{subfigure}[valign=t]{0.49\textwidth}
        \includegraphics[width=\textwidth]{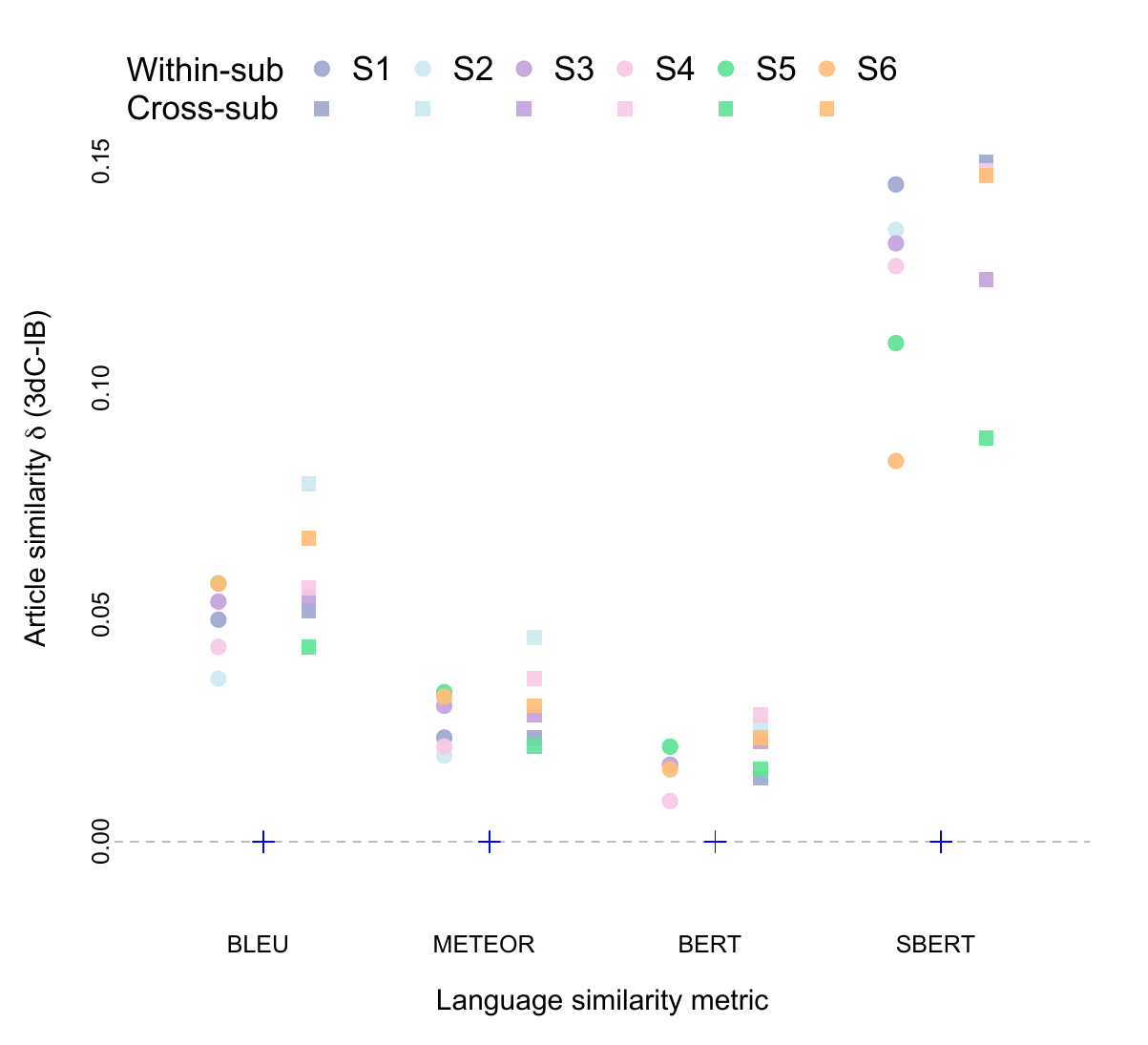}
        \label{fig:within-across}
    \end{subfigure}
    \caption {\small Language similarity scores in a cross-subject setting. \textbf{(a)} Decoder predictions for a test article (2,040 characters) were significantly more similar to the actual stimulus character sequence than both the baseline and expected by chance ($P < 0.05$ for all subjects, one-sided non-parametric test) across all language similarity metrics. To compare across metrics, results are shown as deviations away from the mean of the null distribution (Methods).
    \textbf{(b)} The actual distribution of character rate for a test article (top-left corner) alongside three predicted distributions of character rates. Each subject's character rate model was independently trained and evaluated by predicting the character rate of a test article and assessing the linear correlation between the predicted and the actual character rate distributions. Predicted distributions were significantly higher than expected by chance ($P < 0.05$, one-sided non-parametric test). Cross-subject character rate models performed slightly worse than within-subject character rate models.
    \textbf{(c)} Decoding scores were significantly higher than expected by chance ($P < 0.05$, one-sided non-parametric test) for most timepoints under the SBERT metric.
    \textbf{(d)} Employing the 3dC-IB model, decoder predictions for a test article in a cross-subject setting were significantly more similar to the actual stimulus character sequence than expected by chance ($P < 0.05$ for all subjects, one-sided non-parametric test), demonstrating a performance comparable to that observed in a within-subject setting.}
    \label{fig:across-analysis}
\end{figure}

\subsection*{Are cross-subject models applicable for language decoding?}

A key task for brain-computer interfaces is decoding language or speech from brain recordings of novel, unseen subjects. 
In this cross-subject setting, a decoder is trained on brain responses collected from a set of subjects and then applied to a separate subject who was not included in the training dataset. 
Limited research has ventured into decoding natural language from non-invasive recordings across different subjects \cite{xi2023unicorn, defossez2023decoding}. 
These initiatives have primarily concentrated on identifying words or fragments within a predetermined set of possibilities, leaving questions about the applicability of these approaches for continuous language decoding, which typically involves processing long sequences of over a thousand characters or words.
A notable advancement has shown that it is possible to decode continuous language from the non-invasive recordings of English speakers \cite{fedorenko2010new}. 
However, their cross-subject encoding models considerably lagged behind within-subject models and did not significantly outperform null models. 
Therefore, it remains a major challenge to decipher continuous language from non-invasive brain recordings in a cross-subject setting.

In this set of experiments, we strictly adhered a cross-subject validation protocol. 
A 3dC-IB model was trained on data derived from a subset of subjects and subsequently evaluated on a held-out subject excluded from the training dataset. 
Under this cross-subject setting, the 3dC-IB model exhibited decoding accuracy that was significantly higher than that of the baseline and what would be expected by chance ($P < 0.05$, one-sided non-parametric test; Fig. \ref{fig:across-analysis}a and Extended Data Fig. \ref{fig:across-all}; see Table \ref{tab:main_result_across} for raw scores).
In more than $80\%$ of the timepoints within the test article, the achieved SBERT similarity scores between decoded and actual character sequences were significantly higher than expected by chance (Fig. \ref{fig:across-analysis}c).
Empirical results showed that the cross-subject character rate models performed slightly worse than the within-subject character rate models (Fig. \ref{fig:within-analysis}b and \ref{fig:across-analysis}b).  
However, the performance drop in the character rate models has a modest impact on decoding accuracy as the employed language model can generate character sequences with varying character counts while conveying equivalent or similar meanings.
Surprisingly, the cross-subject models sometimes outperformed their within-subject counterparts, as observed in several instances: subject 1 (S1) under the BLEU and SBERT metrics, subject 2 (S2) across BLEU, BERT, and SBERT metrics, and subject 6 (S6) in terms of BLEU, BERT, and SBERT (Tables \ref{tab:main_result_within} and \ref{tab:main_result_across}).
These results indicate the 3dC-IB's capability to accurately reconstruct the character sequences of the stimulus from non-invasive BOLD signals, highlighting its potential for effective cross-subject decoding in brain-computer interface applications.
In comparison, the linear baseline model \cite{tang2023semantic} did not exhibit a significant improvement over the null models, and for certain subjects, it even performed inferior to the null models (Fig. \ref{fig:across-analysis}a and Extended Data Fig. \ref{fig:across-all}). 
This outcome aligns with the findings presented in \cite{tang2023semantic}, indicating that the linear models' inadequacy in capturing the semantic features encoded in the fMRI recordings across different subjects. 
In contrast, deep neural networks enhanced with a denoising component (i.e., information bottleneck) are able to capture high-level, invariant semantic features shared by different subjects through multiple successive layers of transformation.
Arguably, this demonstrates that there exists a discernible degree of similarity in how identical or similar meanings are represented in the brain activity of different individuals.
This commonality in cognitive processing can be exploited to facilitate language decoding across subjects, revealing the potential of advanced neural network architectures in bridging individual disparities in brain activity patterns for semantic interpretation.

\subsection*{Identifying cortical regions engaged in semantic processing}

\begin{figure}
    \centering
    \begin{subfigure}[valign=t]{0.49\textwidth}
        \includegraphics[width=\textwidth]{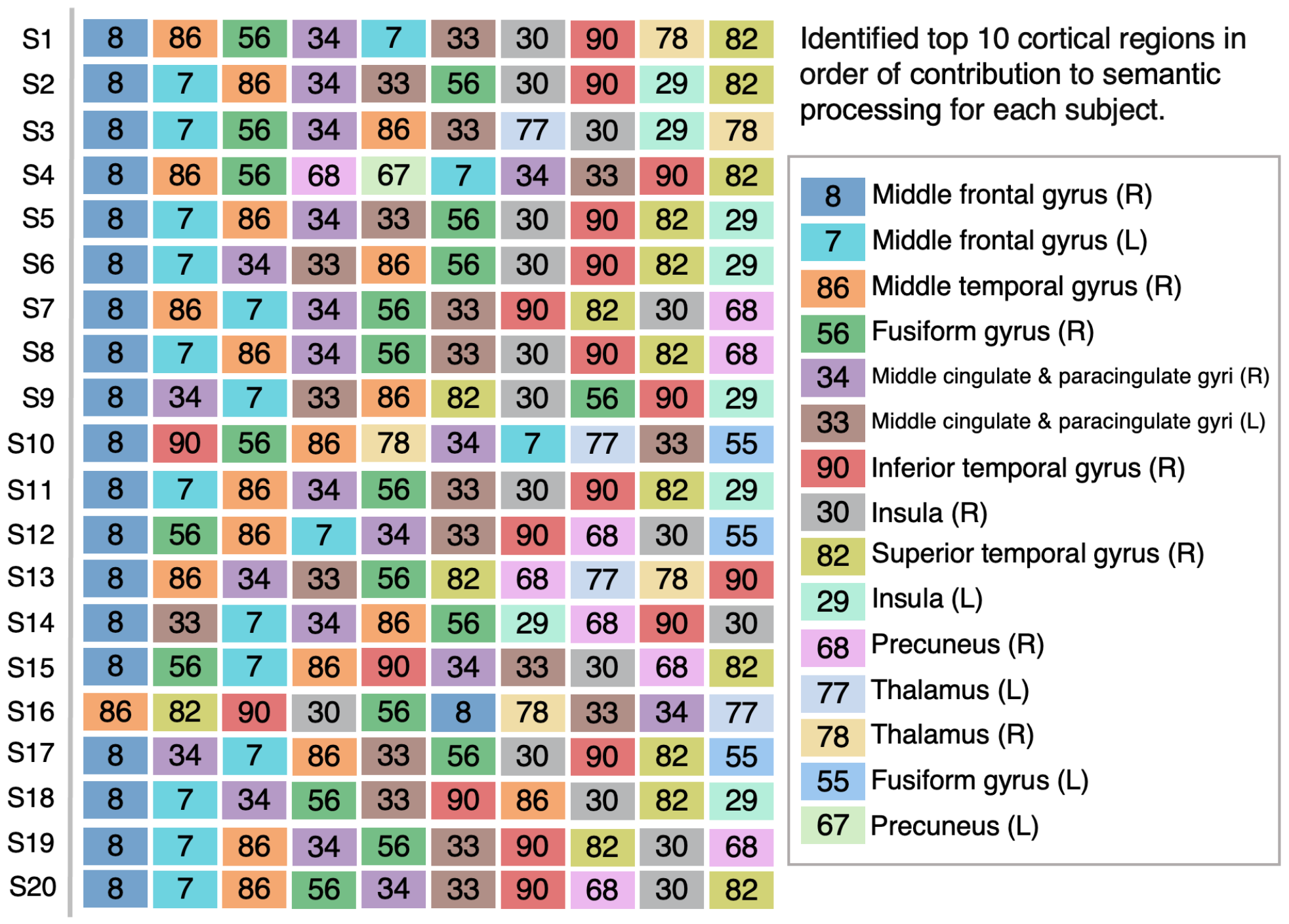}
        \label{fig:region}
    \end{subfigure}
    \begin{subfigure}[valign=t]{0.42\textwidth}
        \includegraphics[width=\textwidth]{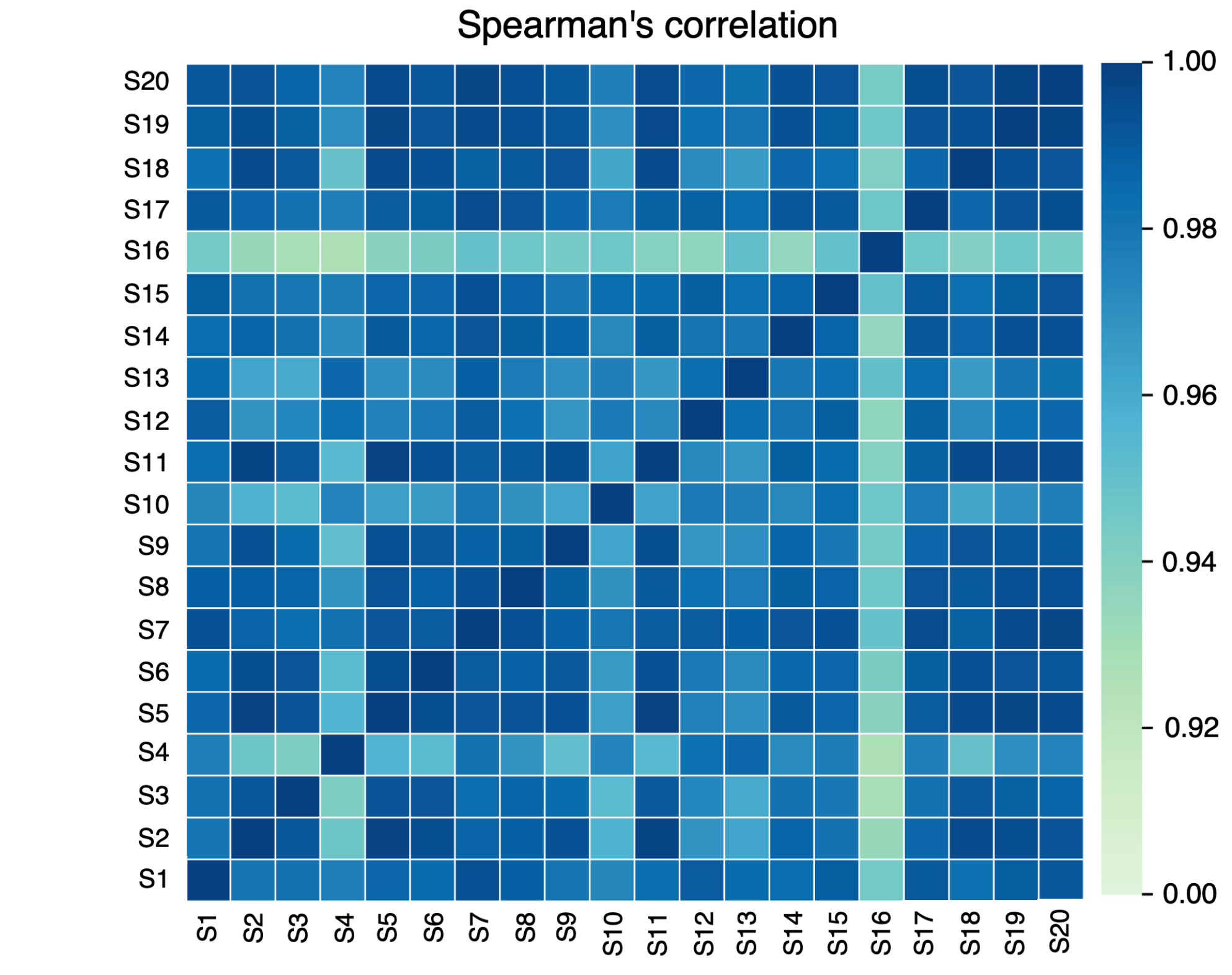}
        \label{fig:correlation}
    \end{subfigure}
    \begin{subfigure}[valign=t]{0.49\textwidth}
        \includegraphics[width=\textwidth]{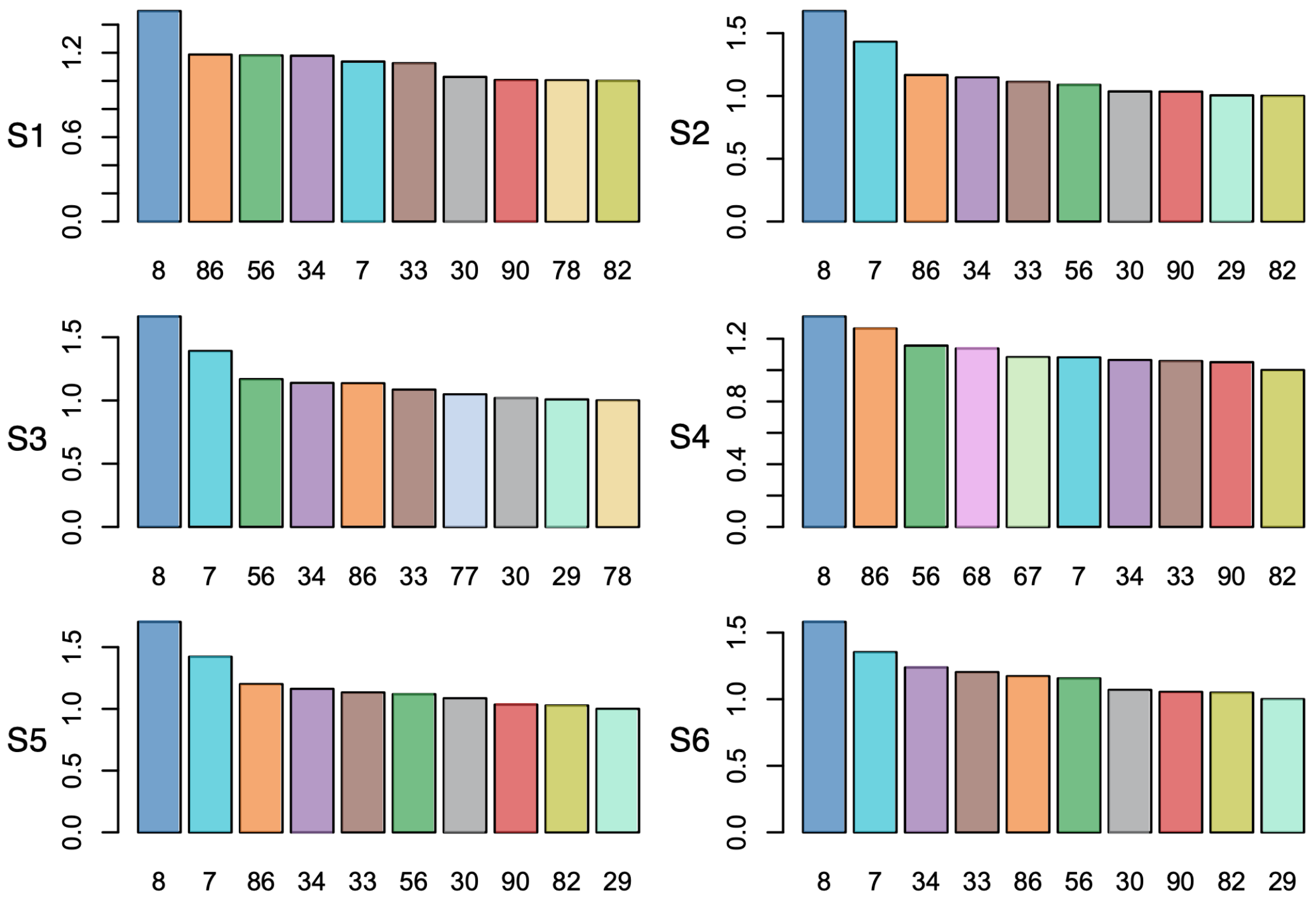}
        \label{fig:region-barplot}
    \end{subfigure}
    \begin{subfigure}[valign=t]{0.42\textwidth}
        \includegraphics[width=\textwidth]{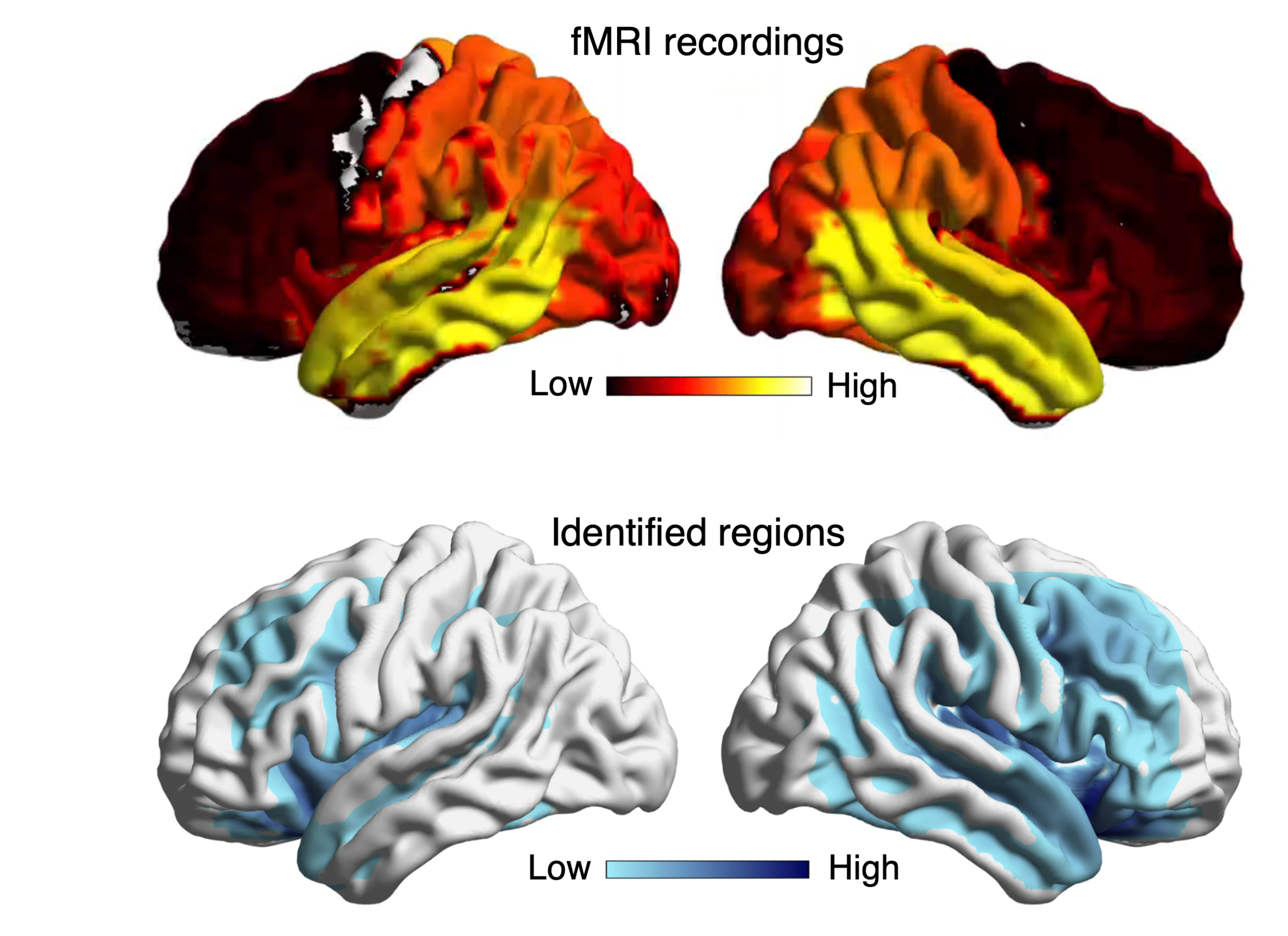}
        \label{fig:visualization}
    \end{subfigure}

    \caption {\small Identified cortical regions. \textbf{(a)} The identified $10$ most significantly contributing cortical regions engaged in language and semantic processing for each subject, with the numeric codes indicating cortical regions defined in the Automated Anatomical Labeling (AAL) template.  
    \textbf{(b)} Spearman’s correlation coefficients calculated for the brain contribution patterns between each pair of subjects, which reveals a notable uniformity in the patterns of brain contribution across subjects.
    \textbf{(c)} The top $10$ identified cortical regions and their relative importance for six subjects. The contribution value of the least contributing cortical region was used as a reference point and set to one unit.
    \textbf{(d)} A typical fMRI recording, as well as the identified regions highlighted using varying shades of color based on their contributions.}
    \label{fig:cortical-region}
\end{figure}

Employing an end-to-end neural network architecture for the decoder allows us to identify more responsive voxels involved in language and semantic processing when subjects are exposed to natural speech narratives. 
These deep network solutions directly map the input data to the desired output without relying on handcrafted feature engineering or intermediate processing steps.
Consequently, the entire process, from input to output (i.e., from BOLD signals to predictions), is encapsulated within the network model itself. 
By leveraging such end-to-end architectures, we are able to compute gradients of decoder predictions with respect to input voxels on the training examples and aggregate the absolute gradients across all voxels within a cortical region. 
The magnitude of summed gradients for each region can be interpreted as the significance of the contribution to semantic processing elicited by speech stimuli, as gradients reflect the extent to which predictions are influenced by changes in input magnitude.

Through the aforementioned process, we identified the $10$ most significantly contributing cortical regions and charted their relative importance for each subject, as depicted in Fig. \ref{fig:cortical-region}a and \ref{fig:cortical-region}c.
The Automated Anatomical Labeling (AAL) template was used as the anatomical parcellation scheme, which provides a standardized method for subdividing the brain into anatomically defined regions, each associated with a specific anatomical label and code (ranging from 1 to 90).
Remarkably, the middle frontal gyrus (MFG) emerged as the most prominently engaged area, showing consistent activation during language and semantic processing across all subjects in this study.
This observation reinforces existing literature that positions the MFG as a pivotal hub for Chinese reading \cite{zhao2017rethinking,wu2012meta,price2012review,guo2022brain} and semantic representations \cite{booth2006specialization}.
Specifically, the left hemisphere's MFG is  closely associated with the orthographic processing required for lexical decision-making tasks, reflecting the unique structure of Chinese characters that incorporate strokes and radicals into a square form to represent syllables and morphemes of the spoken language \cite{coulmas2003writing}.
Alongside the MFG, several cortical regions were identified as critical for various aspects of Chinese language processing.
For example, the middle temporal gyrus was found to be actively involved in syntactic analysis \cite{sun2021independent}, while the fusiform gyrus was associated with orthographic representations for Chinese \cite{booth2006specialization}.
The right hemisphere's inferior temporal gyrus was identified as a critical cortical area engaged in orthographic processing \cite{cao2010cultural}, and the superior temporal gyrus, in conjunction with other cortical regions, was found to be implicated in orthographic processing, phonological identification of word form, and semantic processing \cite{bolger2005cross,zhang2013contribution}.

To assess the similarity of identified cortical regions across different subjects, we assigned each voxel a value corresponding to the aggregated absolute gradients computed over all training examples for each subject.
These values were then organized into a vector, with their positions mirroring those in the original fMRI recordings.
Subsequently, we calculated the Spearman's correlation for each pair of these vectors. 
The Spearman's correlation is, by definition, the Pearson correlation applied to the rank variables, which represent the relative ordering of the observations within each variable.
As illustrated in Fig. \ref{fig:cortical-region}b, a notable uniformity in the patterns of brain contribution was observed across subjects, even though the cortical regions were independently identified for each subject.
While Subject 16 (S16) exhibited a somewhat lower similarity with the others, the Spearman's correlation coefficients between S16 and any other subject remained high (all above $0.85$), and the cortical regions identified for S16 exhibited substantial overlap with those identified for the others.
The ability to identify plausible cortical regions and the high degree of similarity in brain contribution patterns found across subjects
suggest the potential of well-designed deep neural network architectures to provide valuable insights for exploring neural mechanisms.

\section*{Discussion}\label{sec:discussion}

This study demonstrates the feasibility of decoding the meaning of speech stimuli from BOLD signals into a continuous linguistic sequence on a character-by-character basis, taking a further step towards non-invasive brain-computer interfaces for inherently character-based languages.
The use of character-based decoders is particularly advantageous as it allows for a substantially reduced vocabulary size without sacrificing decoding accuracy, thanks to the integration of advanced large language models that ensure the semantic coherence in the decoded texts.
For instance, while there are approximately $3,500$ commonly used Chinese characters, the total number of Chinese words at least exceeds $55,000$.
Given the same amount of training data, the requirement for a smaller vocabulary means that models need to learn fewer patterns, leading to improved decoding accuracy compared to the models burdened with a larger vocabulary.
This strategy is also applicable to word-based languages such as English and French by segmenting words into smaller units (sub-words or tokens), which had been widely used in prominent language models like BERT \cite{vaswani2017attention,devlin2019bert} and GPT-4 \cite{OpenAI2023GPT4TR}.
Moreover, the adoption of a reduced vocabulary could enable the models to process multiple languages or to decode language from multilingual speakers, broadening the scope of application for brain-computer interfaces in multilingual contexts.

BOLD fMRI recordings typically exhibit a low signal-to-noise ratio (SNR). 
These signals also present substantial variability across trials and subjects, making them susceptible to contamination by many artifacts  \cite{hamalainen1993magnetoencephalography,schirrmeister2017deep,king2020encoding}. 
To increase the SNR, exiting approaches either use the average of non-invasive brain responses obtained during different repeats of the same speech stimuli as inputs \cite{tang2023semantic}, or adopt complex prepressing pipelines that involve a variety of techniques including independent component analysis, outlier detection, and artefact correction \cite{haxby2001distributed,kamitani2005decoding,nishimoto2011reconstructing}.
Following these preprocessing steps, it is common practice to apply a linear model specifically trained for each subject.
In contrast, our study explores the use of end-to-end deep network architecture for language decoding, which obviates the need for such complex and troublesome preprocessing steps. 
Particularly, we incorporate the information bottleneck (IB) method as a unique layer within a 3D convolutional neural network framework.
This approach aims to optimize the balance between accuracy and complexity during the compression of brain image inputs. 
It facilitates the identification of voxels with higher responsiveness, thereby improving the SNR of brain activity measurements.

Previous studies have shown the challenges associated with using linear models to effectively decode continuous language from BOLD fMRI recordings in a cross-subject context, and it has been found that cross-subject models exhibit significantly inferior performance compared to their within-subject counterparts \cite{tang2023semantic}.
While deep neural network architectures have been investigated for their potential in analyzing brain signals, these initiatives have primarily concentrated on straightforward classification tasks \cite{defossez2023decoding} or have been limited to analyses involving models developed based on data collected from individual subjects \cite{roy2019deep,dash2018determining}.
In this study, we explored the feasibility of deep neural networks to decode continuous language across subjects. 
A 3D convolutional neural network, augmented with an information bottleneck, was developed to identify invariant semantic features present in BOLD fMRI recordings across various subjects and trials. 
The experimental results demonstrated the ability of this neural architecture to identify the semantic features shared by various subjects, and the developed models in this study achieved performance on par with, and in certain instances, even superior to their within-subject counterparts.
These findings open up the possibility for the application of decoders trained on a separate set of subjects to decode language for unseen subjects.

Contrary to the majority of existing language decoders which map brain activity to explicit motor features \cite{anumanchipalli2019speech} or utilize data from regions associated with motor representations during overt or attempted language production \cite{willett2021high}, our decoder is distinguished by its focus on semantic features \cite{huth2016natural} and predominantly relies on data from regions known for semantic encoding \cite{pasley2012reconstructing,tang2023semantic}.
Furthermore, the end-to-end methodology applied in this study enables the identification of cortical areas that are primarily engaged or activated during semantic processing.
This is achieved by backpropagating errors through the neural networks from the predictions back to the input voxels, thereby estimating the contribution of each voxel to the semantic tasks.
Through this process, we have pinpointed the top-$10$ cortical regions most critically involved in semantic representation, discovering a high degree of consistency in the identified regions and their relative importance across different subjects. Most notably, the bilateral middle frontal gyrus (MFG) emerged as the most active region in semantic processing, confirming its pivotal role in Chinese language processing.

Prior research on the neural basis of language processing has indicated that while many aspects of language processing are universal across languages, there are also language-specific differences that reflect the unique linguistic and cognitive demands of different languages. 
For example, Chinese is a logographic language, where each character represents a meaningful unit. 
The cognitive processing of these characters involves visuospatial analysis and memory areas where the MFG is actively engaged \cite{kuo2004orthographic,tan2005neuroanatomical,booth2006specialization}.
Previous studies also suggested that the MFG's involvement in visuospatial processing might be more pronounced in Chinese due to the logographic nature of its writing system \cite{liu2006dissociated,liu2022functional,wu2012meta,guo2022brain}, whereas its role in phonological and syntactic processing might be more emphasized in English.
These observations confirmed the effectiveness of our approach in tracking the voxel-wise contributions to language production and semantic processing, which also enhances our comprehension of the underlying decoding mechanisms.
Furthermore, the pattern of brain contributions observed in this study paves new avenues for using deep learning models as a tool for exploring neural mechanisms.

In this study, the decoder was primarily designed to recover the semantic meaning of perceived speech stimuli, which may result in the reconstructed language that paraphrases the actual stimuli using different character sequences. 
To improve the fidelity of reconstructing the actual stimuli, future developments could benefit from integrating motor features or additional complementary recordings, such as magnetoencephalography (MEG) or electroencephalography (EEG). 
These motor features or recordings are better at distinguishing between the actual stimuli and their paraphrases since they are more closely related to the surface form of speech stimuli. 
Incorporating these features also could provide users with increased control over the decoder's output, as they are less prone to interference with semantic processes such as perception and memory.
Moreover, leveraging the complementary strengths of fMRI's higher spatial but lower temporal resolution with EEG or MEG's higher temporal but lower spatial resolution could produce a complementary effect, further improving the decoding performance in a non-invasive manner.

Another important factor that could improve decoding performance is the expansion of training data size. 
In this study, we collected fMRI data from $20$ participants, each exposed to $2.7$ hours of naturally spoken narratives. 
These narratives originated from $20$ different articles (Tables \ref{tb:summary-1} to \ref{tb:summary-4}), with durations spanning from $8$ to $12$ minutes each. 
Diverging from earlier studies \cite{tang2023semantic,defossez2023decoding,xi2023unicorn}, where training and testing stories typically originated from homogeneous genres, our collection of $20$ articles spanned a diverse range of genres, including fairy tales, fables, novels, essays, news reports, and scientific articles. 
The relatively limited amount of fMRI data and the broad genre diversity of the stimulus articles might have a notable impact on decoding performance. 
Moving forward, we plan to substantially increase the amount of fMRI training data and to see how far we can go in the cross-subject setting. 
Furthermore, it would be intriguing to evaluate the performance of our neural language decoder for multilingual speakers or in multilingual contexts.


\section*{Methods}\label{sec:methods}

We use a three-dimensional convolutional neural network (3D CNN) \cite{ji20123d,myronenko20193d} to extract features from BOLD signals obtained through fMRI, incorporating an information bottleneck (IB) for refinement. 
These extracted feature vectors were trained to align semantically with the corresponding semantic feature representations of stimulus character sequences produced by a pre-trained BERT \cite{devlin2019bert}.
After training, the learned feature vectors can be used to estimate how likely the BOLD signals are elicited by candidate character sequences proposed by a pre-trained language model.
The IB was introduced to locate relevant voxels and improve signal-to-noise ratios in brain activity by maximizing the mutual information between the extracted features and stimuli while simultaneously minimizing the mutual information between these features and all BOLD signals, which inherently contain a significant amount of noise.

\subsection*{Model}

\subsubsection*{Encoding model}

We perform 3D convolutions to extract features from an fMRI brain image, which is achieved by convolving 3D kernels to the cubes formed by multiple contiguous voxels in spatial dimensions. 
By this convolution, the value at $(x, y, z)$ on the $j$th feature map in the $i$th layer is computed by:

\begin{equation} \label{eq:convolution}
v^{xyz}_{ij} = \text{ReLU}(\sum^{K_i - 1}_{k = 0}\sum^{Q_i - 1}_{q = 0}\sum^{R_i - 1}_{r = 0}\sum^{S_i - 1}_{s = 0} w^{qrs}_{ijk} \times v_{(i - 1)k}^{(x+q)(y+r)(z+s)})
\end{equation}

\noindent where $\text{ReLU}(x) = \max(0, x)$ is a rectifier activation function \cite{fukushima1969visual,hahnloser2000digital,hahnloser2000permitted}, $K_i$ denotes the number of feature maps in the $(i -1)$th layer, and $w^{qrs}_{ijk}$ denotes the weight at the position $(q,r,s)$ of the kernel connected to the $k$th feature map in the previous layer.
$Q_i$, $R_i$, and $S_i$ are the length, height, and width of the kernel, respectively.
It is noteworthy that the bias term, typically present in Equation \eqref{eq:convolution}, is intentionally removed as it ensures that the result of $v^{xyz}_{ij}$ is zero when all input values are zero, indicating the absence of BOLD signals in those voxels.
Following the 3D convolution, 
subsampling operations are performed to reduce the resolution of feature maps by applying max-pooling over local neighborhoods on the feature maps in the previous layer, thereby enhancing invariance to distortions in brain images.
A 3D CNN architecture can be built by stacking multiple layers of convolution and subsampling in an alternating fashion. 
The parameters of 3D CNN (i.e., the kernel weights) are learned using a gradient descent algorithm \cite{rumelhart1986learning,lecun1998gradient} by backpropagating errors through the network.

We developed a 3D CNN architecture with six layers (Fig. \ref{fig:illustration}), consisting of three convolutional layers and three subsampling layers, designed specifically for extracting features from fMRI images. 
The size of input brain images is $53 \times 63 \times 52$, and 3D convolutions with a kernel size $7 \times 7 \times 7$ are first applied on input with $32$ filters.
Subsequently, we apply $2 \times 2 \times 2$ max-pooling on each of the feature maps produced by the first convolutional layer.
The subsequent two 3D convolutional layers use the same kernel size of $7 \times 7 \times 7$ with $128$ and $1,000$ filters, respectively.
Following a fundamental design principle of CNNs, the number of feature maps (or filters) was increased gradually in late layers, thereby facilitating the generation of multiple higher-level features from the same set of lower-level feature maps.
An adaptive max-pooling operation is applied on the last subsampling layer which computes the maximum activations of the units in each of the $1,000$ feature maps.
Finally, a linear layer is employed to reduce dimensionality with a set of $1,000 \times 768$ weights.
Through the sequential application of multiple convolutional, subsampling, and linear layers, an input brain image can be converted into a compact feature vector with $768$ dimensions. 

The semantic feature representation of the stimulus character sequence for each brain image was constructed by a three-step procedure. 
First, for each character-time pair $(c_i, t_i)$ within each training text where the time point $t_i$ is the midpoint of the pronunciation of character $c_i$, we fed the character sequences $(c_{i-5}, \dots, c_{i-1}, c_{i})$ into the BERT (twelve-layer base version) pre-trained for the Chinese language, and extracted the contextual feature representation of $c_i$ from the ninth layer of the BERT. 
Previous studies have shown that contextual feature representations generated by pre-trained artificial neural networks exhibit a robust alignment with observed neural responses, including the response properties of neurons in cortical regions such as the posterior parietal cortex \cite{zipser1988back} and the primary motor cortex \cite{lillicrap2013preference}.
Additionally, prior investigations have indicated that middle layers of language models offer better semantic features for predicting brain responses to natural language stimuli \cite{jain2018incorporating,toneva2019interpreting,lebel2021voxelwise,caucheteux2022brains}.
The first step yields a new set of vector-time pairs $(\boldsymbol{E}_{i}, t_i)$ where the contextual feature vector $\boldsymbol{E}_{i}$ is a $768$-dimensional embedding for character $c_i$. 

Second, these embeddings were resampled at times corresponding to fMRI acquisition by using a three-lobe Lanczos filter \cite{huth2016natural} and for each fMRI
acquisition time point $t$, we obtained a corresponding stimulus feature vector with $768$ dimensions.
Third, the stimulus feature vectors at five different delays $(t-1, t-2, t-3, t-4, t-5)$ were summed with weights $1.0$, $0.7$, $0.5$, $0.3$, and $0.1$ respectively to produce the distributed semantic feature representations of $768$ dimensions.
With a temporal resolution (TR) of $1.5$ s (one point five seconds), this weighted sum combines stimulus feature vectors captured at time points $1.5$ s, $3$ s, $4.5$ s, $6$ s, and $7.5$ s earlier for predicting responses at time $t$. 
For the linear baseline \cite{tang2023semantic}, we adhered to their method, in which the stimulus feature vectors at five different delays were concatenated to yield a $3,840$-dimensional semantic feature representation for the brain image acquired at time point $t$.
The parameters of the 3D CNN and the weights of the linear layers were tuned to align the compact feature vectors of brain images produced by the 3D CNN with these semantic feature representations.


\subsubsection*{Information bottleneck}

The information bottleneck (IB) technique was introduced to find the optimal tradeoff between accuracy and complexity when compressing a random variable $X$, given a joint probability distribution between $X$ and a relevant observed variable $Y$ \cite{tishby1999information}. 
Let $X$ represent an input random variable (an fMRI brain image here), and $Y$ denote a correlated output random variable (a character sequence here). 
Given the joint distribution $p(X, Y)$, our goal is to derive a compressed representation $Z$ of $X$ that effectively predicts $Y$ through learning a stochastic mapping $p(Z|X)$. 
In this study, $Z$ is the compact feature vector generated by the 3D CNN encoder from an fMRI brain image. 
The information bottleneck can be viewed as a rate-distortion problem, wherein a distortion function measures the effectiveness of predicting $Y$ from a compressed representation $Z$ compared to its direct prediction from $X$.
The information bottleneck minimizes the following objective:

\begin{equation}
    \mathcal{L}_{\text{IB}} = - \text{I}(Y;Z) + \beta \text{I}(X;Z)
\end{equation}

\noindent where $\text{I}(Y;Z)$ and $\text{I}(X;Z)$ are the mutual information of $Y$ and $Z$, and of $X$ and $Z$, respectively, and $\beta$ is a Lagrange multiplier. 
Minimizing this objective implies that $Z$ retains less information about $X$ while still preserving sufficient information for predicting $Y$.
The balance between the predictive power and the compression strength is controlled by the Lagrange multiplier $\beta$.
By increasing the value of $\beta$, we ``narrow the bottleneck'', favoring compression over predictive accuracy. 
The IB objective was introduced to maximize the predictive power of $Z$ while mitigating the inclusion of irrelevant and noisy information in $X$ by imposing some constraint on the amount of information that $Z$ carries about $X$.

By the definition of mutual information, we have:
\begin{equation} \label{eq:mi-xz}
    \text{I}(X;Z) = \mathbb{E}_{x,z} \left[ \log \frac{p(x,z)}{p(x)p(z)} \right] = \mathbb{E}_{x,z} \left[ \log \frac{p(z|x)}{p(z)} \right]
\end{equation}

\noindent where $\mathbb{E}_{x,z}[\cdot]$ denotes the expectation with respect to the joint distribution $p(X, Z)$.
In practice, we use the empirical distribution over a sample of $(x,z)$ pairs to approximate the distribution $p(X, Z)$.
In Equation \eqref{eq:mi-xz}, the troublesome term is $p(z) = \mathbb{E}_{x'}[p(z|x')]$, since even estimating it from a given training set necessitates an inner loop over all the brain images $x'$ contained within that set.
To avoid this, a variational IB was often used to replace $p(z)$ with some variational distribution $q(z)$ \cite{li2019specializing}.
This can only increase the value of our objective function since the difference between original and variational forms of this term is a Kullback-Leibler (KL) divergence and hence non-negative.

\begin{equation} \label{eq:vbi-xz}
    \underbrace{\mathbb{E}_{x,z} \left[ \log \frac{p(z|x)}{q(z)} \right]}_{\text{upper bound}} - \underbrace{\mathbb{E}_{x,z} \left[ \log \frac{p(z|x)}{p(z)} \right]}_{\text{I}(X;Z)} = \mathbb{E}_{x}  \left[ D_{\text{KL}}(p(z)||q(z)) \right] \ge 0
\end{equation}

\noindent Since the variational form (i.e., the first term above) is an upper bound for $\text{I}(X;Z)$, we can minimize this upper bound by adjusting both $p(z|x)$ and $q(z)$, making the bound as tight as possible.
The distribution $q(z)$ should be intentionally selected to be a simpler form than $p(z|x)$ while making $q(z)$ similar to the true posterior $p(z|x)$. 
Typically, the distribution $q(z)$ is restricted to a family of Gaussian distributions, and we choose a multivariate Gaussian distribution with a diagonal covariance matrix for $q(z)$.
In this case, the first term in Equation \eqref{eq:vbi-xz} can be rewritten as $D_{\text{KL}}(p(z|x)||q(z))$ so that this KL-divergence can be computed exactly as follows:

\begin{equation}
 D_{\text{KL}}(\mathcal{N}_1||\mathcal{N}_2) = \frac{1}{2}\left(\text{tr}(\Sigma^{-1}_2 \Sigma_1) + (\mu_2 - \mu_1)^{\top} \Sigma^{-1}_2(\mu_2 - \mu_1) - d + \log \frac{\text{det}(\Sigma_2)}{\text{det}(\Sigma_1)}\right)
\end{equation}

\noindent where $\mu_1$ and $\mu_2$ are the means and 
$\Sigma_1$ and $\Sigma_2$ are the covariance matrices of the Gaussian distributions $\mathcal{N}_1$ and $\mathcal{N}_2$, respectively, $d$ the dimensionality of the compressed representation vectors, $\det(\Sigma)$ the determinant of the covariance matrix, and $\text{tr}(\cdot)$ the trace of a matrix.

To maximize $\text{I}(Y;Z)$, we replace $p(y|z)$ with a variational approximation $q(y|z)$. 
The variational version is a lower bound of original $\text{I}(Y;Z)$ since the difference between them is a KL-divergence and hence non-negative again.

\begin{equation} \label{eq:vbi-yz}
    \underbrace{\mathbb{E}_{y,z} \left[ \log \frac{p(y|z)}{p(y)} \right]}_{\text{I}(Y;Z)} - \underbrace{\mathbb{E}_{y,z} \left[ \log \frac{q(y|z)}{p(y)} \right]}_{\text{lower bound}} = \mathbb{E}_{z}  \left[ D_{\text{KL}}(p(y|z)||q(y|z)) \right] \ge 0
\end{equation}

\noindent Therefore, the mutual information between $Y$ and $Z$ can be maximized by maximizing its lower bound.
The constant term $p(y)$ can be safely omitted during the optimization process, and we use the aforementioned 3D CNN as our variational distribution $q(y|z)$, which functions as the decoder.
With these approximations, the final loss function of our approach takes the following form:

\begin{equation} \label{eq:loss}
    \mathcal{L}(X, Y) = \mathbb{E}_{x,y} \left[  \mathbb{E}_{z \sim p(z|x)} \left[-\log q(y|z) + \beta D_{\text{KL}}(p(z|x)||q(z)) \right] \right]
\end{equation}

\noindent We apply stochastic gradient descent to optimize this objective. 
For a stochastic estimate of the objective, we begin by sampling some $(x, y)$ pairs from the training set. Subsequently, we use the 3D CNN decoder to compute the continuous value $z$. Given the continuous nature of $z$, we implement the reparametrization trick designed for multivariate Gaussian distributions \cite{rezende2014stochastic}. 
To compute the stochastic gradient, we run the back-propagation algorithm on this computation.



\subsubsection*{Generative large language model} \label{sec:llm}

Large language models (LLMs) have demonstrated remarkable performance on various natural language processing tasks \cite{devlin2019bert,brown2020language,raffel2020exploring}, and these models, characterized by their huge parameter counts---often numbering in the tens or hundreds of billions-----enable them to capture intricate language patterns and nuances. This capability is learned through pre-training on large corpora of diverse text data.
Recent advancements in the realm of LLMs are Generative Pre-trained Transformers (GPTs) \cite{radford2018improving,radford2019language,wei2021finetuned,chen2021evaluating,ouyang2022training,touvron2023llama,chowdhery2023palm,OpenAI2023GPT4TR}. 
GPTs leverage multi-head self-attention mechanisms for parallelized processing of input data, enabling the capture of long-range dependencies to predict the next token in a sentence or a sequence of text based on the context of preceding tokens.

We fine-tuned Baichuan-2 (the base version equipped with seven billion parameters)\footnote{https://huggingface.co/baichuan-inc/Baichuan2-7B-Base/tree/main (accessed on November 11, 2023)} on a corpus consisting of $14,988$ short stories (over $3$ million Chinese characters) from STORAL\footnote{https://github.com/thu-coai/moralstory (accessed  on November 15, 2023)}, $6,500$ news (over $9$ million characters)
from THUCNews\footnote{https://www.heywhale.com/mw/dataset/5de4b6d0ca27f8002c4c530a (accessed  on November 9, 2023)}, $919$ documents (over $280$ million characters) from WikiPedia\footnote{https://www.heywhale.com/mw/dataset/5d1ee7939f53a9002ce5910e/file (accessed on October 8, 2023)}, and $27$ novels (over $2$ million characters) from modern Chinese literature. 
The text used for training the decoder and testing was excluded from the dataset employed in the fine-tuning process.
Baichuan-2, the second generation of large-scale open-source, generative language models tailored for the Chinese language, was pre-trained on a massive corpus comprising $2.6$ trillion tokens.
We fine-tuned Baichuan-2 for $3$ epochs using a context length of $1024$.
During the fine-tuning stage, we used a standard language modeling objective to maximize the following likelihood for a given Chinese character sequence $S = \{c_1, c_2, \dots, c_n\}$ consisting of $n$ characters within the training corpus:

\begin{equation}
    \mathcal{L}(S) = \sum_{i=1}^n -\log p(c_i|c_{i-h}, \dots, c_{i-1})
\end{equation}

\noindent where $h$ denotes the maximum number of the context length, and punctuation was deliberately removed from the fine-tuning corpus. 
The parameters of Baichuan-2 model were tuned using stochastic gradient descent \cite{robbins1951stochastic}.  

\subsubsection*{Character rate model}

We established a character rate model for each subject to predict when characters were perceived. 
The character rate at each fMRI acquisition was defined as the number of stimulus characters occurring since the preceding acquisition.
Following the method proposed in \cite{tang2023semantic}, we employed regularized linear regression \cite{huth2016natural} to fit a set of weights predicting the character rate based on brain responses. 
For the prediction of character rates during perceived speech, we utilized brain responses from all available cortical regions.
To achieve this, a separate linear temporal filter incorporating five delays $(t + 1, t + 2, t + 3, t + 4, t + 5)$ was fitted for each voxel. 
With a temporal resolution (TR) of $1.5$ s, this involved concatenating responses from $1.5$ s, $3$ s, $4.5$ s, $6$ s, and $7.5$ s later to predict the character rate at time $t$. 
Given these brain responses, the model predicts the character rate at each acquisition. 
The temporal interval between consecutive acquisitions ($1.5$ s) was uniformly divided by the predicted character rates, rounded to the nearest non-negative integers, to estimate character onset times.

\subsubsection*{Character-based decoder}

In principle, the identification of the most probable stimulus character sequence could be achieved by assessing the similarity between recorded brain responses and possible character sequences in the form of their feature vectors and then selecting the one with the highest resemblance. 
However, the substantial search space for possible character sequences poses a formidable challenge for any algorithm to handle effectively.
To address this, we employed a generative language model to narrow down the search, proposing few plausible partial character sequences. 
These generated sequences aim to ensure both semantic and syntactic coherence in the resulting candidate sequences.
Building on the insights of \cite{tang2023semantic}, we employed a beam-search algorithm \cite{tillmann2003word} to maintain the $k$ most likely partial character sequences throughout the decoding process.

When new characters need to be predicted, the generative large language model produces continuations for each candidate in the beam.
The number of characters that should be generated was determined by the character rate model.
The language model predicts the probability distribution for the next character conditioned on the last $50$ of previously predicted characters.
Given the susceptibility of maximization-based decoding methods to degeneration, characterized by incoherent output or repetitive loops, we employed nucleus sampling \cite{holtzman2019curious} for sequence generation. 
This sampling proposes characters belonging to the top $\rho$ percent of the probability mass and having a probability within a factor $\eta$ of the most likely character. 
Furthermore, these candidate characters proposed by this sampling were filtered out if they did not appear at least twice in the character list from the dataset used to train the encoding model.

The encoding model evaluates and ranks each continuation by comparing it to recorded brain responses.
The $k$ most likely continuations across all candidates are retained in the beam for the next.
This iterative process continues until the number of predicted characters matches the quantity anticipated by the character rate model for a given sequence of brain images.
Finally, the decoder outputs the candidate character sequence with the highest similarity score against the recorded brain responses.

\subsubsection*{Hyper-parameters}


Unless otherwise specified, model parameters were initialized randomly, and training employed the AdamW optimizer \cite{loshchilov2017decoupled} with a weight decay of $0.01$ and a gradient clip range of $(-1, 1)$. The batch size was set to $128$, the number of epochs to $100$, and the learning rate to $0.00001$. 
For cross-subject experiments, the number of epochs was set to $10$ due to the larger training set compared to within-subject experiments. 
In the final loss function of Eq. (\ref{eq:loss}), the magnitude of the KL-divergence term outweighs that of the first term. 
To mitigate this imbalance, we first scale down the value of the KL-divergence term by a factor of $100$, and then set $\beta$ to $1.0$, thereby ensuring equal importance in optimizing both terms.
The training dataset comprised all $20$ articles and their corresponding audios (refer to Table \ref{tb:summary-1} to \ref{tb:summary-4} for descriptions), excluding article $14$ (``\emph{Ordinary World}'') for testing and article $13$ (``\emph{Chronicle of a Blood Merchant}'') for the validation set.
In the cross-subject setting, for each test subject, we used the data from all the other $19$ subjects as the training set, excluding the data associated with the subject intended for testing. The trained model was evaluated on the held-out data generated from the unseen test subject.
The hyperparameters of models were tuned on the validation set only. 
During decoding with the beam search algorithm, the beam size was set to $200$.
The percent of probability mass $\rho$ during nucleus sampling  was set to $90\%$, and the value of factor $\eta$ was set to $0.1$.
For each setting, the average accuracy is obtained over $5$ runs with different random initializations.

\subsection*{Datasets}

\subsubsection*{Subjects}

The task-based fMRI data were collected from a total of $20$ participants (age $23.7 \pm 1.8$ years, $10$ females and $10$ males). 
All participants were healthy and had normal hearing. The participants were postgraduate students with the same educational level. 
Before participation, fully informed consent was obtained from all participants by Fudan University.
The research procedures and ethical guidelines were also approved by the Fudan University institutional review board. 
Every subject was compensated at a rate of CNY $100$ per hour. 
The sample size was not pre-specified using statistical methods, and no binding was performed as there were no experimental groups in the fMRI analysis.
All available data were included in the analysis, and no exclusions were made.

\subsubsection*{fMRI data collection}

These fMRI data were acquired using an ultrahigh-field 7T Siemens Terra MR scanner at the Zhangjiang International Brain Imaging Centre of Fudan University, Shanghai, China. A $32$-channel Siemens volume coil (Siemens Healthcare) was used for data acquisition. Functional scans were obtained using an echo planar imaging (EPI) sequence with the following parameters: repetition time (TR) $= 1,500$ ms, echo time (TE) $= 25$ ms, flip angle $= 65$ degrees, voxel size $= 1.5 \times 1.5 \times 1.5$ mm, matrix size $= 128 \times 128$, field of view (FOV) $= 1,920 \times 1,920$ mm, slice thickness $= 1.5$ mm, and number of slices $= 96$. For better pre-processing quality, we also collected the T1-weighted anatomical data for each subject on a separate 3T scanner (Siemens Prisma) with the following parameters: TR$ = 2,500$ ms, TE $= 2.22$ ms, flip angle $= 8$ degrees, matrix size $= 300 \times 320$, FOV $= 240 \times 256$ mm, slice thickness $= 0.8$ mm, and number of slices $= 208$. 

\subsubsection*{Experimental tasks}


In the experiment, a collection of $20$ articles, each ranging from $8$ to $12$ minutes in duration, served as stimuli for the Chinese natural speech task. These articles spanned diverse genres, encompassing fairy tales, fables, novels, essays, news, and scientific essays. 
A comprehensive summary of each article is provided in Tables \ref{tb:summary-1} to \ref{tb:summary-4}. 
Each article was presented in a separate fMRI scanning session, and a total of $20$ sessions were conducted for each participant. 
The duration of each scan was customize to match the length of the corresponding article, with an additional $2$-second silence period both preceding and following the article's presentation.
The $20$ fMRI scans were scheduled over different days, resulting in a cumulative scanning time of approximately $162$ minutes.
Participants experienced the articles through Sensimetrics S14 in-ear piezoelectric headphones. To ensure accuracy, each article underwent manual transcription by a single listener, and subsequent alignment of the audio with the transcript was independently verified by two postgraduate students, ensuring a thorough cross-check of the alignment accuracy.

\subsubsection*{fMRI data pre-processing}


The fMRI data were preprocessed using Statistical Parametric Mapping-12 (SPM12)\footnote{http://www.fil.ion.ucl.ac.uk/spm}. To ensure T1 equilibrium, several volumes were intentionally omitted before the trigger. Subsequently, temporal realignment to the middle EPI volume and correction for head movement were implemented. The structural images for each participant were registered to the mean EPI image, segmented, and normalized to Montreal Neurological Institute (MNI) space. The realigned EPI volumes were then normalized to MNI space, employing deformation field parameters derived from the structural image normalization. Finally, the normalized EPI volumes underwent spatial smoothing with a $6 \times 6 \times 6$ mm Gaussian kernel and high-pass filtering to mitigate noise.

\subsubsection*{Cortical regions}

The task-based fMRI data used for extracting semantic representations encompassed the whole brain, including, but not limited to, the anterior temporal lobe (ATL) region, the prefrontal region, and the parietal-temporal-occipital association region.


The choice of ATL regions stems from their crucial role in integrating semantic information \cite{ralph2017neural,bi2021dual}. Specifically, the ATL regions, anatomically delineated as the six regions identified in the Harvard-Oxford Atlas \cite{jenkinson2012fsl} with a grey matter probability exceeding $0.2$, including the temporal pole, the the anterior superior temporal gyrus, the anterior middle temporal gyrus, the anterior inferior temporal gyrus, the anterior temporal fusiform cortex, and the anterior parahippocampal gyrus.


The parietal-temporal-occipital association region has gained recognition for its crucial role in facilitating the integration of diverse perceptual information \cite{binder2011neurobiology} and has also exhibited potential for semantic prediction \cite{huth2016natural}. This region encompasses a composite of $38$ and $90$ cortical regions identified in the Automated Anatomical Labeling (AAL) template \cite{tzourio2002automated}, including the bilateral calcarine, cuneus, lingual, superior occipital, middle occipital, inferior occipital, fusiform, postcentral, superior parietal, inferior parietal, supramarginal, angular, precuneus, heschl, superior temporal, middle temporal, inferior temporal, superior temporal pole, and middle temporal pole.


The activation of the prefrontal region has been observed across various language tasks \cite{fedorenko2014reworking} and is recognized as a component of the semantic map underlying language decoding \cite{huth2016natural}. The prefrontal region mask was constructed from 20 out of 90 cortical regions outlined in the AAL template. 
These regions include the bilateral precentral, superior frontal, superior orbitofrontal, middle frontal, middle orbitofrontal, inferior orbitofrontal, inferior pars opercularis, inferior pars triangularis, superior and medial frontal and medial orbitofrontal.


\subsection*{Evaluation}

\subsubsection*{Language similarity metrics}

We employed four similarity metrics--- BLEU, METEOR, BERT Score, and SBERT---to evaluate the performance of decoders by comparing decoded character sequences to reference character sequences. 
For the perceived speech, stimulus transcripts served as the reference sequences, and the temporal alignment for each pronunciation in the played audios was manually annotated and validated by two post-graduate students.

BLEU (bilingual evaluation understudy) remains one of the most popular automated metrics for evaluating the quality of texts translated by machine from one natural language to another \cite{papineni2002bleu}. 
This evaluation is based on the quantification of overlapping $n$-grams between two textual sequences. 
In alignment with the approach outlined in \cite{tang2023semantic}, we adopted the unigram variant BLEU-1.
METEOR \cite{banerjee2005meteor} calculates a similarity score by combining unigram-precision (the number of predicted $1$-grams present in the reference sequence), unigram-recall (the number of reference $1$-grams found in the predicted sequence), and a measure of fragmentation designed to directly capture the well-ordered arrangement of matched characters in the predicted sequence relative to the reference.
BERT Score \cite{zhang2020bertscore} leverages a bidirectional transformer-based language model to generate a contextual embedding for each character in the predicted and reference sequences and then computes a similarity score over these embeddings.
We used the recall variant of BERT Score, incorporating inverse document frequency (IDF) \cite{sparck1972statistical} as importance weights. 
These weights were computed across texts in the training dataset to enhance the the precision of the evaluation by assigning greater weights to less-frequently occurring characters over their more-frequent counterparts.

The three metrics described above primarily focus on character-based evaluation. 
However, it is crucial to take into account factors such as character order and the variations in expressing the same meaning when assessing the similarity of textual sequences, particularly in the context of continuous languages.
Consequently, we also used SBERT \cite{reimers2019sentence} as a semantic textual similarity metric for pairs of character sequences. 
For the development of SBERT, the Siamese network architecture \cite{bromley1993signature,chicco2021siamese} was used to produce fixed-sized feature vectors for input character sequences, and the cosine similarity was used to compute the similarity score between a pair of character sequences in their form of features vectors.
SBERT was used for all analysis where the language similarity metric is not specified.

We computed the similarity scores between the predicted and reference character sequences within an each window centered around every second of the stimulus. 
For a given stimulus, we considered all successive windows of size $20$ seconds, sliding over the stimulus with a one-second interval. 
Subsequently, these window similarity scores were averaged across all windows to evaluate the decoder's performance in predicting the full stimulus speech.
We established a lower bound for each metric by
calculating the mean similarity between the actual stimulus characters and ten null character sequences.
These null sequences were generated by the fined-tuned generative language model using ten initial characters as context but without using any brain data. 


\subsubsection*{Statistical testing}

In order to assess the the statistical significance of the similarity scores produced by the decoder, null character sequences were created through sampling from a finely-tuned generative language model without using any brain data.
Given that language similarity metrics can be affected by the length of predicted sequences, we first employed the character rate model to estimate the number of characters to be generated based on brain images, and then generated null sequences in accordance with the predicted character counts. This approach allowed us to effectively isolate the decoder's capacity to capture semantic information from the brain data, minimizing the impact of variations in sequence length on the evaluation of similarity scores.

We applied the same beam search procedures in the generation of null character sequences as the actual decoder. 
The null model employed nucleus sampling at each predicted character time to generate continuations from the language model while maintaining a beam of ten candidate partial sequence. 
The only difference between the actual decoder and the null model lies in the assignment of a random generation probability to each continuation in the null model, as opposed to ranking them based on brain data. 
After iterating through all predicted character time points, the null model yielded the candidate sequence with the highest likelihood. 
This process was repeated $200$ times to generate null sequences. 
By abstaining from employing brain data to select characters, this process closely mirrors the actual decoder, embodying the null hypothesis that the decoder fails to recover meaningful information about the stimulus from the brain data.
To establish a null distribution of decoding scores, we evaluated the null sequences against a reference sequence, and then conducted a comparison between the observed decoding scores and this null distribution using a one-sided non-parametric test. 
The resultant $P$-values were computed as the fraction of null sequences exhibiting a decoding score greater than or equal to the observed decoding score.
Unless otherwise specified, all tests were performed within each subject and replicated across the all subjects. 
The range across subjects was reported for all quantitative results.
Normal (or Gaussian) distribution assumptions were made for data, but formal testing was not conducted due to the limited number of subjects.


\subsubsection*{Ablation study and training dynamics}

\begin{figure} 
    \centering
    \begin{subfigure}[valign=t]{0.44\textwidth}
        \includegraphics[width=\textwidth]{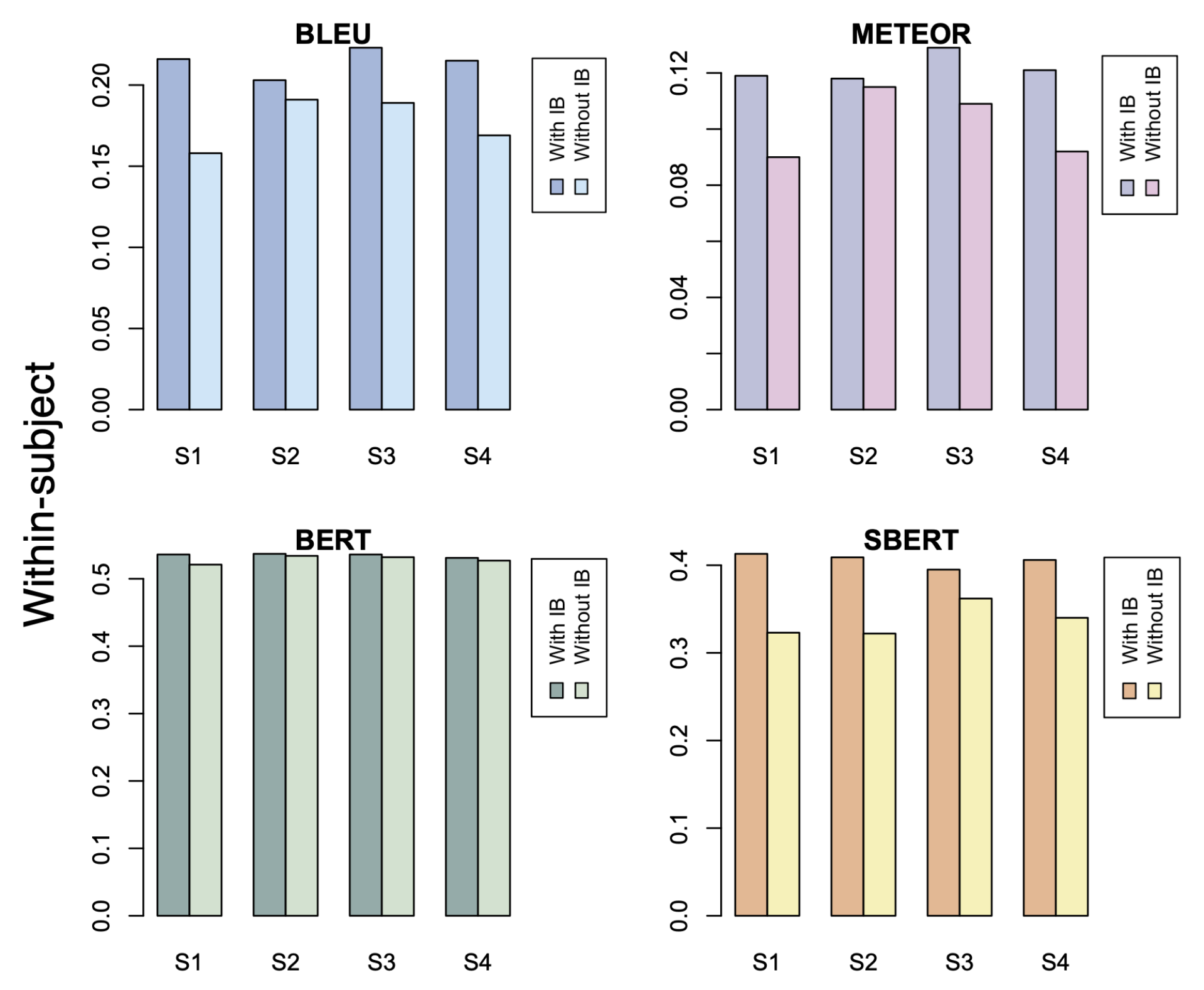}
    \end{subfigure}
    \begin{subfigure}[valign=t]{0.44\textwidth}
        \includegraphics[width=\textwidth]{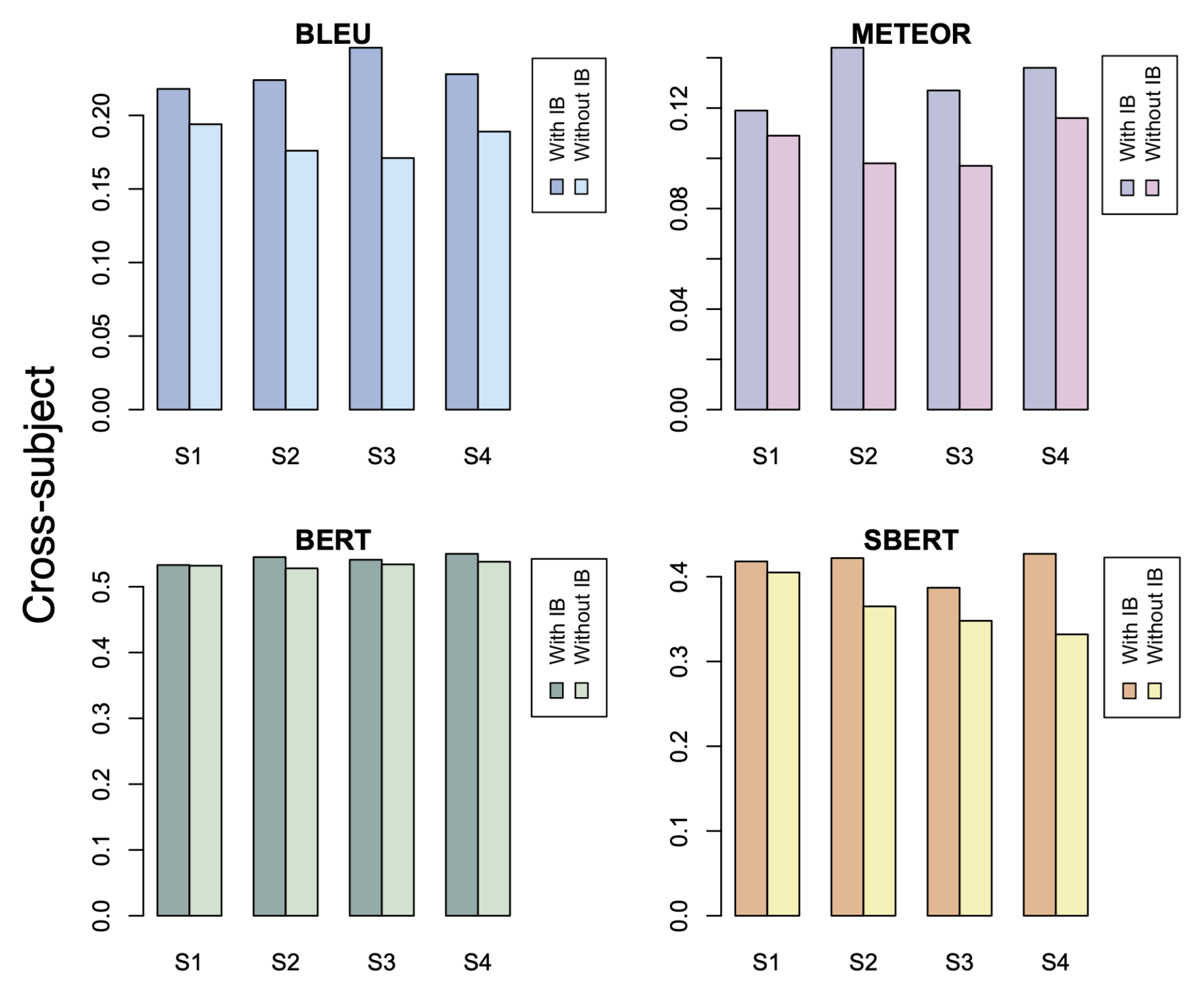}
    \end{subfigure}
    \begin{subfigure}[valign=t]{0.44\textwidth}
         \includegraphics[width=\textwidth]{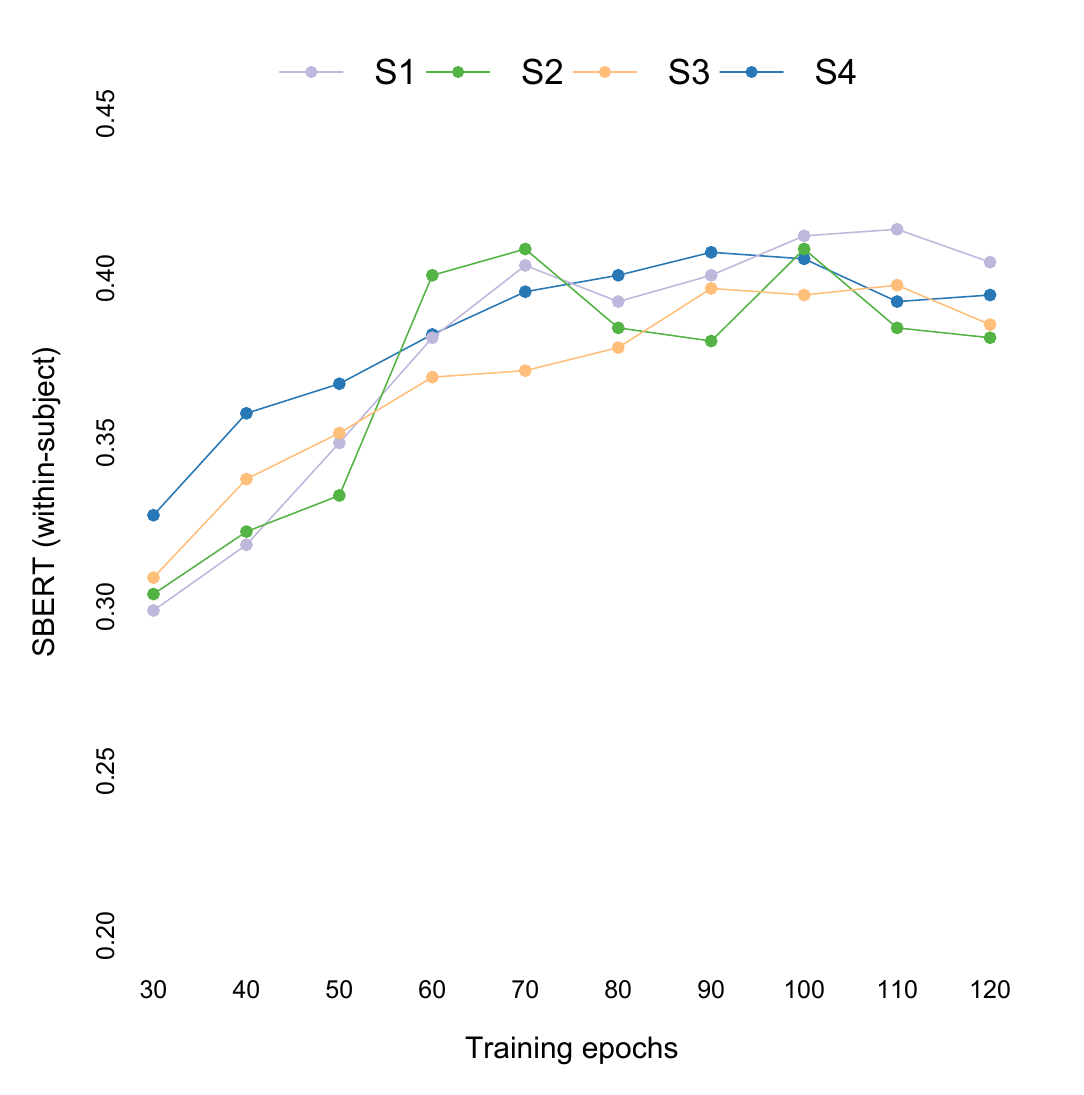}
    \end{subfigure}
    \begin{subfigure}[valign=t]{0.44\textwidth}
        \includegraphics[width=\textwidth]{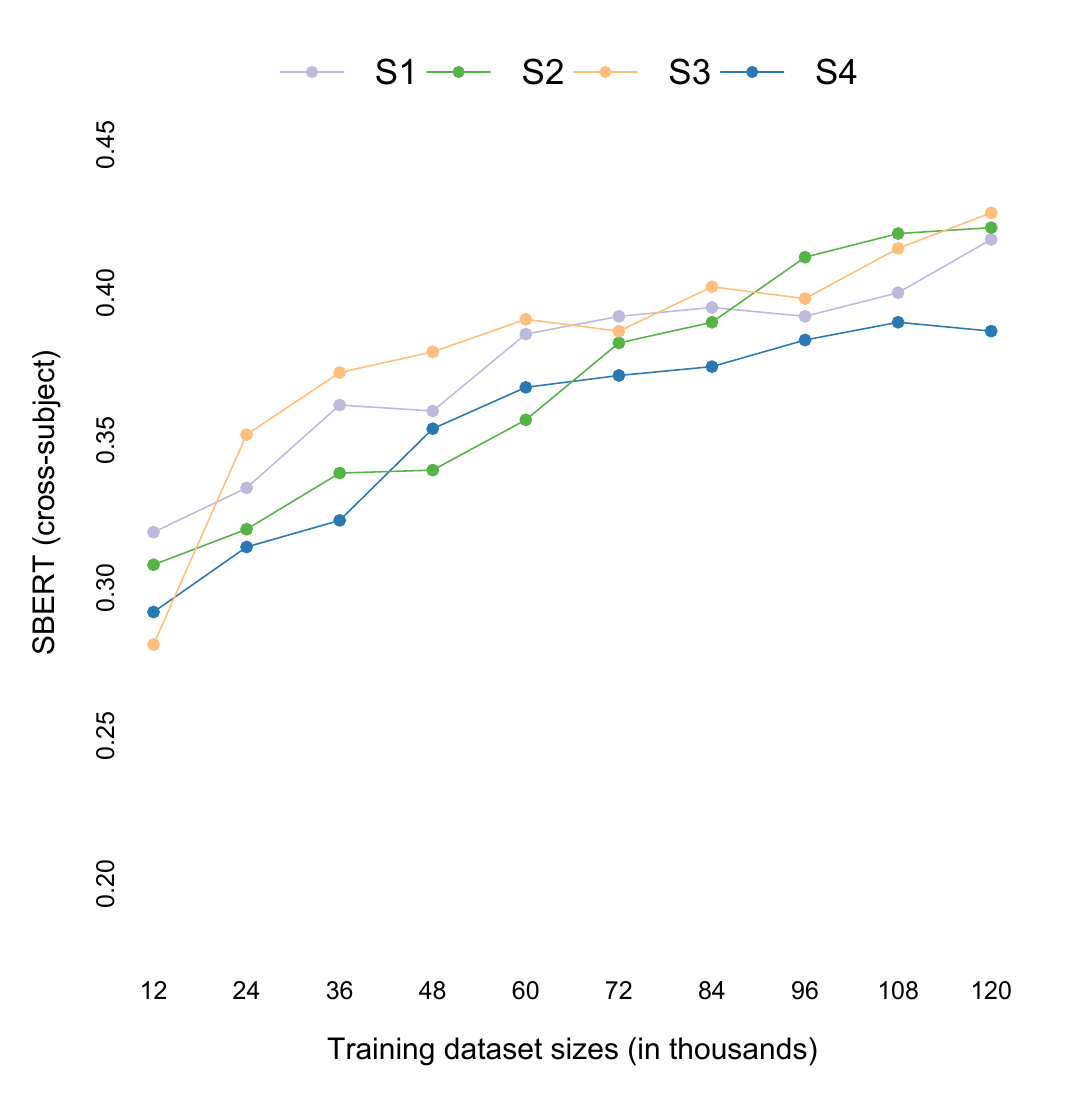}

    \end{subfigure}

    \caption{\small Exploring the impact of information bottleneck, training epochs, and dataset size on model performance. \textbf{(a)} A comparison of language similarity scores between our ``full-fledged'' 3dC-IB model (denoted as ``With IB'') and its ablated variant without the information bottleneck (denoted as ``Without IB'') in the within-subject setting. \textbf{(b)} A similar comparison in the cross-subject setting. The 3dC-IB model's predictions for a test article were significantly more similar to the actual stimuli than those provided by its ablated version across all language similarity metrics, in both within-subject and cross-subject contexts. \textbf{(c)} Variation of language similarity scores with the number of training epochs in the within-subject context. Decoding scores generally showed a trend of increasing with the number of epochs up to $110$, after which overfitting was observed. \textbf{(d)} To test if decoding performance is limited by the size of the training dataset in the cross-subject setting, decoders were trained with varying amounts of data, ranging from $10\%$ to $100\%$ (i.e., $12$k to $120$k). As the size of the training dataset increased decoding performance was improved, indicating potential for further enhancement of our decoder's performance with the availability of more training data.
    } 
    \label{fig:ablation-study}
\end{figure}

To investigate the significance of the information bottleneck (IB) within our neural network-based decoder and assess its impact on decoding performance, we conducted an evaluation of a modified version of our decoder where the IB layer was removed. 
This variant was implemented by removing the last term from the loss function of Eq. (\ref{eq:loss}). 
The bar plots depicted in Fig. \ref{fig:ablation-study} (a) and (b) reveal that the IB component contributes to enhanced decoding accuracy across four metrics in both within-subject and cross-subject settings. 
By ``narrowing the bottleneck'' and reducing the noise present in raw BOLD fMRI recordings, the IB layer proves as a pivotal component in boosting decoding performance. 
While Figure \ref{fig:ablation-study} presents results for four subjects, consistent trends were observed across all $20$ subjects in this ablation study, demonstrating the significance of the IB layer and its efficacy in improving the signal-to-noise ratio of brain activities.

To examine the impact of varying the number of training epochs on decoding performance, we evaluated decoders trained with different epoch counts. 
In Figure \ref{fig:ablation-study} (c), we showed the variation of SBERT scores alongside the number of training epochs in the within-subject context. 
Generally, language similarity scores exhibit an upward trend with an increasing number of training epochs, until indications of overfitting become apparent beyond $110$ epochs.
In the context of within-subject analysis, our training dataset consists of a finite set of fMRI recordings, obtained during a $2.7$-hour session of listening to naturally spoken narratives.
Consequently, evaluating whether the size of the training dataset impacts decoding performance within subjects becomes inconsequential.
However, for cross-subject models, we can access a substantial corpus of $120$k fMRI images for training, even though the data from the test subject have been excluded. 
This allows us to explore how decoding performance is influenced by variations in the size of the training dataset. 
To explore this relationship, we partitioned the entire training dataset into ten equally-sized segments for each test subject and proceeded to train decoders on subsets of increasing sizes. 
As illustrated in Fig. \ref{fig:ablation-study} (d), a discernible trend emerges wherein decoding scores consistently exhibit incremental improvement with each augmentation in the size of the training dataset. 
This observation suggests promising prospects for further enhancing the performance of our decoder through the provision of additional training data.









\bibliography{reference}

\begin{thebibliography}{100}
\expandafter\ifx\csname url\endcsname\relax
  \def\url#1{\burl{#1}}\fi
\expandafter\ifx\csname urlprefix\endcsname\relax\def\urlprefix{URL }\fi
\providecommand{\bibinfo}[2]{#2}
\providecommand{\eprint}[2][]{\url{#2}}
\providecommand{\doi}[1]{\url{https://doi.org/#1}}
\bibcommenthead

\bibitem{zhang2013contribution}
\bibinfo{author}{Zhang, M.} \emph{et~al.}
\newblock \bibinfo{title}{The contribution of the left mid-fusiform cortical thickness to {C}hinese and {E}nglish reading in a large {C}hinese sample}.
\newblock \emph{\bibinfo{journal}{Neuroimage}} \textbf{\bibinfo{volume}{65}}, \bibinfo{pages}{250--256} (\bibinfo{year}{2013}).

\bibitem{bolger2005cross}
\bibinfo{author}{Bolger, D.~J.}, \bibinfo{author}{Perfetti, C.~A.} \& \bibinfo{author}{Schneider, W.}
\newblock \bibinfo{title}{Cross-cultural effect on the brain revisited: Universal structures plus writing system variation}.
\newblock \emph{\bibinfo{journal}{Human brain mapping}} \textbf{\bibinfo{volume}{25}}, \bibinfo{pages}{92--104} (\bibinfo{year}{2005}).

\bibitem{sun2021independent}
\bibinfo{author}{Sun, Z.}, \bibinfo{author}{Shi, Y.}, \bibinfo{author}{Guo, P.}, \bibinfo{author}{Yang, Y.} \& \bibinfo{author}{Zhu, Z.}
\newblock \bibinfo{title}{Independent syntactic representation identified in left front-temporal cortex during {C}hinese sentence comprehension}.
\newblock \emph{\bibinfo{journal}{Brain and Language}} \textbf{\bibinfo{volume}{214}}, \bibinfo{pages}{104907} (\bibinfo{year}{2021}).

\bibitem{cao2010cultural}
\bibinfo{author}{Cao, F.} \emph{et~al.}
\newblock \bibinfo{title}{Cultural constraints on brain development: Evidence from a developmental study of visual word processing in mandarin {C}hinese}.
\newblock \emph{\bibinfo{journal}{Cerebral Cortex}} \textbf{\bibinfo{volume}{20}}, \bibinfo{pages}{1223--1233} (\bibinfo{year}{2010}).

\bibitem{coulmas2003writing}
\bibinfo{author}{Coulmas, F.}
\newblock \emph{\bibinfo{title}{Writing systems: An introduction to their linguistic analysis}}  (\bibinfo{publisher}{Cambridge University Press}, \bibinfo{year}{2003}).

\bibitem{zhao2017rethinking}
\bibinfo{author}{Zhao, R.}, \bibinfo{author}{Fan, R.}, \bibinfo{author}{Liu, M.}, \bibinfo{author}{Wang, X.} \& \bibinfo{author}{Yang, J.}
\newblock \bibinfo{title}{Rethinking the function of brain regions for reading {C}hinese characters in a meta-analysis of f{MRI} studies}.
\newblock \emph{\bibinfo{journal}{Journal of Neurolinguistics}} \textbf{\bibinfo{volume}{44}}, \bibinfo{pages}{120--133} (\bibinfo{year}{2017}).

\bibitem{price2012review}
\bibinfo{author}{Price, C.~J.}
\newblock \bibinfo{title}{A review and synthesis of the first 20 years of {PET} and f{MRI} studies of heard speech, spoken language and reading}.
\newblock \emph{\bibinfo{journal}{Neuroimage}} \textbf{\bibinfo{volume}{62}}, \bibinfo{pages}{816--847} (\bibinfo{year}{2012}).

\bibitem{guo2022brain}
\bibinfo{author}{Guo, W.}, \bibinfo{author}{Geng, S.}, \bibinfo{author}{Cao, M.} \& \bibinfo{author}{Feng, J.}
\newblock \bibinfo{title}{The brain connectome for chinese reading}.
\newblock \emph{\bibinfo{journal}{Neuroscience Bulletin}} \textbf{\bibinfo{volume}{38}}, \bibinfo{pages}{1097--1113} (\bibinfo{year}{2022}).

\bibitem{kuo2004orthographic}
\bibinfo{author}{Kuo, W.-J.} \emph{et~al.}
\newblock \bibinfo{title}{Orthographic and phonological processing of {C}hinese characters: an {fMRI} study}.
\newblock \emph{\bibinfo{journal}{Neuroimage}} \textbf{\bibinfo{volume}{21}}, \bibinfo{pages}{1721--1731} (\bibinfo{year}{2004}).

\bibitem{liu2006dissociated}
\bibinfo{author}{Liu, C.-L.} \emph{et~al.}
\newblock \bibinfo{title}{Dissociated roles of the middle frontal gyri in the processing of {C}hinese characters}.
\newblock \emph{\bibinfo{journal}{Neuroreport}} \textbf{\bibinfo{volume}{17}}, \bibinfo{pages}{1397--1401} (\bibinfo{year}{2006}).

\bibitem{booth2006specialization}
\bibinfo{author}{Booth, J.~R.} \emph{et~al.}
\newblock \bibinfo{title}{Specialization of phonological and semantic processing in {C}hinese word reading}.
\newblock \emph{\bibinfo{journal}{Brain research}} \textbf{\bibinfo{volume}{1071}}, \bibinfo{pages}{197--207} (\bibinfo{year}{2006}).

\bibitem{wu2012meta}
\bibinfo{author}{Wu, C.-Y.}, \bibinfo{author}{Ho, M.-H.~R.} \& \bibinfo{author}{Chen, S.-H.~A.}
\newblock \bibinfo{title}{A meta-analysis of {fMRI} studies on {C}hinese orthographic, phonological, and semantic processing}.
\newblock \emph{\bibinfo{journal}{Neuroimage}} \textbf{\bibinfo{volume}{63}}, \bibinfo{pages}{381--391} (\bibinfo{year}{2012}).

\bibitem{liu2022functional}
\bibinfo{author}{Liu, C.~Y.}, \bibinfo{author}{Tao, R.}, \bibinfo{author}{Qin, L.}, \bibinfo{author}{Matthews, S.} \& \bibinfo{author}{Siok, W.~T.}
\newblock \bibinfo{title}{Functional connectivity during orthographic, phonological, and semantic processing of {C}hinese characters identifies distinct visuospatial and phonosemantic networks}.
\newblock \emph{\bibinfo{journal}{Human Brain Mapping}} \textbf{\bibinfo{volume}{43}}, \bibinfo{pages}{5066--5080} (\bibinfo{year}{2022}).

\bibitem{tan2005neuroanatomical}
\bibinfo{author}{Tan, L.~H.}, \bibinfo{author}{Laird, A.~R.}, \bibinfo{author}{Li, K.} \& \bibinfo{author}{Fox, P.~T.}
\newblock \bibinfo{title}{Neuroanatomical correlates of phonological processing of {C}hinese characters and alphabetic words: A meta-analysis}.
\newblock \emph{\bibinfo{journal}{Human brain mapping}} \textbf{\bibinfo{volume}{25}}, \bibinfo{pages}{83--91} (\bibinfo{year}{2005}).

\bibitem{dash2018determining}
\bibinfo{author}{Dash, D.} \emph{et~al.}
\newblock \bibinfo{title}{Determining the optimal number of {MEG} trials: A machine learning and speech decoding perspective}.
\newblock \emph{\bibinfo{journal}{Brain Informatics: International Conference}} \bibinfo{pages}{163--172} (\bibinfo{year}{2018}).

\bibitem{roy2019deep}
\bibinfo{author}{Roy, Y.} \emph{et~al.}
\newblock \bibinfo{title}{Deep learning-based electroencephalography analysis: a systematic review}.
\newblock \emph{\bibinfo{journal}{Journal of neural engineering}} \textbf{\bibinfo{volume}{16}}, \bibinfo{pages}{051001} (\bibinfo{year}{2019}).

\bibitem{nishimoto2011reconstructing}
\bibinfo{author}{Nishimoto, S.} \emph{et~al.}
\newblock \bibinfo{title}{Reconstructing visual experiences from brain activity evoked by natural movies}.
\newblock \emph{\bibinfo{journal}{Current biology}} \textbf{\bibinfo{volume}{21}}, \bibinfo{pages}{1641--1646} (\bibinfo{year}{2011}).

\bibitem{kamitani2005decoding}
\bibinfo{author}{Kamitani, Y.} \& \bibinfo{author}{Tong, F.}
\newblock \bibinfo{title}{Decoding the visual and subjective contents of the human brain}.
\newblock \emph{\bibinfo{journal}{Nature neuroscience}} \textbf{\bibinfo{volume}{8}}, \bibinfo{pages}{679--685} (\bibinfo{year}{2005}).

\bibitem{haxby2001distributed}
\bibinfo{author}{Haxby, J.~V.} \emph{et~al.}
\newblock \bibinfo{title}{Distributed and overlapping representations of faces and objects in ventral temporal cortex}.
\newblock \emph{\bibinfo{journal}{Science}} \textbf{\bibinfo{volume}{293}}, \bibinfo{pages}{2425--2430} (\bibinfo{year}{2001}).

\bibitem{king2020encoding}
\bibinfo{author}{King, J.-R.} \emph{et~al.}
\newblock \bibinfo{title}{Encoding and decoding framework to uncover the algorithms of cognition}.
\newblock \emph{\bibinfo{journal}{The Cognitive Neurosciences,}} \textbf{\bibinfo{volume}{6}}, \bibinfo{pages}{691--702} (\bibinfo{year}{2020}).

\bibitem{schirrmeister2017deep}
\bibinfo{author}{Schirrmeister, R.~T.} \emph{et~al.}
\newblock \bibinfo{title}{Deep learning with convolutional neural networks for eeg decoding and visualization}.
\newblock \emph{\bibinfo{journal}{Human brain mapping}} \textbf{\bibinfo{volume}{38}}, \bibinfo{pages}{5391--5420} (\bibinfo{year}{2017}).

\bibitem{hamalainen1993magnetoencephalography}
\bibinfo{author}{H{\"a}m{\"a}l{\"a}inen, M.}, \bibinfo{author}{Hari, R.}, \bibinfo{author}{Ilmoniemi, R.~J.}, \bibinfo{author}{Knuutila, J.} \& \bibinfo{author}{Lounasmaa, O.~V.}
\newblock \bibinfo{title}{Magnetoencephalography—theory, instrumentation, and applications to noninvasive studies of the working human brain}.
\newblock \emph{\bibinfo{journal}{Reviews of modern Physics}} \textbf{\bibinfo{volume}{65}}, \bibinfo{pages}{413} (\bibinfo{year}{1993}).

\bibitem{vaswani2017attention}
\bibinfo{author}{Vaswani, A.} \emph{et~al.}
\newblock \bibinfo{title}{Attention is all you need}.
\newblock \emph{\bibinfo{journal}{Advances in neural information processing systems}} \textbf{\bibinfo{volume}{30}} (\bibinfo{year}{2017}).

\bibitem{loshchilov2016sgdr}
\bibinfo{author}{Loshchilov, I.} \& \bibinfo{author}{Hutter, F.}
\newblock \bibinfo{title}{{SGDR:} stochastic gradient descent with warm restarts}.
\newblock \emph{\bibinfo{journal}{Proceedings of the International Conference on Learning Representations (ICLR)}}  (\bibinfo{year}{2017}).

\bibitem{loshchilov2017decoupled}
\bibinfo{author}{Loshchilov, I.} \& \bibinfo{author}{Hutter, F.}
\newblock \bibinfo{title}{Decoupled weight decay regularization}.
\newblock \emph{\bibinfo{journal}{Proceedings of the International Conference on Learning Representations}}  (\bibinfo{year}{2019}).

\bibitem{chicco2021siamese}
\bibinfo{author}{Chicco, D.}
\newblock \bibinfo{title}{Siamese neural networks: An overview}.
\newblock \emph{\bibinfo{journal}{Artificial neural networks}} \bibinfo{pages}{73--94} (\bibinfo{year}{2021}).

\bibitem{bromley1993signature}
\bibinfo{author}{Bromley, J.}, \bibinfo{author}{Guyon, I.}, \bibinfo{author}{LeCun, Y.}, \bibinfo{author}{S{\"a}ckinger, E.} \& \bibinfo{author}{Shah, R.}
\newblock \bibinfo{title}{Signature verification using a ``{S}iamese'' time delay neural network}.
\newblock \emph{\bibinfo{journal}{Proceedings of Advances in neural information processing systems}} \textbf{\bibinfo{volume}{6}} (\bibinfo{year}{1993}).

\bibitem{reimers2019sentence}
\bibinfo{author}{Reimers, N.} \& \bibinfo{author}{Gurevych, I.}
\newblock \bibinfo{title}{Sentence-{BERT}: Sentence embeddings using siamese {BERT}-networks}.
\newblock \emph{\bibinfo{journal}{Proceedings of Conference on Empirical Methods in Natural Language Processing and the International Joint Conference on Natural Language Processing}}  (\bibinfo{year}{2019}).

\bibitem{sparck1972statistical}
\bibinfo{author}{Sparck~Jones, K.}
\newblock \bibinfo{title}{A statistical interpretation of term specificity and its application in retrieval}.
\newblock \emph{\bibinfo{journal}{Journal of documentation}} \textbf{\bibinfo{volume}{28}}, \bibinfo{pages}{11--21} (\bibinfo{year}{1972}).

\bibitem{zhang2020bertscore}
\bibinfo{author}{Zhang, T.}, \bibinfo{author}{Kishore, V.}, \bibinfo{author}{Wu, F.}, \bibinfo{author}{Weinberger, K.~Q.} \& \bibinfo{author}{Artzi, Y.}
\newblock \bibinfo{title}{{BERTS}core: Evaluating text generation with {BERT}}.
\newblock \emph{\bibinfo{journal}{Proceddings of the International Conference on Learning Representations}}  (\bibinfo{year}{2020}).

\bibitem{banerjee2005meteor}
\bibinfo{author}{Banerjee, S.} \& \bibinfo{author}{Lavie, A.}
\newblock \bibinfo{title}{{METEOR}: An automatic metric for mt evaluation with improved correlation with human judgments}.
\newblock \emph{\bibinfo{journal}{Proceedings of the ACL workshop on intrinsic and extrinsic evaluation measures for machine translation and/or summarization}} \bibinfo{pages}{65--72} (\bibinfo{year}{2005}).

\bibitem{papineni2002bleu}
\bibinfo{author}{Papineni, K.}, \bibinfo{author}{Roukos, S.}, \bibinfo{author}{Ward, T.} \& \bibinfo{author}{Zhu, W.-J.}
\newblock \bibinfo{title}{{BLEU}: a method for automatic evaluation of machine translation}.
\newblock \emph{\bibinfo{journal}{Proceedings of the Annual Meeting of the Association for Computational Linguistics}} \bibinfo{pages}{311--318} (\bibinfo{year}{2002}).

\bibitem{holtzman2019curious}
\bibinfo{author}{Holtzman, A.}, \bibinfo{author}{Buys, J.}, \bibinfo{author}{Du, L.}, \bibinfo{author}{Forbes, M.} \& \bibinfo{author}{Choi, Y.}
\newblock \bibinfo{title}{The curious case of neural text degeneration}.
\newblock \emph{\bibinfo{journal}{Proceddings of the International Conference on Learning Representations}}  (\bibinfo{year}{2020}).

\bibitem{robbins1951stochastic}
\bibinfo{author}{Robbins, H.} \& \bibinfo{author}{Monro, S.}
\newblock \bibinfo{title}{A stochastic approximation method}.
\newblock \emph{\bibinfo{journal}{The annals of mathematical statistics}} \bibinfo{pages}{400--407} (\bibinfo{year}{1951}).

\bibitem{ouyang2022training}
\bibinfo{author}{Ouyang, L.} \emph{et~al.}
\newblock \bibinfo{title}{Training language models to follow instructions with human feedback}.
\newblock \emph{\bibinfo{journal}{Proceddings of the Advances in Neural Information Processing Systems}} \textbf{\bibinfo{volume}{35}}, \bibinfo{pages}{27730--27744} (\bibinfo{year}{2022}).

\bibitem{chen2021evaluating}
\bibinfo{author}{Chen, M.} \emph{et~al.}
\newblock \bibinfo{title}{Evaluating large language models trained on code}.
\newblock \emph{\bibinfo{journal}{arXiv preprint arXiv:2107.03374}}  (\bibinfo{year}{2021}).

\bibitem{radford2019language}
\bibinfo{author}{Radford, A.} \emph{et~al.}
\newblock \bibinfo{title}{Language models are unsupervised multitask learners}.
\newblock \emph{\bibinfo{journal}{OpenAI}} \textbf{\bibinfo{volume}{1}}, \bibinfo{pages}{9} (\bibinfo{year}{2019}).

\bibitem{wei2021finetuned}
\bibinfo{author}{Wei, J.} \emph{et~al.}
\newblock \bibinfo{title}{Finetuned language models are zero-shot learners}.
\newblock \emph{\bibinfo{journal}{Proceddings of the International Conference on Learning Representations}}  (\bibinfo{year}{2022}).

\bibitem{OpenAI2023GPT4TR}
\bibinfo{author}{OpenAI}.
\newblock \bibinfo{title}{{GPT}-4 technical report}.
\newblock \emph{\bibinfo{journal}{ArXiv}} \textbf{\bibinfo{volume}{abs/2303.08774}} (\bibinfo{year}{2023}).

\bibitem{touvron2023llama}
\bibinfo{author}{Touvron, H.} \emph{et~al.}
\newblock \bibinfo{title}{{LL}a{MA}: Open and efficient foundation language models}.
\newblock \emph{\bibinfo{journal}{arXiv preprint arXiv:2302.13971}}  (\bibinfo{year}{2023}).

\bibitem{chowdhery2023palm}
\bibinfo{author}{Chowdhery, A.} \emph{et~al.}
\newblock \bibinfo{title}{Palm: Scaling language modeling with pathways}.
\newblock \emph{\bibinfo{journal}{Journal of Machine Learning Research}} \textbf{\bibinfo{volume}{24}}, \bibinfo{pages}{1--113} (\bibinfo{year}{2023}).

\bibitem{radford2018improving}
\bibinfo{author}{Radford, A.}, \bibinfo{author}{Narasimhan, K.}, \bibinfo{author}{Salimans, T.}, \bibinfo{author}{Sutskever, I.} \emph{et~al.}
\newblock \bibinfo{title}{Improving language understanding by generative pre-training}.
\newblock \emph{\bibinfo{journal}{OpenAI}}  (\bibinfo{year}{2018}).

\bibitem{raffel2020exploring}
\bibinfo{author}{Raffel, C.} \emph{et~al.}
\newblock \bibinfo{title}{Exploring the limits of transfer learning with a unified text-to-text transformer}.
\newblock \emph{\bibinfo{journal}{The Journal of Machine Learning Research}} \textbf{\bibinfo{volume}{21}}, \bibinfo{pages}{5485--5551} (\bibinfo{year}{2020}).

\bibitem{brown2020language}
\bibinfo{author}{Brown, T.} \emph{et~al.}
\newblock \bibinfo{title}{Language models are few-shot learners}.
\newblock \emph{\bibinfo{journal}{Proceddings of the Advances in neural information processing systems}} \textbf{\bibinfo{volume}{33}}, \bibinfo{pages}{1877--1901} (\bibinfo{year}{2020}).

\bibitem{rezende2014stochastic}
\bibinfo{author}{Rezende, D.~J.}, \bibinfo{author}{Mohamed, S.} \& \bibinfo{author}{Wierstra, D.}
\newblock \bibinfo{title}{Stochastic backpropagation and approximate inference in deep generative models}.
\newblock \emph{\bibinfo{journal}{Proceddings of the International conference on machine learning}} \bibinfo{pages}{1278--1286} (\bibinfo{year}{2014}).

\bibitem{li2019specializing}
\bibinfo{author}{Li, X.~L.} \& \bibinfo{author}{Eisner, J.}
\newblock \bibinfo{title}{Specializing word embeddings (for parsing) by information bottleneck}.
\newblock \emph{\bibinfo{journal}{Proceddings of the Conference on Empirical Methods in Natural Language Processing and the International Joint Conference on Natural Langauge Processing}}  (\bibinfo{year}{2019}).

\bibitem{logothetis2003underpinnings}
\bibinfo{author}{Logothetis, N.~K.}
\newblock \bibinfo{title}{The underpinnings of the bold functional magnetic resonance imaging signal}.
\newblock \emph{\bibinfo{journal}{Journal of Neuroscience}} \textbf{\bibinfo{volume}{23}}, \bibinfo{pages}{3963--3971} (\bibinfo{year}{2003}).

\bibitem{tishby1999information}
\bibinfo{author}{Tishby, N.}, \bibinfo{author}{Pereira, F.~C.} \& \bibinfo{author}{Bialek, W.}
\newblock \bibinfo{title}{The information bottleneck method}.
\newblock \emph{\bibinfo{journal}{Proceddings of the Annual Allerton Conference on Communication, Control, and Computing}}  (\bibinfo{year}{1999}).

\bibitem{lebel2021voxelwise}
\bibinfo{author}{LeBel, A.}, \bibinfo{author}{Jain, S.} \& \bibinfo{author}{Huth, A.~G.}
\newblock \bibinfo{title}{Voxelwise encoding models show that cerebellar language representations are highly conceptual}.
\newblock \emph{\bibinfo{journal}{Journal of Neuroscience}} \textbf{\bibinfo{volume}{41}}, \bibinfo{pages}{10341--10355} (\bibinfo{year}{2021}).

\bibitem{toneva2019interpreting}
\bibinfo{author}{Toneva, M.} \& \bibinfo{author}{Wehbe, L.}
\newblock \bibinfo{title}{Interpreting and improving natural-language processing (in machines) with natural language-processing (in the brain)}.
\newblock \emph{\bibinfo{journal}{Proceddings of the Conference on Neural Information Processing Systems}} \textbf{\bibinfo{volume}{32}} (\bibinfo{year}{2019}).

\bibitem{jain2018incorporating}
\bibinfo{author}{Jain, S.} \& \bibinfo{author}{Huth, A.~G.}
\newblock \bibinfo{title}{Incorporating context into language encoding models for {fMRI}}.
\newblock \emph{\bibinfo{journal}{Proceddings of the Conference on Neural Information Processing Systems}}  (\bibinfo{year}{2018}).

\bibitem{caucheteux2022brains}
\bibinfo{author}{Caucheteux, C.} \& \bibinfo{author}{King, J.-R.}
\newblock \bibinfo{title}{Brains and algorithms partially converge in natural language processing}.
\newblock \emph{\bibinfo{journal}{Communications Biology}} \textbf{\bibinfo{volume}{5}}, \bibinfo{pages}{134} (\bibinfo{year}{2022}).

\bibitem{huth2016natural}
\bibinfo{author}{Huth, A.~G.}, \bibinfo{author}{De~Heer, W.~A.}, \bibinfo{author}{Griffiths, T.~L.}, \bibinfo{author}{Theunissen, F.~E.} \& \bibinfo{author}{Gallant, J.~L.}
\newblock \bibinfo{title}{Natural speech reveals the semantic maps that tile human cerebral cortex}.
\newblock \emph{\bibinfo{journal}{Nature}} \textbf{\bibinfo{volume}{532}}, \bibinfo{pages}{453--458} (\bibinfo{year}{2016}).

\bibitem{lillicrap2013preference}
\bibinfo{author}{Lillicrap, T.~P.} \& \bibinfo{author}{Scott, S.~H.}
\newblock \bibinfo{title}{Preference distributions of primary motor cortex neurons reflect control solutions optimized for limb biomechanics}.
\newblock \emph{\bibinfo{journal}{Neuron}} \textbf{\bibinfo{volume}{77}}, \bibinfo{pages}{168--179} (\bibinfo{year}{2013}).

\bibitem{zipser1988back}
\bibinfo{author}{Zipser, D.} \& \bibinfo{author}{Andersen, R.~A.}
\newblock \bibinfo{title}{A back-propagation programmed network that simulates response properties of a subset of posterior parietal neurons}.
\newblock \emph{\bibinfo{journal}{Nature}} \textbf{\bibinfo{volume}{331}}, \bibinfo{pages}{679--684} (\bibinfo{year}{1988}).

\bibitem{lecun1998gradient}
\bibinfo{author}{LeCun, Y.}, \bibinfo{author}{Bottou, L.}, \bibinfo{author}{Bengio, Y.} \& \bibinfo{author}{Haffner, P.}
\newblock \bibinfo{title}{Gradient-based learning applied to document recognition}.
\newblock \emph{\bibinfo{journal}{Proceedings of the IEEE}} \textbf{\bibinfo{volume}{86}}, \bibinfo{pages}{2278--2324} (\bibinfo{year}{1998}).

\bibitem{rumelhart1986learning}
\bibinfo{author}{Rumelhart, D.~E.}, \bibinfo{author}{Hinton, G.~E.} \& \bibinfo{author}{Williams, R.~J.}
\newblock \bibinfo{title}{Learning representations by back-propagating errors}.
\newblock \emph{\bibinfo{journal}{Nature}} \textbf{\bibinfo{volume}{323}}, \bibinfo{pages}{533--536} (\bibinfo{year}{1986}).

\bibitem{hahnloser2000permitted}
\bibinfo{author}{Hahnloser, R.} \& \bibinfo{author}{Seung, H.~S.}
\newblock \bibinfo{title}{Permitted and forbidden sets in symmetric threshold-linear networks}.
\newblock \emph{\bibinfo{journal}{Proceedings of the Advances in Neural Information Processing Systems}} \textbf{\bibinfo{volume}{13}} (\bibinfo{year}{2000}).

\bibitem{hahnloser2000digital}
\bibinfo{author}{Hahnloser, R.~H.}, \bibinfo{author}{Sarpeshkar, R.}, \bibinfo{author}{Mahowald, M.~A.}, \bibinfo{author}{Douglas, R.~J.} \& \bibinfo{author}{Seung, H.~S.}
\newblock \bibinfo{title}{Digital selection and analogue amplification coexist in a cortex-inspired silicon circuit}.
\newblock \emph{\bibinfo{journal}{Nature}} \textbf{\bibinfo{volume}{405}}, \bibinfo{pages}{947--951} (\bibinfo{year}{2000}).

\bibitem{fukushima1969visual}
\bibinfo{author}{Fukushima, K.}
\newblock \bibinfo{title}{Visual feature extraction by a multilayered network of analog threshold elements}.
\newblock \emph{\bibinfo{journal}{IEEE Transactions on Systems Science and Cybernetics}} \textbf{\bibinfo{volume}{5}}, \bibinfo{pages}{322--333} (\bibinfo{year}{1969}).

\bibitem{devlin2019bert}
\bibinfo{author}{Devlin, J.}, \bibinfo{author}{Chang, M.-W.}, \bibinfo{author}{Lee, K.} \& \bibinfo{author}{Toutanova, K.}
\newblock \bibinfo{title}{{BERT}: Pre-training of deep bidirectional transformers for language understanding}.
\newblock \emph{\bibinfo{journal}{Proceedings of the Annual Conference of the North American Chapter of the Association for Computational Linguistics: Human Language Technologies}}  (\bibinfo{year}{2019}).

\bibitem{myronenko20193d}
\bibinfo{author}{Myronenko, A.}
\newblock \bibinfo{title}{3d {MRI} brain tumor segmentation using autoencoder regularization}.
\newblock \emph{\bibinfo{journal}{Brainlesion: Glioma, Multiple Sclerosis, Stroke and Traumatic Brain Injuries: 4th International Workshop}} \bibinfo{pages}{311--320} (\bibinfo{year}{2019}).

\bibitem{ji20123d}
\bibinfo{author}{Ji, S.}, \bibinfo{author}{Xu, W.}, \bibinfo{author}{Yang, M.} \& \bibinfo{author}{Yu, K.}
\newblock \bibinfo{title}{{3D} convolutional neural networks for human action recognition}.
\newblock \emph{\bibinfo{journal}{IEEE Transactions on Pattern Analysis and Machine Intelligence}} \textbf{\bibinfo{volume}{35}}, \bibinfo{pages}{221--231} (\bibinfo{year}{2012}).

\bibitem{tillmann2003word}
\bibinfo{author}{Tillmann, C.} \& \bibinfo{author}{Ney, H.}
\newblock \bibinfo{title}{Word reordering and a dynamic programming beam search algorithm for statistical machine translation}.
\newblock \emph{\bibinfo{journal}{Computational Linguistics}} \textbf{\bibinfo{volume}{29}}, \bibinfo{pages}{97--133} (\bibinfo{year}{2003}).

\bibitem{fedorenko2010new}
\bibinfo{author}{Fedorenko, E.}, \bibinfo{author}{Hsieh, P.-J.}, \bibinfo{author}{Nieto-Casta{\~n}{\'o}n, A.}, \bibinfo{author}{Whitfield-Gabrieli, S.} \& \bibinfo{author}{Kanwisher, N.}
\newblock \bibinfo{title}{New method for {fMRI} investigations of language: defining {ROI}s functionally in individual subjects}.
\newblock \emph{\bibinfo{journal}{Journal of Neurophysiology}} \textbf{\bibinfo{volume}{104}}, \bibinfo{pages}{1177--1194} (\bibinfo{year}{2010}).

\bibitem{dash2020decoding}
\bibinfo{author}{Dash, D.}, \bibinfo{author}{Ferrari, P.} \& \bibinfo{author}{Wang, J.}
\newblock \bibinfo{title}{Decoding imagined and spoken phrases from non-invasive neural (meg) signals}.
\newblock \emph{\bibinfo{journal}{Frontiers in Neuroscience}} \textbf{\bibinfo{volume}{14}}, \bibinfo{pages}{290} (\bibinfo{year}{2020}).

\bibitem{pereira2018toward}
\bibinfo{author}{Pereira, F.} \emph{et~al.}
\newblock \bibinfo{title}{Toward a universal decoder of linguistic meaning from brain activation}.
\newblock \emph{\bibinfo{journal}{Nature Communications}} \textbf{\bibinfo{volume}{9}}, \bibinfo{pages}{963} (\bibinfo{year}{2018}).

\bibitem{mitchell2008predicting}
\bibinfo{author}{Mitchell, T.~M.} \emph{et~al.}
\newblock \bibinfo{title}{Predicting human brain activity associated with the meanings of nouns}.
\newblock \emph{\bibinfo{journal}{Science}} \textbf{\bibinfo{volume}{320}}, \bibinfo{pages}{1191--1195} (\bibinfo{year}{2008}).

\bibitem{farwell1988talking}
\bibinfo{author}{Farwell, L.~A.} \& \bibinfo{author}{Donchin, E.}
\newblock \bibinfo{title}{Talking off the top of your head: toward a mental prosthesis utilizing event-related brain potentials}.
\newblock \emph{\bibinfo{journal}{Electroencephalography and Clinical Neurophysiology}} \textbf{\bibinfo{volume}{70}}, \bibinfo{pages}{510--523} (\bibinfo{year}{1988}).

\bibitem{wang1999reading}
\bibinfo{author}{Wang, J.}, \bibinfo{author}{Chen, H.-C.}, \bibinfo{author}{Radach, R.} \& \bibinfo{author}{Inhoff, A.}
\newblock \emph{\bibinfo{title}{Reading {C}hinese script: A cognitive analysis}}  (\bibinfo{publisher}{Psychology Press}, \bibinfo{year}{1999}).

\bibitem{cattell1886time}
\bibinfo{author}{Cattell, J.~M.}
\newblock \bibinfo{title}{The time it takes to see and name objects}.
\newblock \emph{\bibinfo{journal}{Mind}} \textbf{\bibinfo{volume}{11}}, \bibinfo{pages}{63--65} (\bibinfo{year}{1886}).

\bibitem{xi2023unicorn}
\bibinfo{author}{Xi, N.} \emph{et~al.}
\newblock \bibinfo{title}{Uni{C}o{RN}: Unified cognitive signal reconstruction bridging cognitive signals and human language}.
\newblock \emph{\bibinfo{journal}{arXiv preprint arXiv:2307.05355}}  (\bibinfo{year}{2023}).

\bibitem{defossez2023decoding}
\bibinfo{author}{D{\'e}fossez, A.}, \bibinfo{author}{Caucheteux, C.}, \bibinfo{author}{Rapin, J.}, \bibinfo{author}{Kabeli, O.} \& \bibinfo{author}{King, J.-R.}
\newblock \bibinfo{title}{Decoding speech perception from non-invasive brain recordings}.
\newblock \emph{\bibinfo{journal}{Nature Machine Intelligence}} \bibinfo{pages}{1--11} (\bibinfo{year}{2023}).

\bibitem{tang2023semantic}
\bibinfo{author}{Tang, J.}, \bibinfo{author}{LeBel, A.}, \bibinfo{author}{Jain, S.} \& \bibinfo{author}{Huth, A.~G.}
\newblock \bibinfo{title}{Semantic reconstruction of continuous language from non-invasive brain recordings}.
\newblock \emph{\bibinfo{journal}{Nature Neuroscience}} \bibinfo{pages}{1--9} (\bibinfo{year}{2023}).

\bibitem{ozcelik2023science}
\bibinfo{author}{Ozcelik, F.} \& \bibinfo{author}{VanRullen, R.}
\newblock \bibinfo{title}{Natural scene reconstruction from {fMRI} signals using generative latent diffusion}.
\newblock \emph{\bibinfo{journal}{Scientific Reports}} \textbf{\bibinfo{volume}{13}} (\bibinfo{year}{2023}).

\bibitem{pasley2012reconstructing}
\bibinfo{author}{Pasley, B.~N.} \emph{et~al.}
\newblock \bibinfo{title}{Reconstructing speech from human auditory cortex}.
\newblock \emph{\bibinfo{journal}{PLoS Biology}} \textbf{\bibinfo{volume}{10}}, \bibinfo{pages}{e1001251} (\bibinfo{year}{2012}).

\bibitem{anumanchipalli2019speech}
\bibinfo{author}{Anumanchipalli, G.~K.}, \bibinfo{author}{Chartier, J.} \& \bibinfo{author}{Chang, E.~F.}
\newblock \bibinfo{title}{Speech synthesis from neural decoding of spoken sentences}.
\newblock \emph{\bibinfo{journal}{Nature}} \textbf{\bibinfo{volume}{568}}, \bibinfo{pages}{493--498} (\bibinfo{year}{2019}).

\bibitem{moses2021neuroprosthesis}
\bibinfo{author}{Moses, D.~A.} \emph{et~al.}
\newblock \bibinfo{title}{Neuroprosthesis for decoding speech in a paralyzed person with anarthria}.
\newblock \emph{\bibinfo{journal}{New England Journal of Medicine}} \textbf{\bibinfo{volume}{385}}, \bibinfo{pages}{217--227} (\bibinfo{year}{2021}).

\bibitem{willett2021high}
\bibinfo{author}{Willett, F.~R.}, \bibinfo{author}{Avansino, D.~T.}, \bibinfo{author}{Hochberg, L.~R.}, \bibinfo{author}{Henderson, J.~M.} \& \bibinfo{author}{Shenoy, K.~V.}
\newblock \bibinfo{title}{High-performance brain-to-text communication via handwriting}.
\newblock \emph{\bibinfo{journal}{Nature}} \textbf{\bibinfo{volume}{593}}, \bibinfo{pages}{249--254} (\bibinfo{year}{2021}).

\bibitem{pei2011decoding}
\bibinfo{author}{Pei, X.}, \bibinfo{author}{Barbour, D.~L.}, \bibinfo{author}{Leuthardt, E.~C.} \& \bibinfo{author}{Schalk, G.}
\newblock \bibinfo{title}{Decoding vowels and consonants in spoken and imagined words using electrocorticographic signals in humans}.
\newblock \emph{\bibinfo{journal}{Journal of Neural Engineering}} \textbf{\bibinfo{volume}{8}}, \bibinfo{pages}{046028} (\bibinfo{year}{2011}).

\bibitem{akbari2019towards}
\bibinfo{author}{Akbari, H.}, \bibinfo{author}{Khalighinejad, B.}, \bibinfo{author}{Herrero, J.~L.}, \bibinfo{author}{Mehta, A.~D.} \& \bibinfo{author}{Mesgarani, N.}
\newblock \bibinfo{title}{Towards reconstructing intelligible speech from the human auditory cortex}.
\newblock \emph{\bibinfo{journal}{Scientific Reports}} \textbf{\bibinfo{volume}{9}}, \bibinfo{pages}{874} (\bibinfo{year}{2019}).

\bibitem{stavisky2018decoding}
\bibinfo{author}{Stavisky, S.~D.} \emph{et~al.}
\newblock \bibinfo{title}{Decoding speech from intracortical multielectrode arrays in dorsal ``arm/hand'' areas of human motor cortex}.
\newblock \emph{\bibinfo{journal}{Proceedgins of the Annual International Conference of the IEEE Engineering in Medicine and Biology Society}} \bibinfo{pages}{93--97} (\bibinfo{year}{2018}).

\bibitem{bib1}
\bibinfo{author}{Campbell, S.~L.} \& \bibinfo{author}{Gear, C.~W.}
\newblock \bibinfo{title}{The index of general nonlinear {D}{A}{E}{S}}.
\newblock \emph{\bibinfo{journal}{Numer. {M}ath.}} \textbf{\bibinfo{volume}{72}}, \bibinfo{pages}{173--196} (\bibinfo{year}{1995}).

\bibitem{bib2}
\bibinfo{author}{Slifka, M.~K.} \& \bibinfo{author}{Whitton, J.~L.}
\newblock \bibinfo{title}{Clinical implications of dysregulated cytokine production}.
\newblock \emph{\bibinfo{journal}{J. {M}ol. {M}ed.}} \textbf{\bibinfo{volume}{78}}, \bibinfo{pages}{74--80} (\bibinfo{year}{2000}).

\bibitem{bib3}
\bibinfo{author}{Hamburger, C.}
\newblock \bibinfo{title}{Quasimonotonicity, regularity and duality for nonlinear systems of partial differential equations}.
\newblock \emph{\bibinfo{journal}{Ann. Mat. Pura. Appl.}} \textbf{\bibinfo{volume}{169}}, \bibinfo{pages}{321--354} (\bibinfo{year}{1995}).

\bibitem{bib4}
\bibinfo{author}{Geddes, K.~O.}, \bibinfo{author}{Czapor, S.~R.} \& \bibinfo{author}{Labahn, G.}
\newblock \emph{\bibinfo{title}{Algorithms for {C}omputer {A}lgebra}}  (\bibinfo{publisher}{Kluwer}, \bibinfo{address}{Boston}, \bibinfo{year}{1992}).

\bibitem{bib5}
\bibinfo{author}{Broy, M.}
\newblock \bibinfo{title}{ in \textit{Software engineering---from auxiliary to key technologies}} (eds \bibinfo{editor}{Broy, M.} \& \bibinfo{editor}{Denert, E.}) \emph{\bibinfo{booktitle}{Software Pioneers}} \bibinfo{pages}{10--13} (\bibinfo{publisher}{Springer}, \bibinfo{address}{New {Y}ork}, \bibinfo{year}{1992}).

\bibitem{bib6}
\bibinfo{editor}{Seymour, R.~S.} (ed.) \emph{\bibinfo{title}{Conductive {P}olymers}}  (\bibinfo{publisher}{Plenum}, \bibinfo{address}{New {Y}ork}, \bibinfo{year}{1981}).

\bibitem{bib7}
\bibinfo{author}{Smith, S.~E.}
\newblock \bibinfo{editor}{Zaimis, E.} (ed.) \emph{\bibinfo{title}{Neuromuscular blocking drugs in man}}.
\newblock (ed.\bibinfo{editor}{Zaimis, E.}) \emph{\bibinfo{booktitle}{Neuromuscular junction. {H}andbook of experimental pharmacology}}, Vol.~\bibinfo{volume}{42}, \bibinfo{pages}{593--660} (\bibinfo{publisher}{Springer}, \bibinfo{address}{Heidelberg}, \bibinfo{year}{1976}).

\bibitem{bib8}
\bibinfo{author}{Chung, S.~T.} \& \bibinfo{author}{Morris, R.~L.}
\newblock \bibinfo{title}{Isolation and characterization of plasmid deoxyribonucleic acid from streptomyces fradiae} (\bibinfo{year}{1978}).
\newblock \bibinfo{note}{Paper presented at the 3rd international symposium on the genetics of industrial microorganisms, University of {W}isconsin, {M}adison, 4--9 June 1978}.

\bibitem{bib9}
\bibinfo{author}{Hao, Z.}, \bibinfo{author}{AghaKouchak, A.}, \bibinfo{author}{Nakhjiri, N.} \& \bibinfo{author}{Farahmand, A.}
\newblock \bibinfo{title}{Global integrated drought monitoring and prediction system (gidmaps) data sets} (\bibinfo{year}{2014}).
\newblock \bibinfo{note}{Figshare \url{https://doi.org/10.6084/m9.figshare.853801}}.

\bibitem{bib10}
\bibinfo{author}{Babichev, S.~A.}, \bibinfo{author}{Ries, J.} \& \bibinfo{author}{Lvovsky, A.~I.}
\newblock \bibinfo{title}{Quantum scissors: teleportation of single-mode optical states by means of a nonlocal single photon} (\bibinfo{year}{2002}).
\newblock \bibinfo{note}{Preprint at \url{https://arxiv.org/abs/quant-ph/0208066v1}}.

\bibitem{bib11}
\bibinfo{author}{Beneke, M.}, \bibinfo{author}{Buchalla, G.} \& \bibinfo{author}{Dunietz, I.}
\newblock \bibinfo{title}{Mixing induced {CP} asymmetries in inclusive {B} decays}.
\newblock \emph{\bibinfo{journal}{Phys. {L}ett.}} \textbf{\bibinfo{volume}{B393}}, \bibinfo{pages}{132--142} (\bibinfo{year}{1997}).

\bibitem{bib12}
\bibinfo{author}{Stahl, B.}
\newblock \bibinfo{title}{deep{SIP}: deep learning of {S}upernova {I}a {P}arameters}, \bibinfo{version}{0.42}.
\newblock \bibinfo{howpublished}{Astrophysics {S}ource {C}ode {L}ibrary} (\bibinfo{year}{2020}).
\newblock {\href{https://ascl.net/2006.023}{{ascl:2006.023}}}.

\bibitem{bib13}
\bibinfo{author}{Abbott, T. M.~C.} \emph{et~al.}
\newblock \bibinfo{title}{{Dark Energy Survey Year 1 Results: Constraints on Extended Cosmological Models from Galaxy Clustering and Weak Lensing}}.
\newblock \emph{\bibinfo{journal}{Phys. Rev. D}} \textbf{\bibinfo{volume}{99}}, \bibinfo{pages}{123505} (\bibinfo{year}{2019}).

\bibitem{ralph2017neural}
\bibinfo{author}{Ralph, M. A.~L.}, \bibinfo{author}{Jefferies, E.}, \bibinfo{author}{Patterson, K.} \& \bibinfo{author}{Rogers, T.~T.}
\newblock \bibinfo{title}{The neural and computational bases of semantic cognition}.
\newblock \emph{\bibinfo{journal}{Nature reviews neuroscience}} \textbf{\bibinfo{volume}{18}}, \bibinfo{pages}{42--55} (\bibinfo{year}{2017}).

\bibitem{bi2021dual}
\bibinfo{author}{Bi, Y.}
\newblock \bibinfo{title}{Dual coding of knowledge in the human brain}.
\newblock \emph{\bibinfo{journal}{Trends in Cognitive Sciences}} \textbf{\bibinfo{volume}{25}}, \bibinfo{pages}{883--895} (\bibinfo{year}{2021}).

\bibitem{desikan2006automated}
\bibinfo{author}{Desikan, R.~S.} \emph{et~al.}
\newblock \bibinfo{title}{An automated labeling system for subdividing the human cerebral cortex on mri scans into gyral based regions of interest}.
\newblock \emph{\bibinfo{journal}{Neuroimage}} \textbf{\bibinfo{volume}{31}}, \bibinfo{pages}{968--980} (\bibinfo{year}{2006}).

\bibitem{jenkinson2012fsl}
\bibinfo{author}{Jenkinson, M.}, \bibinfo{author}{Beckmann, C.~F.}, \bibinfo{author}{Behrens, T.~E.}, \bibinfo{author}{Woolrich, M.~W.} \& \bibinfo{author}{Smith, S.~M.}
\newblock \bibinfo{title}{{FSL} (the {FMRIB} software library)}.
\newblock \emph{\bibinfo{journal}{Neuroimage}} \textbf{\bibinfo{volume}{62}}, \bibinfo{pages}{782--790} (\bibinfo{year}{2012}).

\bibitem{fedorenko2014reworking}
\bibinfo{author}{Fedorenko, E.} \& \bibinfo{author}{Thompson-Schill, S.~L.}
\newblock \bibinfo{title}{Reworking the language network}.
\newblock \emph{\bibinfo{journal}{Trends in Cognitive Sciences}} \textbf{\bibinfo{volume}{18}}, \bibinfo{pages}{120--126} (\bibinfo{year}{2014}).

\bibitem{binder2011neurobiology}
\bibinfo{author}{Binder, J.~R.} \& \bibinfo{author}{Desai, R.~H.}
\newblock \bibinfo{title}{The neurobiology of semantic memory}.
\newblock \emph{\bibinfo{journal}{Trends in Cognitive Sciences}} \textbf{\bibinfo{volume}{15}}, \bibinfo{pages}{527--536} (\bibinfo{year}{2011}).

\bibitem{tzourio2002automated}
\bibinfo{author}{Tzourio-Mazoyer, N.} \emph{et~al.}
\newblock \bibinfo{title}{Automated anatomical labeling of activations in {SPM} using a macroscopic anatomical parcellation of the {MNI} {MRI} single-subject brain}.
\newblock \emph{\bibinfo{journal}{Neuroimage}} \textbf{\bibinfo{volume}{15}}, \bibinfo{pages}{273--289} (\bibinfo{year}{2002}).

\end{thebibliography}

\begin{appendices}

\section{Extended Data}\label{sec:A1}


\begin{figure}[ht]
    \centering
    \begin{subfigure}[b]{0.85\textwidth}
        \includegraphics[width=\textwidth]{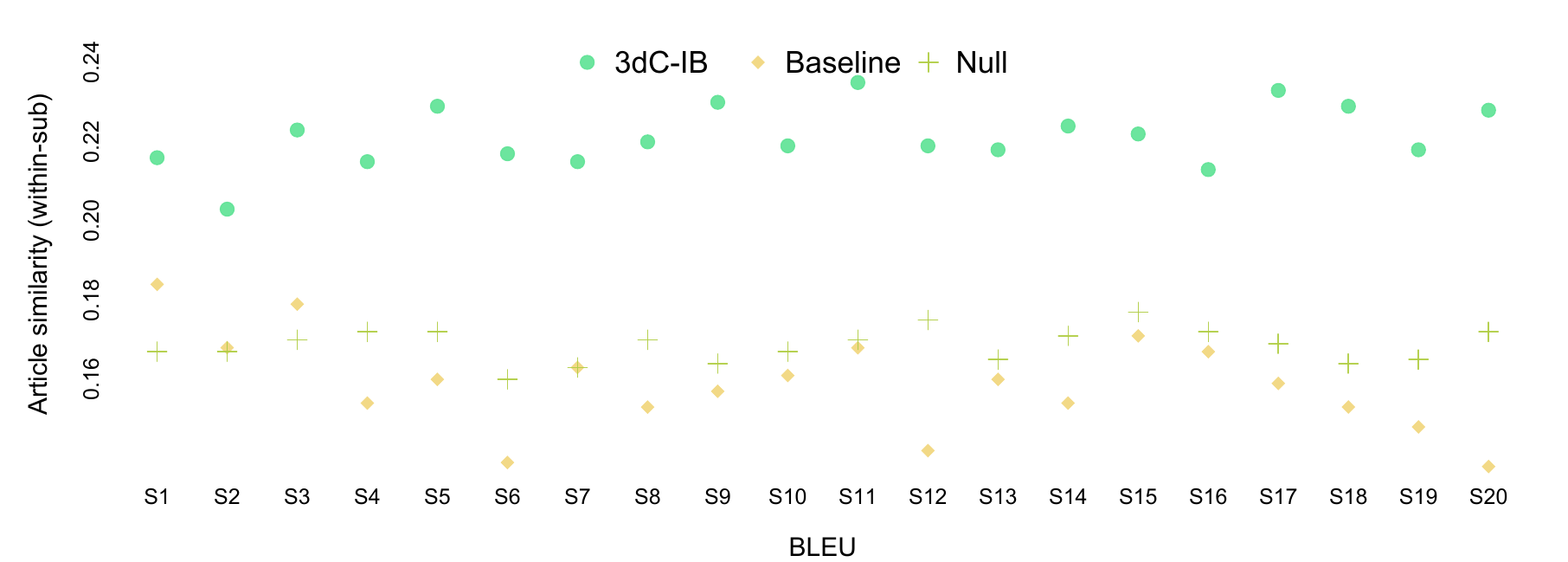}
        \label{fig:within-all-bleu}
    \end{subfigure}
    \begin{subfigure}[b]{0.85\textwidth}
        \includegraphics[width=\textwidth]{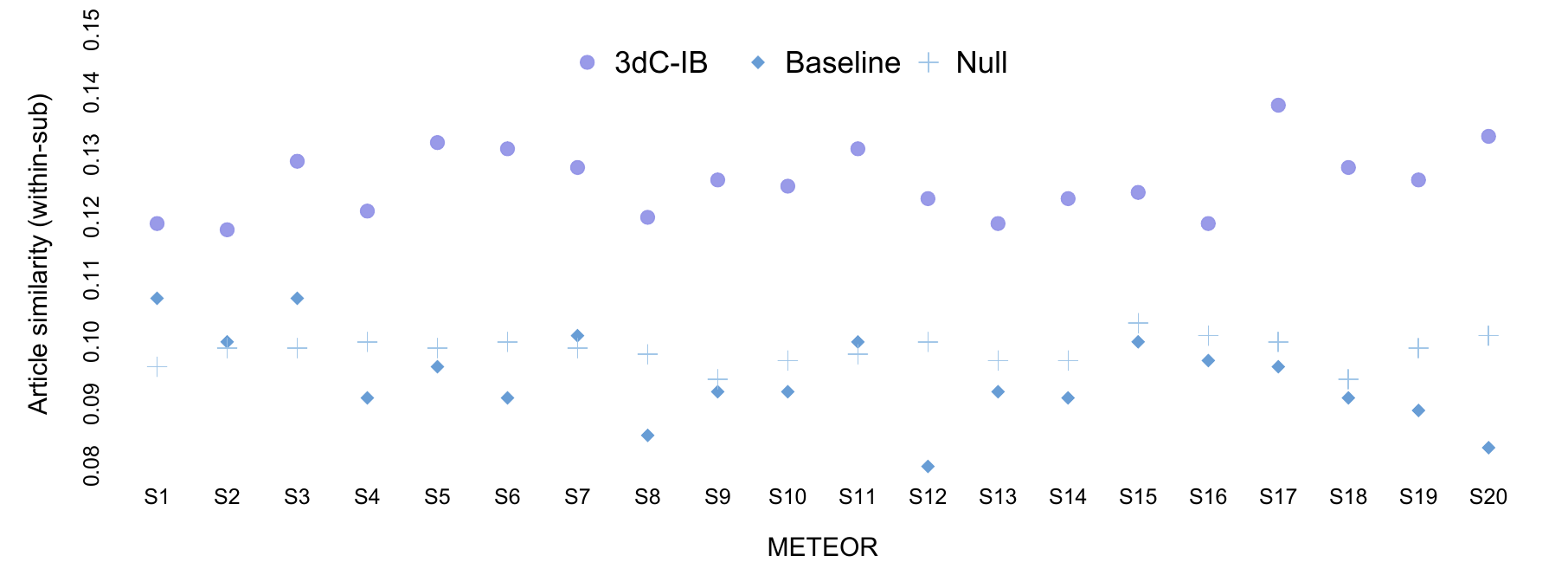}
        \label{fig:within-all-meteor}
    \end{subfigure}
    \begin{subfigure}[b]{0.85\textwidth}
        \includegraphics[width=\textwidth]{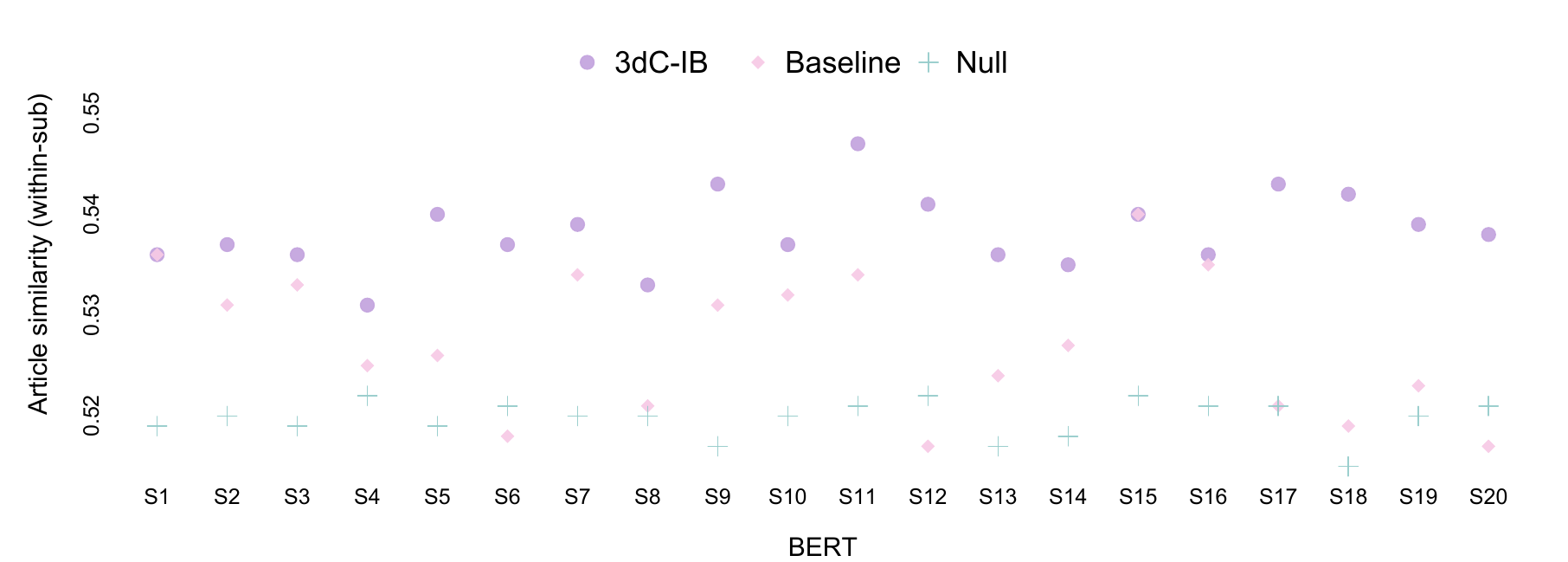}
        \label{fig:within-all-bert}
    \end{subfigure}
    \begin{subfigure}[b]{0.85\textwidth}
        \includegraphics[width=\textwidth]{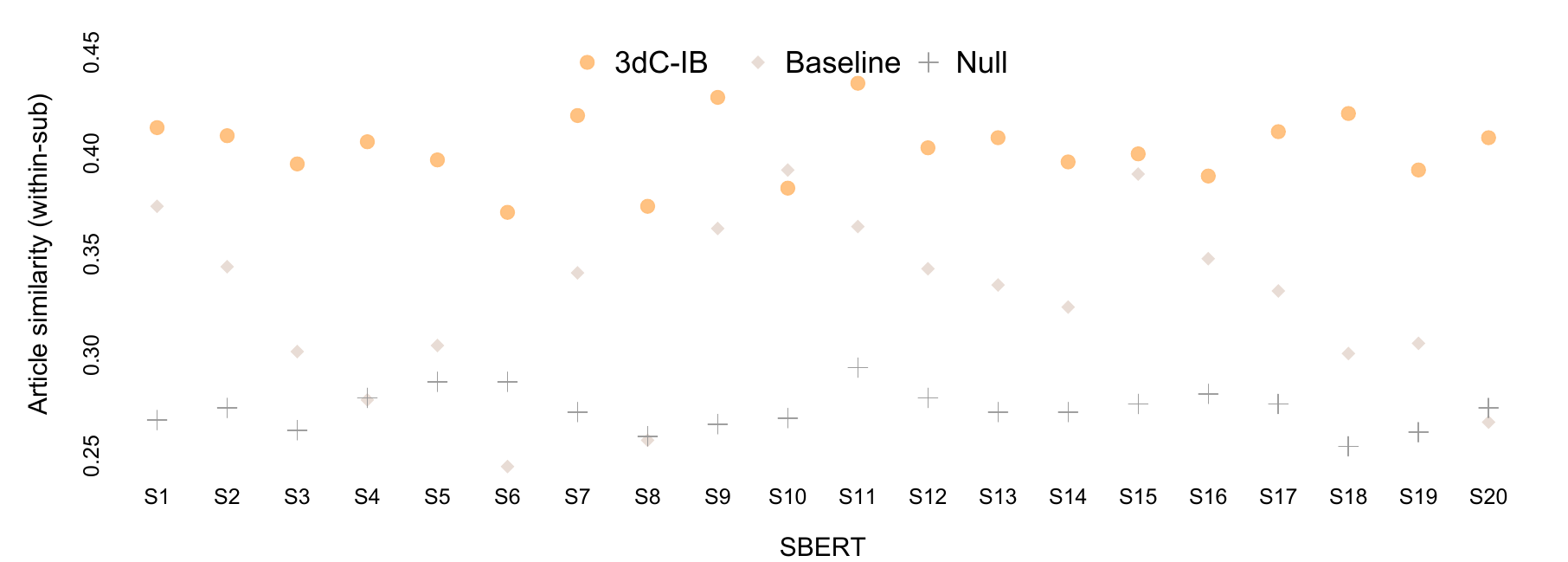}
        \label{fig:within-all-sbert}
    \end{subfigure}
    \caption{\small Language similarity scores in a within-subject setting. \textbf{(a)} BLEU,  \textbf{(b)} METEOR, \textbf{(C)} BERT, and \textbf{(d)} SBERT. Decoder predictions for a test article ($2,040$ characters) were significantly more similar to the actual stimulus character sequence compared to both the predictions of the baseline and those expected by chance ($P < 0.05$ for all subjects, one-sided non-parametric test).}
    \label{fig:within-all}
\end{figure}

\clearpage

\begin{figure}[ht]
    \centering
    \begin{subfigure}[b]{0.80\textwidth}
        \includegraphics[width=\textwidth]{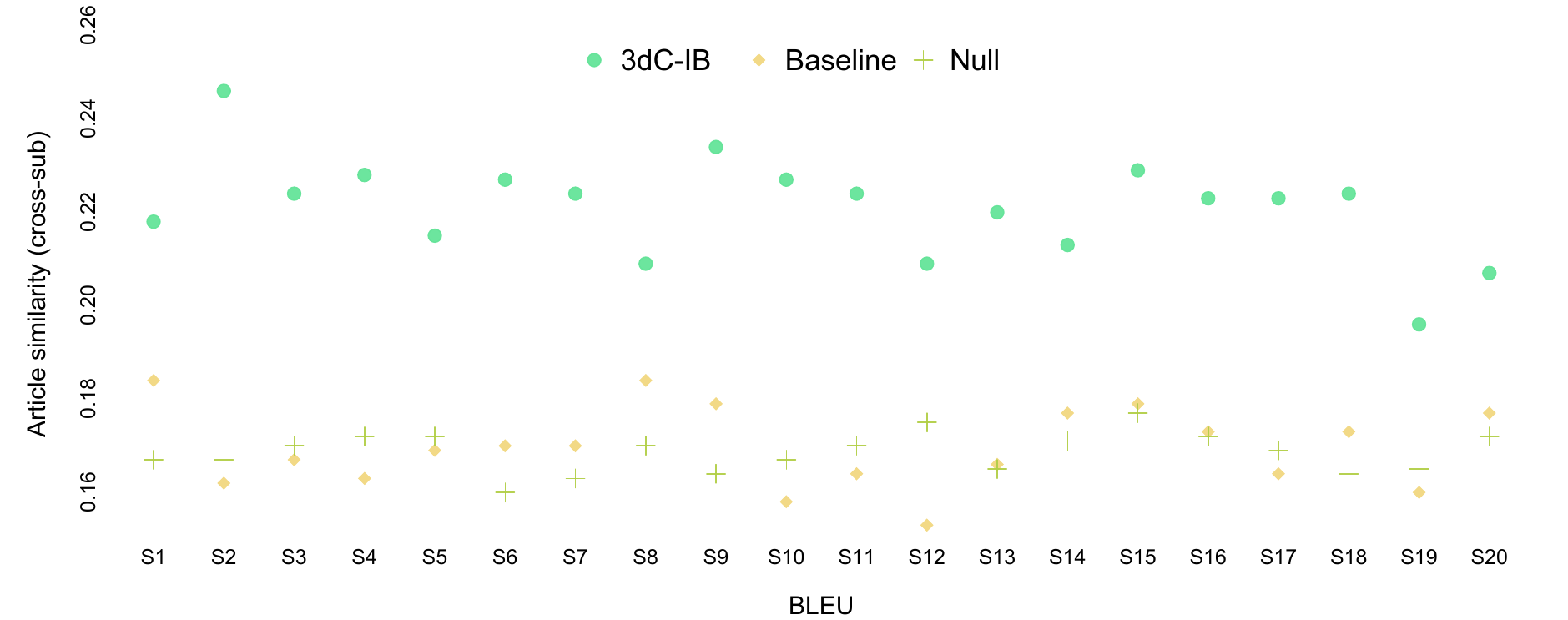}
        \label{fig:across-all-bleu}
    \end{subfigure}
    \begin{subfigure}[b]{0.80\textwidth}
        \includegraphics[width=\textwidth]{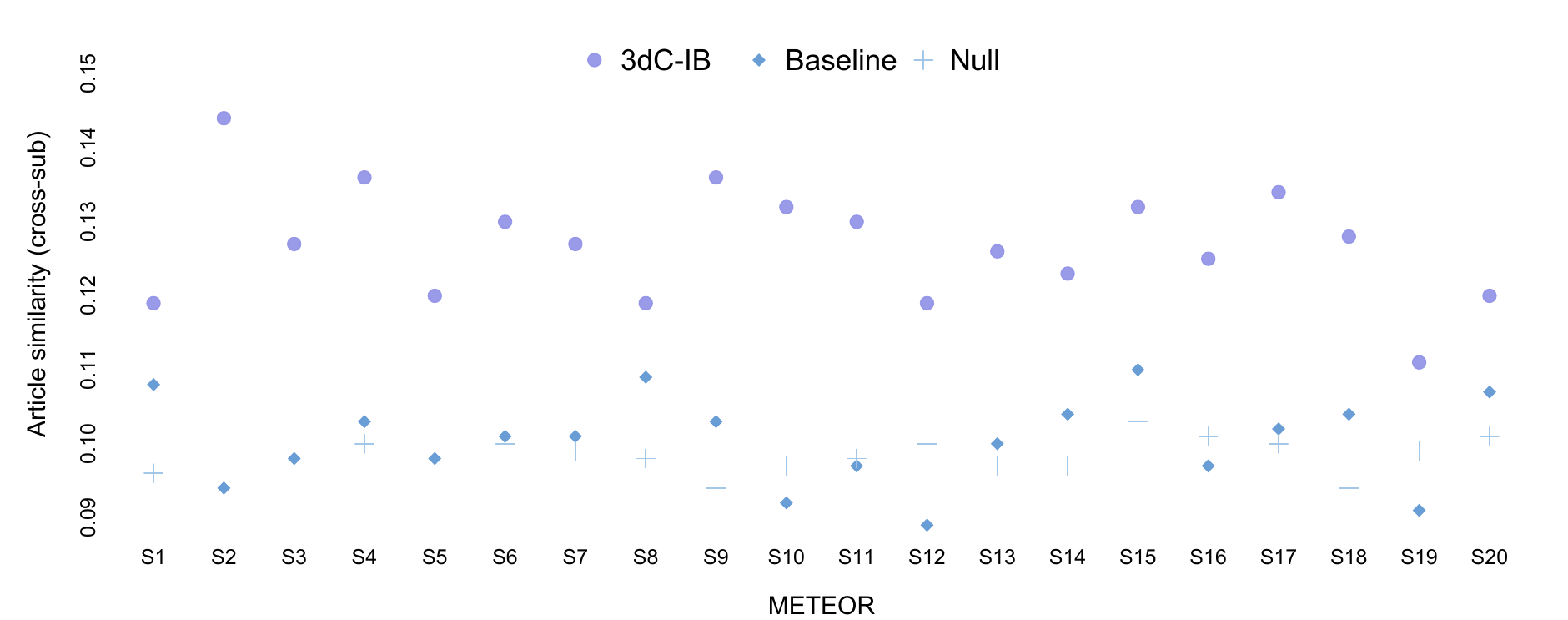}
        \label{fig:across-all-meteor}
    \end{subfigure}
    \begin{subfigure}[b]{0.80\textwidth}
        \includegraphics[width=\textwidth]{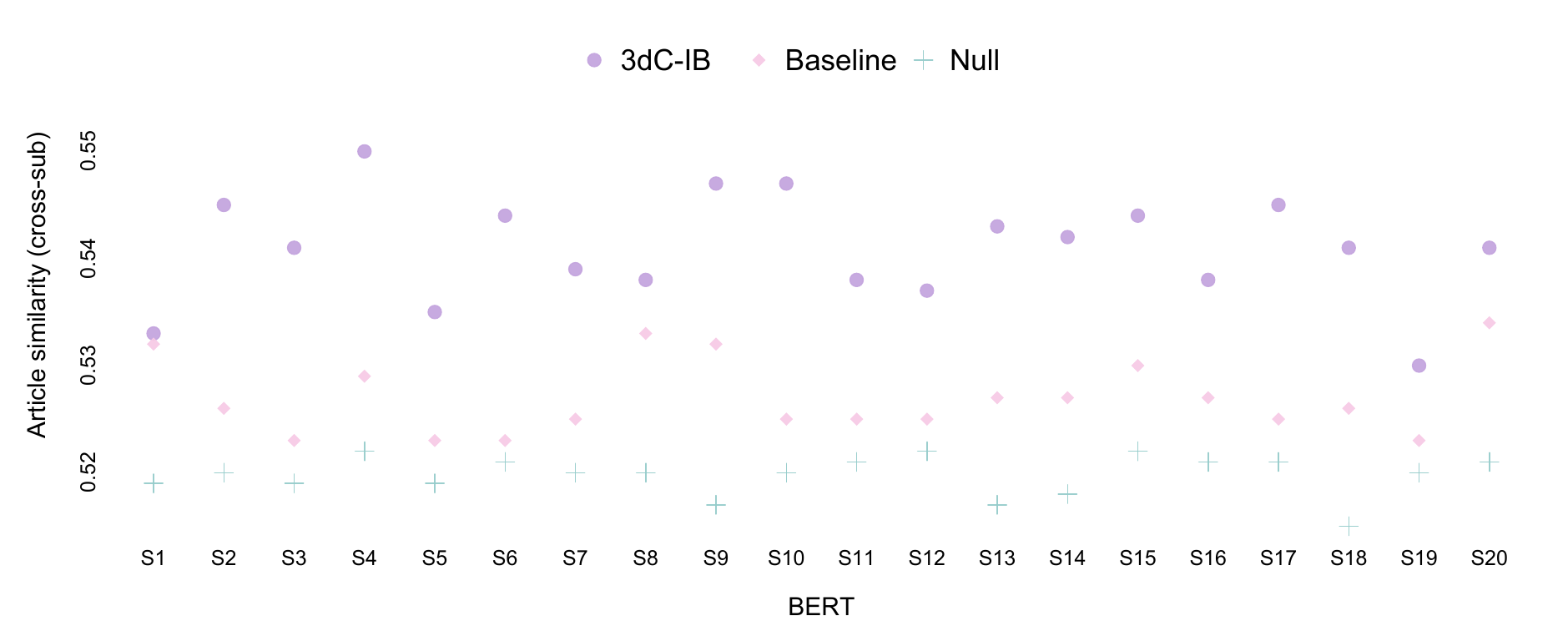}
        \label{fig:across-all-bert}
    \end{subfigure}
    \begin{subfigure}[b]{0.80\textwidth}
        \includegraphics[width=\textwidth]{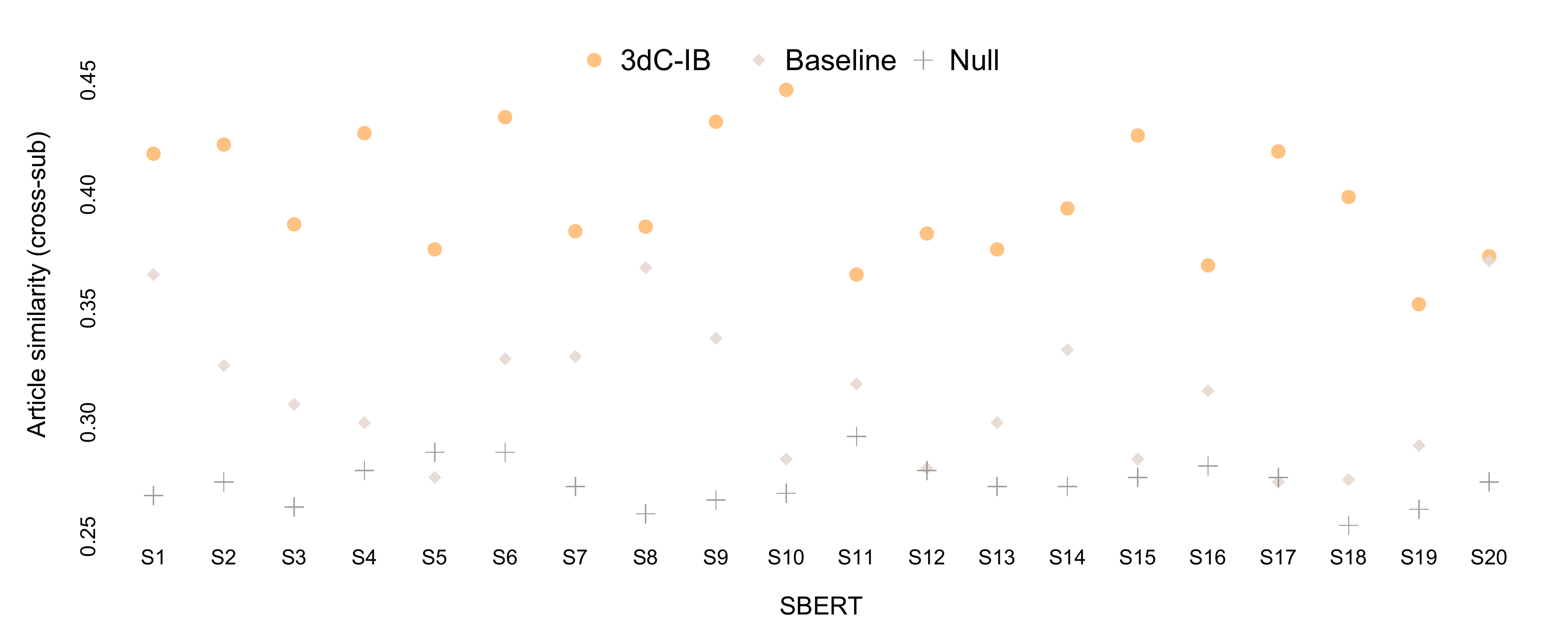}
        \label{fig:across-all-sbert}
    \end{subfigure}
    \caption{\small Language similarity scores in a cross-subject setting. \textbf{(a)} BLEU,  \textbf{(b)} METEOR, \textbf{(C)} BERT, and \textbf{(d)} SBERT. Decoder predictions for a test article ($2,040$ characters) were significantly more similar to the actual stimulus character sequence compared to both the predictions of the baseline and those expected by chance ($P < 0.05$ for all subjects, one-sided non-parametric test).}
    \label{fig:across-all}
\end{figure}

\clearpage

\begin{table}[h]
  \centering
  \caption{Summary of each article of the Chinese natural speech stimulus (part 1).}
  \label{tb:summary-1}
  \begin{tabular}{p{0.4cm}<{\centering}|p{2.4cm}<{\raggedright}|p{1.2cm}<{\raggedright}|p{7cm}}
    \hline
    \textbf{No.} & \multicolumn{1}{c|}{\textbf{Title}} & \multicolumn{1}{c|}{\textbf{Topic}} & \multicolumn{1}{c}{\textbf{Synopsis}} \\
    \hline
    $1$ & The Rabbit Who Went to the Moon & fairy tale & This story narrates the tale of the space agency's quest to find a rabbit to accompany the Brave spacecraft on a journey to the moon, where it would be left behind for scientific experiments. A white rabbit is chosen and undergoes a unique adventure, ultimately returning to Earth as an unexpected hero and global sensation. \\
    \hline
    $2$ & Severing the Friendship & fable & This story narrates the close friendship between Guan Ning and Hua Xin during their youth, which is put to the test when they discover gold and encounter a procession of a high-ranking official. Guan Ning emphasizes hard work and moral values, while Hua Xin is tempted by wealth without labor and material luxury. In the end, Guan Ning decides to sever their friendship because of their differing aspirations and values, symbolized by cutting a mat in half. \\
    \hline
    $3$ & The Family & novel & The story revolves around Gao Juexin, the elder brother of Gao Juemin. Despite sharing the same mother and household, their circumstances differ greatly. Gao Juexin is the eldest son and the firstborn in the extended family, shaping his destiny from birth. Gifted with good looks and intelligence, he earns admiration from his parents and teachers. However, his future takes an unexpected turn when his mother passes away and his father arranges a marriage for him, shattering his dreams of further education and the love he had for another woman. \\
    \hline
    $4$ & The Class Adviser & novel & The story unfolds as Song Baoqi, a former delinquent, is released from detention, leading to his enrollment in Zhang Junshi's class. Despite doubts from his colleagues, Teacher Zhang takes on the challenge of guiding and mentoring Song Baoqi, showcasing his commitment to making a difference in the troubled student's life. This story explores the complexities of education, understanding, and trust. \\
    \hline
    $5$ & The Color of Light & essay & When C\'{e}zanne painted apples in blue, it altered people's perceptions of color, from Matisse's blue sunflowers to Picasso's vibrant red figures. Artists chase absolute reality, though it is often elusive. The author recalls a moment at a party where the blue light transformed apples, evoking the essence of C\'{e}zanne's art. This essay explores how light and atmosphere influence our perception of color and how modern artists challenge traditional notions of reality. \\
    \hline
    $6$ & The Hometown Banyan Trees & essay & The author reflects on two venerable banyan trees near their residence, offering refreshing shade and a haven for children's play. These banyans evoke cherished memories, and the author frequently visits them with their child, reliving moments from the past, including the legend of a snake spirit. These banyan trees symbolize the unity and warmth of the village community, enduring even as time passes and individuals venture far from their hometown. \\
    \hline
  \end{tabular}
\end{table}

\begin{table}[h]
  \centering
  \caption{Summary of each article of the Chinese natural speech stimulus (part 2).}
  \label{tb:summary-2}
  \begin{tabular}{p{0.4cm}<{\centering}|p{2.4cm}<{\raggedright}|p{1.2cm}<{\raggedright}|p{7cm}}
    \hline
    \textbf{No.} & \multicolumn{1}{c|}{\textbf{Title}} & \multicolumn{1}{c|}{\textbf{Topic}} & \multicolumn{1}{c}{\textbf{Synopsis}} \\
    \hline    
    $7$ & Police Arrest 26 Suspects After Airing of ``Crazy Personal Information Black Market'' & news  & In February, CCTV News Channel broadcasted a program revealing the illegal trade of personal information online. Following this report, the Public Security Ministry formed a special investigative team. After nearly three months of investigation spanning Beijing and other regions, the case has been successfully resolved, resulting in the apprehension of $26$ individuals connected to criminal activities involving the theft and sale of personal information. \\
    \hline
    $8$ & Huabei Connects to the Central Bank's Credit System, Gradually Covering All User Groups & news & This news tells that as early spending becomes a daily habit, digital financial products like Huabei, Jiebei, JD Bai Tiao, and Weilidai gain popularity alongside credit cards. Recently, Huabei has upgraded its services by integrating with the central bank's credit system. While not all users are included, specific groups have been integrated, enabling the monthly reporting of their financial information. This step aims to enhance credit assessments' accuracy and improve the overall credit system. \\
    \hline
    $9$ & Petroleum Resources & scientific explanatory essay & Petroleum is formed underground under high pressure over extended periods and requires drilling for extraction, followed by separation and processing in refineries. Despite the increasing challenges of petroleum exploration, there are substantial undiscovered petroleum resources. However, petroleum extraction can have environmental impacts, and advanced technology and strict laws are helping to control these adverse impacts on the environment. \\
    \hline
    $10$ & Bombed by Negative News, Here's How to Maintain Mental Health & news & When bombarded by negative news, people often feel their mood affected, but this is a natural response of the human brain to threatening information. Over-empathizing or getting too immersed in negative comments can exacerbate negative emotions. Media literacy and the ability to discern false information become crucial, and active participation in public discussions and actions helps alleviate feelings of powerlessness. Focusing on change and taking action, rather than passively succumbing to news, is the best way to mitigate the sense of helplessness brought about by ``distant cries.'' \\
    \hline
    $11$ & School in the Clouds & fairy tale & Laughing Cat, Mouse Qiu Qiu, and the narrator visit Mouse Xiao Bai's hostess's villa, where they are initially alarmed by strange voices. However, they soon realize the villa is empty except for Lory, a voice-mimicking parrot. A sudden storm separates Laughing Cat and Mouse Qiu Qiu, but when Laughing Cat awakens, he reunites with Mouse Xiao Bai and together they rescue Mouse Qiu Qiu from a tree. Little White then shares the enchanting story of how he and his hostess arrived, riding a magical oiled-paper umbrella, and invites Laughing Cat to meet her the next day. \\
    \hline
  \end{tabular}
\end{table}

\begin{table}[h]
  \centering
  \caption{Summary of each article of the Chinese natural speech stimulus (part 3).}
  \label{tb:summary-3}
  \begin{tabular}{p{0.4cm}<{\centering}|p{2.4cm}<{\raggedright}|p{1.2cm}<{\raggedright}|p{7cm}}
    \hline
    \textbf{No.} & \multicolumn{1}{c|}{\textbf{Title}} & \multicolumn{1}{c|}{\textbf{Topic}} & \multicolumn{1}{c}{\textbf{Synopsis}} \\
    \hline   
    $12$ & The Battle of the Yellow Emperor Against Chi You & fable & In this ancient Chinese myth, Chi You, a descendant of the Flame Emperor, waged a war against the Yellow Emperor but was captured. With the help of the Kuafu tribe, Chi You's forces regained strength, challenging the Yellow Emperor. Seeking guidance from Xuan Nu, a celestial being, the Yellow Emperor obtained a magical sword and strategic knowledge. Equipped with these newfound resources, the Yellow Emperor reversed the course of the war, defeating Chi You and his allies. Chi You was executed, and his shackles became a grove of enduring red maple trees, signifying his legacy. This story illustrates the importance of wisdom, strategy, and unity in overcoming formidable challenges. \\
    \hline
    $13$ & Chronicle of a Blood Merchant & novel & This story narrates how a man named Xu Sanguan leisurely lounges in a melon field, savoring the sweet watermelon and engaging in a conversation about different melon varieties with his uncle. Unexpectedly, Xu Sanguan declares his desire to get married. Shortly after this declaration, he crosses paths with a young woman named Xu Yulan, and a newfound bond blossoms between them. \\
    \hline
    $14$ & Ordinary World & novel & The story describes Sun Shaoping's challenging life as a high school student from a poor rural background. He struggles with hunger and self-esteem issues due to his family's financial difficulties, yet he feels a sense of pride for having the opportunity to attend school in a larger world. Despite the hardships, he cherishes the experience of leaving his remote village for a broader horizon. \\
    \hline
    $15$ & Poetic Night Rain & essay & This essay reflects the profound impact of night rain on travelers, invoking both longing for home and a sense of tranquility. It illustrates how this natural phenomenon can shape emotions and even alter the course of history, affecting generals, advisors, kings, and heroes. Through the poet's perspective, night rain connects life's everyday experiences to deeper meanings, making it a rich source of inspiration. \\
    \hline
    $16$ & The Five Key Flavors of Chinese Food & essay & This article highlights the diverse palate of people across China. The tastes of Chinese food are traditionally categorized into five flavors: sour, sweet, bitter, spicy, salty, encompassing the vinegar affection of Shanxi residents, the spicy cravings of Sichuan locals, the fondness for sweets among Guangdong inhabitants, and the distinctive flavors of Beijing's stinky tofu, etc. It underscores the richness of Chinese cuisine and the depth of its culinary culture. \\
    \hline
  \end{tabular}
\end{table}

\begin{table}[h]
  \centering
  \caption{Summary of each article of the Chinese natural speech stimulus (part 4).}
  \label{tb:summary-4}
  \begin{tabular}{p{0.4cm}<{\centering}|p{2.4cm}<{\raggedright}|p{1.2cm}<{\raggedright}|p{7cm}}
    \hline
    \textbf{No.} & \multicolumn{1}{c|}{\textbf{Title}} & \multicolumn{1}{c|}{\textbf{Topic}} & \multicolumn{1}{c}{\textbf{Synopsis}} \\
    \hline 
    $17$ & Safeguarding the Vital Thread of Ethnic Unity  & news & This article showcases how Yunnan, a province rich in ethnic diversity, has made ethnic unity and progress a top priority, in line with the President of China, Xi Jinping's vision. Through focused poverty reduction initiatives, improved infrastructure, and economic diversification, Yunnan has uplifted its ethnic minority communities, bolstered by legal measures to safeguard and advance ethnic development, solidifying its unwavering dedication to ethnic unity. \\
    \hline
    $18$ & Chinese President Xi Jinping Empowers Entrepreneurs for Greater Economic Impact & news & Chinese President Xi Jinping chaired a symposium with entrepreneurs, recognizing their vital role in China's economy, especially during the COVID-19 pandemic. He encouraged them to uphold values like innovation, integrity, and social responsibility while addressing domestic and global challenges and promoting economic cooperation through reforms for a thriving global economy. \\
    \hline
    $19$ & Flowing Water on Mars & scientific explanatory essay & This article discusses how Mars' photos imply a history of substantial liquid water, seen through river-like runoff and outflow channels. These signs point to a thicker atmosphere and warmer climate about 4 billion years ago. Although debates continue regarding ancient Martian oceans and lakes, the presence of outflow channels stands as compelling evidence of prior water abundance, mostly preserved in subterranean ice, especially within polar regions. \\
    \hline
    $20$ & How Does the Brain Perceive the Passage of Time? & scientific explanatory essay & This article delves into the intricate ways our brains perceive time, highlighting the complexity of our internal timing systems. Different brain regions may host various internal clocks operating at different speeds, which collectively enable our ability to process time. Experiments, including neuroimaging and animal studies, indicate that neurons encode relative time rather than absolute time. This understanding may offer insights into broader questions regarding how the brain constructs our perception of reality. \\
    \hline
  \end{tabular}
\end{table}

\clearpage





\end{appendices}

\end{document}